\documentclass[runningheads]{llncs}

\usepackage{eccv}

\usepackage{eccvabbrv}

\usepackage{graphicx}
\usepackage{booktabs}

\usepackage[accsupp]{axessibility}  

\usepackage{microtype}

\usepackage[pagebackref,breaklinks,colorlinks,citecolor=eccvblue]{hyperref}

\usepackage{orcidlink}

\PassOptionsToPackage{table}{xcolor}
\usepackage[dvipsnames]{xcolor}
\usepackage[resetlabels]{multibib}
\usepackage{subfiles}
\usepackage{csquotes}
\usepackage{pifont}
\usepackage{textcomp}
\usepackage{todonotes}
\usepackage{blindtext}
\usepackage{soul}
\usepackage{graphicx}
\usepackage{mathtools}
\usepackage{multirow}
\usepackage{flushend}
\usepackage{booktabs}
\usepackage[accsupp]{axessibility} 
\usepackage{cuted}

\usepackage{amsmath} 
\usepackage{amssymb} 
\usepackage{amsfonts}
\usepackage{bm} 
\usepackage{siunitx}
\usepackage{cancel}
\usepackage{microtype}
\usepackage{xcolor}
\usepackage{caption}
\usepackage[export]{adjustbox}

\definecolor{mygreen}{HTML}{00A64F}
\definecolor{myred}{HTML}{ED1B23}

\newcommand{\cmark}{\text{\ding{51}}}
\newcommand{\xmark}{\text{\ding{55}}}

\newcommand{\tabref}[1]{Tab.~\ref{#1}}
\newcommand{\secref}[1]{Sec.~\ref{#1}}
\newcommand{\figref}[1]{Fig.~\ref{#1}}
\newcommand{\net}{LetsMap}

\begin{document}

\title{\net: Unsupervised Representation Learning for Label-Efficient Semantic BEV Mapping} 

\titlerunning{\net:  Unsupervised Representation Learning for Semantic BEV Mapping}

\author{Nikhil Gosala\inst{1} \and
K\"ursat Petek\inst{1} \and
B Ravi Kiran \inst{2} \and 
Senthil Yogamani \inst{3} \and \\
Paulo Drews-Jr \inst{4} \and
Wolfram Burgard \inst{5} \and 
Abhinav Valada \inst{1}}

\authorrunning{N.~Gosala et al.}

\institute{University of Freiburg, Germany \and
Qualcomm SARL France \and
QT Technologies Ireland Limited \and
Federal University of Rio Grande, Brazil \and
University of Technology Nuremberg, Germany \\
\url{http://letsmap.cs.uni-freiburg.de}}

\maketitle
\begin{abstract}
Semantic Bird's Eye View (BEV) maps offer a rich representation with strong occlusion reasoning for various decision making tasks in autonomous driving. However, most BEV mapping approaches employ a fully supervised learning paradigm that relies on large amounts of human-annotated BEV ground truth data. 
In this work, we address this limitation by proposing the first unsupervised representation learning approach to generate semantic BEV maps from a monocular frontal view (FV) image in a label-efficient manner. Our approach pretrains the network to independently reason about scene geometry and scene semantics using two disjoint neural pathways in an unsupervised manner and then finetunes it for the task of semantic BEV mapping using only a small fraction of labels in the BEV. We achieve label-free pretraining by exploiting spatial and temporal consistency of FV images to learn scene geometry while relying on a novel temporal masked autoencoder formulation to encode the scene representation. Extensive evaluations on the KITTI-360 and nuScenes datasets demonstrate that our approach performs on par with the existing state-of-the-art approaches while using only $1\%$ of BEV labels and no additional labeled data.
\end{abstract}    
\section{Introduction}
\label{sec:introduction}
Semantic Bird's Eye View (BEV) maps are essential for autonomous driving as they offer rich, occlusion-aware information for height-agnostic applications including object tracking, collision avoidance, and motion control. Instantaneous BEV map estimation that does not rely on large amounts of annotated data is crucial for the rapid deployment of autonomous vehicles in novel domains.
However, the majority of existing BEV mapping approaches follow a fully supervised learning paradigm and thus rely on large amounts of annotated data in BEV, which is extremely arduous to obtain and hinders the scalability of autonomous vehicles to novel environments~\cite{cit:bev-seg-pan2020vpn, cit:bev-seg-lu2019ved, cit:bev-seg-pon, cit:bev-seg-lss}. 
Recent works circumvent this problem by leveraging frontal view (FV) semantic labels for learning both scene geometry and generating BEV pseudolabels~\cite{cit:bev-seg-skyeye}, or by leveraging semi-supervised learning using pairs of labeled and unlabeled samples~\cite{cit:bev-seg-s2g2}. However, the reliance on FV labels as well as the integrated network design of both approaches gives rise to three main challenges: (1)~FV labels offer scene geometry supervision only along class boundaries which limits the geometric reasoning ability of the model; (2)~FV labels are dataset-specific and any change in class definition mandates full model retraining; and (3)~tightly coupled network designs hinder the quick adoption of latest advances from literature.\looseness=-1

\begin{figure}[t]
    \centering
    \includegraphics[width=0.79\linewidth]{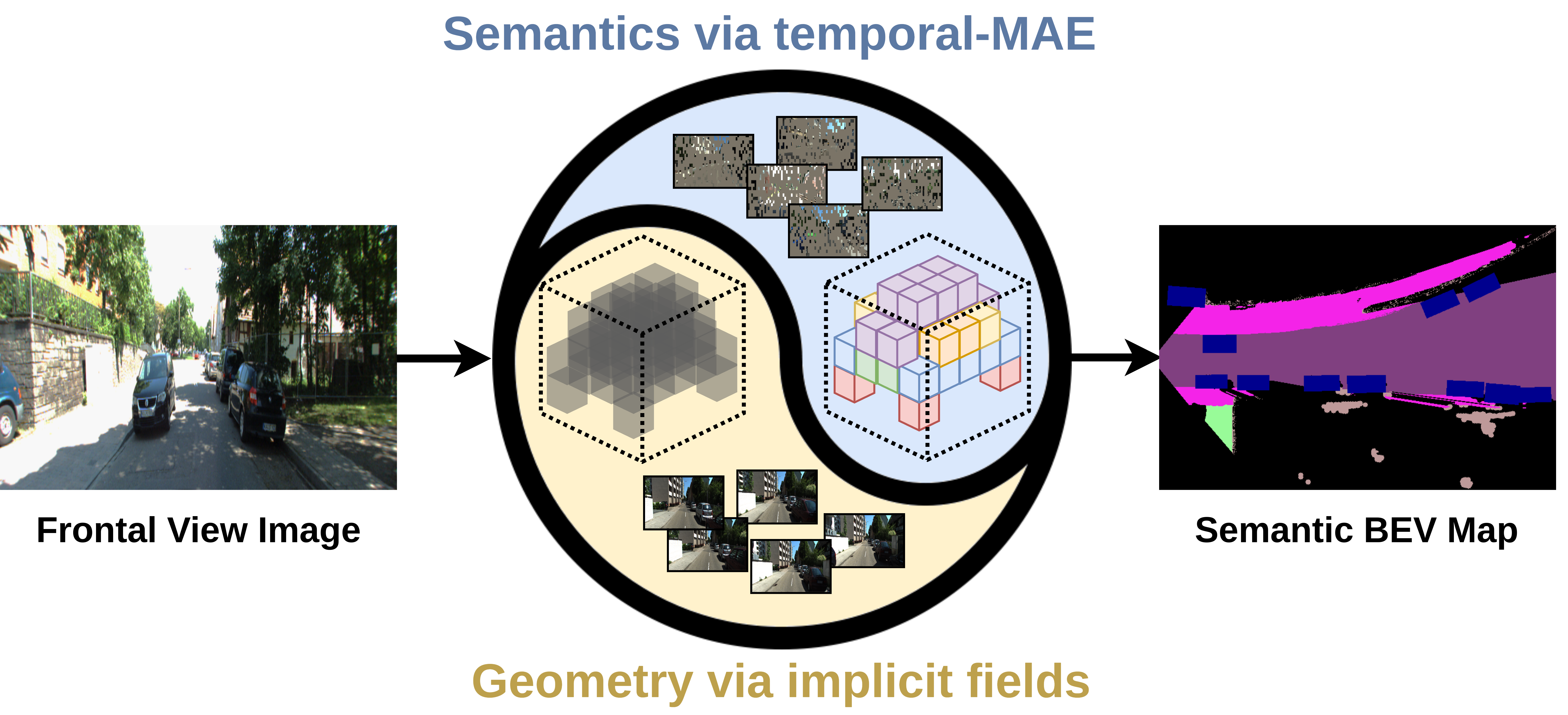}
    \vspace{-0.2cm}
    \caption{\net: The first unsupervised framework for label-efficient semantic BEV mapping. We use RGB image sequences to independently learn scene geometry (yellow) and scene representation (blue) in an unsupervised pretraining step, before adapting it to semantic BEV mapping in a label-efficient finetuning step.}
    \label{fig:teaser}
    \vspace{-0.5cm}
\end{figure}

In this work, we address these limitations by proposing the first unsupervised representation learning framework for predicting semantic BEV maps from monocular FV images in a label-efficient manner. Our approach, \net, utilizes the spatiotemporal consistency and dense representation offered by FV image sequences to alleviate the need for manually annotated data. To this end, we disentangle the two sub-tasks of semantic BEV mapping, i.e., \textit{scene geometry modeling} and \textit{scene representation learning}, into two disjoint neural pathways (\figref{fig:teaser}) and learn them using an unsupervised pretraining step. We then finetune the resultant model for semantic BEV mapping using only a small fraction of labels in BEV. \net~explicitly learns to model the scene geometry via the geometric pathway by leveraging implicit fields, while learning scene representations via the semantic pathway using a novel temporal masked autoencoder (T-MAE) mechanism. During pretraining, we supervise the geometric pathway by exploiting the spatial and temporal consistency of the multi-camera FV images across multiple timesteps and train the semantic pathway by enforcing reconstruction of the FV images for both the current and future timesteps using the masked image of only the current timestep. We extensively evaluate \net~on the KITTI-360~\cite{cit:dataset-kitti360} and nuScenes~\cite{cit:dataset-nuscenes} datasets and demonstrate that our approach performs on par with existing fully-supervised and self-supervised approaches while using only $1\%$ of BEV labels, without leveraging any additional labeled data.\looseness=-1
\section{Related Work}
\label{sec:related-work}

In this section, we discuss existing work on semantic BEV mapping, scene geometry estimation from monocular cameras, and image-based scene representation learning.

{\parskip=3pt
\noindent\textbf{BEV Segmentation}: Monocular semantic BEV mapping methods typically focus on learning a lifting mechanism to transform features from FV to BEV. Early works of VED~\cite{cit:bev-seg-lu2019ved} and VPN~\cite{cit:bev-seg-pan2020vpn} learn the transformation without using scene geometry, which limits their performance in the real world. PON~\cite{cit:bev-seg-pon} solves this issue by incorporating scene geometry into the network design while LSS~\cite{cit:bev-seg-lss} learns a depth distribution to transform features from FV to BEV. PanopticBEV~\cite{cit:bev-seg-panopticbev} splits the world into \textit{flat} and \textit{non-flat} regions and transforms them to BEV using two disjoint pathways. Recent methods use transformers to generate BEV features from both single image~\cite{cit:bev-seg-tiim} and multi-view images~\cite{cit:bev-seg-bevformerv2}. Some works also use multi-modal data to augment monocular cameras~\cite{cit:bev-seg-hdmapnet, cit:bev-seg-bevfusion, cit:bev-seg-simplebev, schramm2024bevcar}. All the aforementioned approaches follow a fully supervised learning paradigm and rely on vast amounts of resource-intensive human-annotated semantic BEV labels. Recent works reduce reliance on BEV ground truth labels by combining labeled and unlabeled images in a semi-supervised manner~\cite{cit:bev-seg-s2g2} or by leveraging FV labels to generate BEV pseudolabels and train the network in a self-supervised manner~\cite{cit:bev-seg-skyeye}. However, these approaches rely on additional labeled data or use tightly coupled network designs which limits their ability to scale to new environments or incorporate the latest advances in literature. In this paper, we propose a novel unsupervised label-efficient approach that first learns scene geometry and scene representation in a modular, label-free manner before adapting to semantic BEV mapping using only a small fraction of BEV semantic labels.} 

{\parskip=3pt
\noindent\textbf{Monocular Scene Geometry Estimation}:
Scene geometry estimation is a fundamental challenge in computer vision and is a core component of 3D scene reconstruction. Initial approaches use techniques such as multi-view stereo~\cite{furukawa2009accurate} and visual SLAM~\cite{cit:visual-slam,vodisch2022continual} while recent approaches leverage learnable functions in the form of ray distance functions~\cite{cit:ray-distance-functions} or implicit neural fields~\cite{cit:nerf-orig}. Early neural radiance fields-based approaches were optimized on single scenes and relied on substantial amounts of training data~\cite{cit:nerf-orig}. PixelNeRF~\cite{cit:pixel-nerf} addresses these issues by conditioning NeRF on input images, enabling simultaneous optimization across different scenes. Recent works improve upon PixelNeRF by decoupling color from scene density estimation~\cite{cit:behind-the-scenes}, and by using a tri-planar representation to query the neural field from any world point~\cite{cit:neo360}. In our approach, we leverage implicit fields to generate the volumetric density from a single monocular FV image to constrain features from the uniformly-lifted 2D scene representation features.}

{\parskip=3pt
\noindent\textbf{Scene Representation Learning}: Early works used augmentations such as image permutation~\cite{cit:ssl-image-permutation}, rotation prediction~\cite{cit:ssl-image-rotation}, noise discrimination~\cite{hindel2023inod}, and frame ordering~\cite{lang2024self} to learn scene representation; which were primitive and lacked generalization across diverse tasks. \cite{cit:ssl-moco, cit:ssl-simclr} propose using contrastive learning to learn scene representation, and \cite{cit:ssl-swav} builds upon this paradigm by removing the need for negative samples during training. Recent works propose masked autoencoders~\cite{cit:ssl-mae} wherein masked input image patches are predicted by the network using the learned high-level understanding of the scene. More recently, foundation models such as DINO~\cite{cit:dino-v1} and DINOv2~\cite{cit:dino-v2} employ self-distillation on large amounts of curated data to learn rich representations of the scene. However, all these approaches work on single timestep images and fail to leverage scene consistency over multiple timesteps. In this work, we explicitly enforce scene consistency over multiple timesteps by proposing a novel temporal masked autoencoding strategy to learn rich scene representations.}
\section{Technical Approach}
\label{sec:techincal-approach}

\begin{figure}[t]
    \centering
    \includegraphics[width=\textwidth]{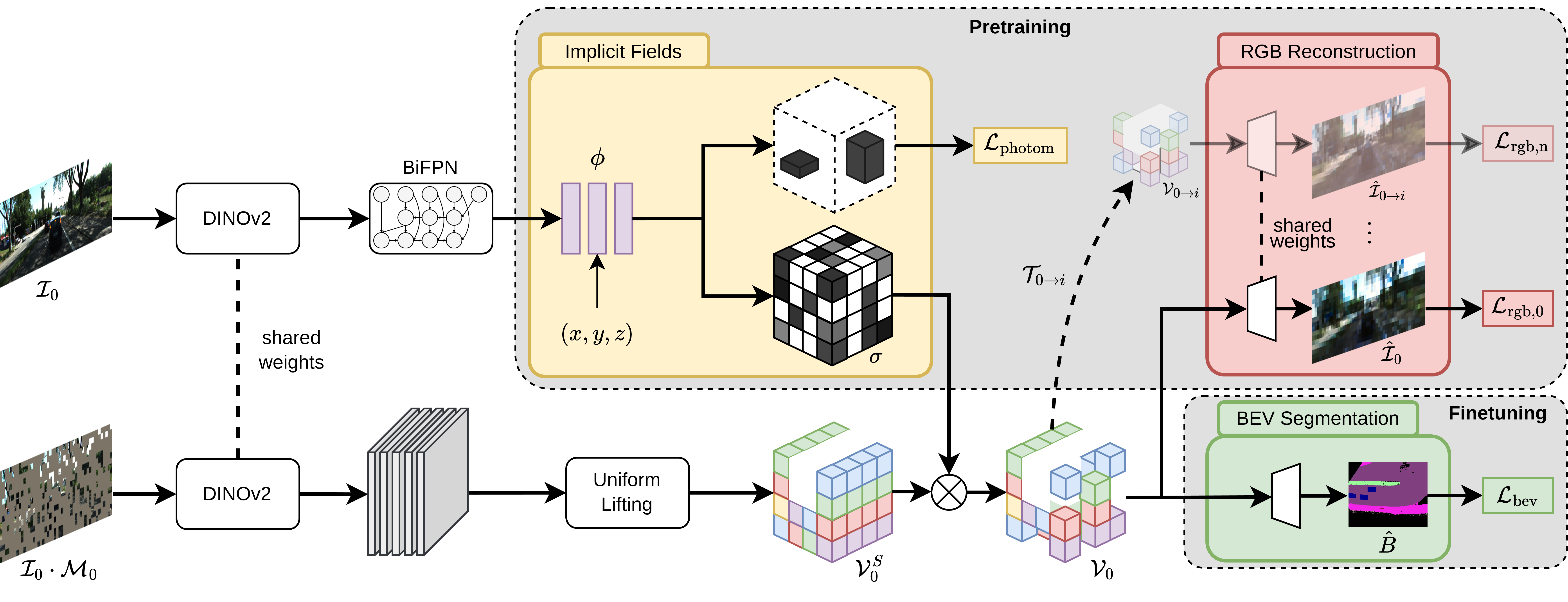}
    \vspace{-0.2cm}
    \caption{Overview of \net, our novel unsupervised representation learning framework for label-efficient semantic BEV mapping. The crux of our approach is to leverage FV image sequences to independently model scene geometry and learn scene representation using two disjoint pathways following an unsupervised training paradigm. The resulting model is then finetuned on a small fraction of BEV labels to the task of semantic BEV mapping.}
    \label{fig:overall-network}
    \vspace{-0.2cm}
\end{figure}

In this section, we present an overview of \net, the first unsupervised learning framework for predicting semantic BEV maps from monocular FV images using a label-efficient training paradigm. An overview of our framework is illustrated in \figref{fig:overall-network}. The key idea of our approach is to leverage sequences of multi-camera FV images to learn the two core sub-tasks of semantic BEV mapping, i.e., \textit{scene geometry modeling} and \textit{scene representation learning}, using two disjoint neural pathways following a label-free paradigm, before adapting it to the downstream task in a label-efficient manner. We achieve this desired behavior by splitting the training protocol into sequential FV pretraining and BEV finetuning stages. The FV pretraining stage learns to explicitly model the scene geometry by enforcing scene consistency over multiple views using the photometric loss ($\mathcal{L}_{\text{photom}}$, \secref{subsec:scene-geometry}) while learning the scene representation by reconstructing a masked input image over multiple timesteps using the reconstruction loss ($\mathcal{L}_{\text{rgb}}$, \secref{subsec:scene-semantics}).
Upon culmination of the pretraining phase, the finetuning phase adapts the network to the task of semantic BEV mapping using the cross-entropy loss on the tiny fraction of available BEV labels ($\mathcal{L}_{\text{bev}}$, \secref{subsec:bev-finetuning}). The total loss of the network is thus computed as:
\begin{equation}
\mathcal{L} = 
\begin{cases}
    \mathcal{L}_{\text{photom}} + \mathcal{L}_{\text{rgb}} & \text{when pretraining} \\
    \mathcal{L}_{\text{bev}}              & \text{when finetuning}
\end{cases}.
\end{equation}

\subsection{Network Architecture}
Our proposed \net~architecture, as shown in \figref{fig:overall-network}, consists of a pretrained DINOv2~\cite{cit:dino-v2} (ViT-b) backbone to generate multi-scale features from an input image; a geometry pathway comprising a convolution-based adapter followed by an implicit neural field to predict the scene geometry; a semantic pathway encompassing a sparse convolution-based adapter to capture representation-specific features; an RGB reconstruction head to facilitate reconstruction of the masked input image patches over multiple timesteps; and a BEV semantic head to generate a semantic BEV map from the input monocular FV image during the finetuning phase. 

During pretraining, an input image $\mathcal{I}_0$ is processed by the backbone to generate feature maps of three scales. The geometry pathway, $\mathcal{G}$, processes these multi-scale features using a BiFPN~\cite{tan2020efficientdet} layer followed by an implicit field module to generate the volumetric density of the scene at the current timestep. In a parallel branch, a masking module first randomly masks non-overlapping patches in $\mathcal{I}_0$ and the backbone then processes the visible patches to generate the corresponding image features. The semantic pathway $\mathcal{S}$ then generates the representation-specific features using a five-layer adapter that ensures propagation of masked regions using the convolution masking strategy outlined in~\cite{cit:spark-masked-convolution}. We then uniformly lift the resultant 2D features to 3D using the camera projection equation and multiply them with the volumetric density computed from $\mathcal{G}$ to generate scene-consistent voxel features. We warp the voxel grid to multiple timesteps using the ego-motion and collapse it into 2D by applying the camera projection equation along the depth dimension. The RGB reconstruction head then predicts the pixel values for each of the masked patches to reconstruct the image at different timesteps. During finetuning, we disable image masking and orthographically collapse the voxel features along the height dimension to generate the BEV features. A BEV semantic head processes these features to generate semantic BEV predictions. 

\subsection{Geometric Pathway}
\label{subsec:scene-geometry}

\begin{figure}
    \centering
    \begin{subfigure}{0.40\linewidth}
        \includegraphics[width=\linewidth]{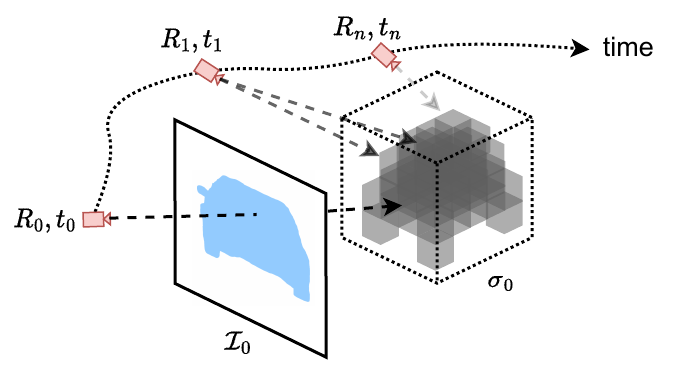}
        \vspace{-0.2cm}
        \caption{}
        \label{fig:nerf-arch}
    \end{subfigure}
    \hfill
    \begin{subfigure}{0.59\linewidth}
        \setlength{\tabcolsep}{0.05cm}
        {
        \renewcommand{\arraystretch}{0.2}
        \newcolumntype{M}[1]{>{\centering\arraybackslash}m{#1}}
        \begin{tabular}{M{3.5cm}M{3.5cm}M{3.5cm}}
        \includegraphics[width=\linewidth, height=0.455\linewidth, frame]{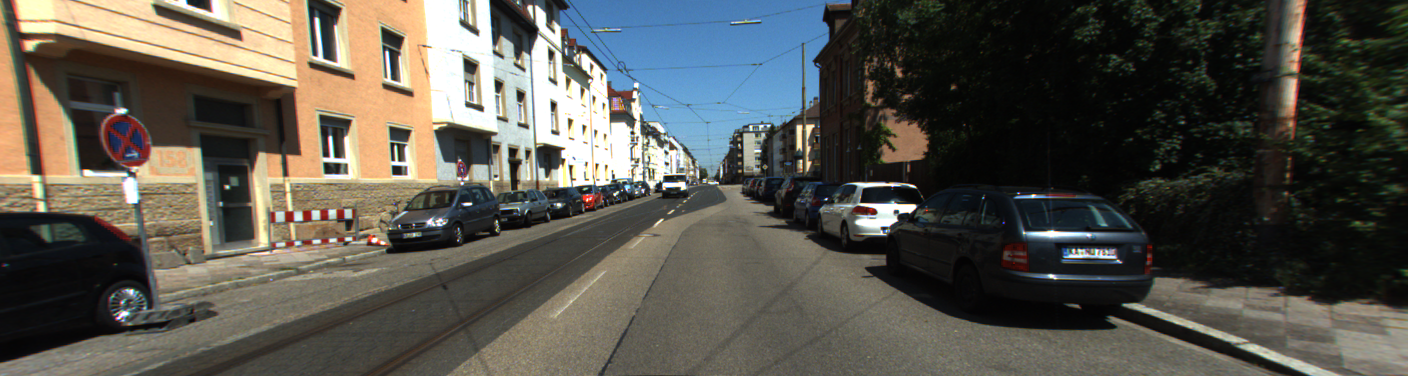} & \includegraphics[width=\linewidth, height=0.455\linewidth, frame]{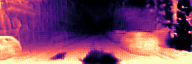} \\
        \\
        \includegraphics[width=\linewidth, height=0.455\linewidth, frame]{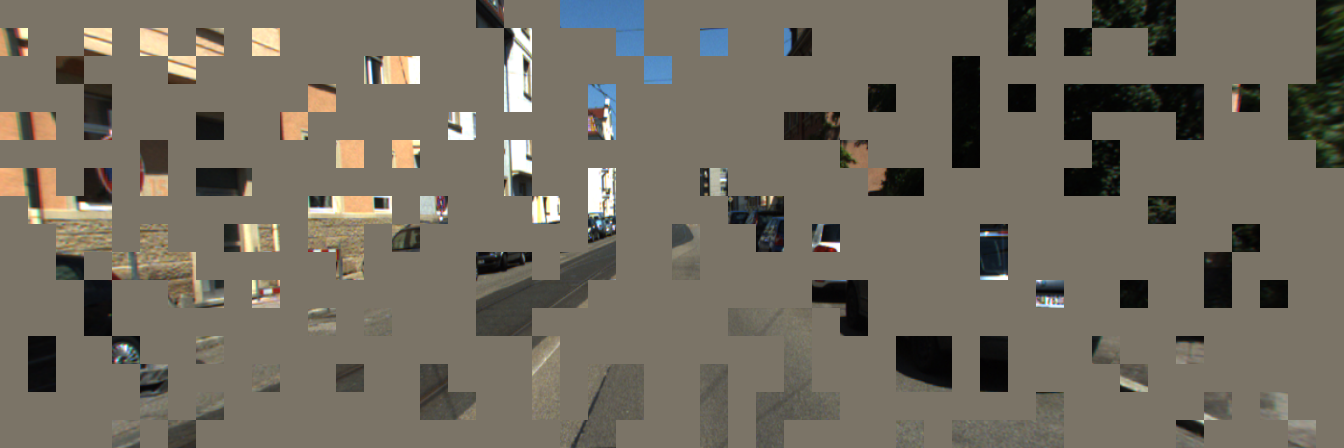} & \includegraphics[width=\linewidth, height=0.455\linewidth, frame]{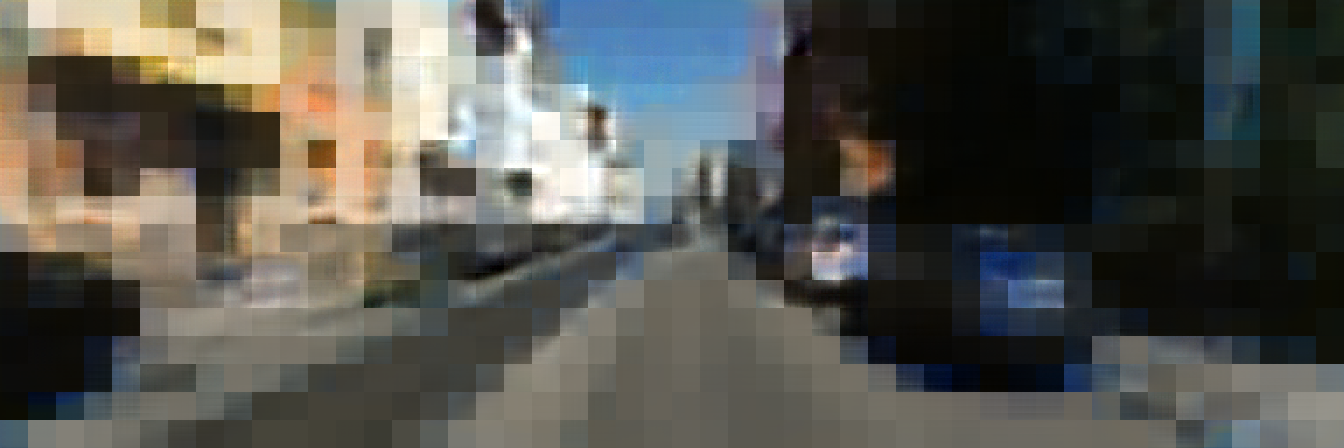} \\
        \\
        \end{tabular}
        }   
        \caption{}
        \label{fig:fv-pred}
    \end{subfigure}
    \caption{(a) An illustration of our neural implicit field module. It leverages spatio-temporal consistency offered by multi-camera images to model the scene geometry. (b) FV predictions from our unsupervised pretraining step. A FV image (top left) is processed by the geometry pathway to generate a volumetric density of the field which generates a depth map (top right) upon ray casting. Parallelly, a masked FV image (bottom left) is processed by the semantic pathway to reconstruct the masked image (bottom right).}
    \vspace{-0.5cm}
\end{figure}

The goal of the geometric pathway $\mathcal{G}$ is to explicitly model scene geometry in a label-free manner using only the spatio-temporal images obtained from cameras onboard an autonomous vehicle. Explicit scene geometry modeling allows the network to reason about occlusions and disocclusions in the scene, thus improving the quality of predictions in the downstream task. To this end, we design the task of scene geometry learning using an implicit field formulation wherein the main goal is to estimate the volumetric density of the scene in the camera coordinate system given a monocular FV image, as shown in \figref{fig:nerf-arch}. We multiply the estimated volumetric density with the uniformly-lifted semantic features to generate the geometrically consistent semantic features (see \secref{subsec:scene-semantics}). 

We generate the volumetric density for the scene by following the idea of image-conditioned NeRF outlined in~\cite{cit:pixel-nerf}. Firstly, we retrieve the image features $f$ for randomly sampled points, $\mathbf{x}=(x,y,z)$, along every camera ray by projecting them onto the 2D image plane and computing the value for each projection location using bilinear interpolation. We then pass the image features along with their positional encodings into a two-layer MLP, $\phi$, to estimate the volumetric density, $\sigma_\mathbf{x}$, at each of the sampled locations. Mathematically, the volumetric density at location $\mathbf{x}$ is computed as:
\begin{equation}
    \sigma_\mathbf{x} = \phi(f_{\mathbf{u}_\mathbf{x}}, \gamma(\mathbf{u}_\mathbf{x}, d_\mathbf{x})),
\end{equation}
where $\gamma(\cdot, \cdot)$ represents the sinusoidal positional encoding computed using the 2D projection $\mathbf{u}_\mathbf{x}$ of $\mathbf{x}$ on the image plane and the distance $d_\mathbf{x}$ of $\mathbf{x}$ from the camera origin.

During training, we optimize $\phi$ by first computing the depth map from $\sigma$ and then computing the photometric loss between the multi-view FV images at both the current as well as future timesteps. Specifically, for a camera ray through pixel location $\mathbf{u}$, we estimate the corresponding depth $\hat{d}_\mathbf{u}$ by computing the integral of intermediate depths over the probability of ray termination at a given distance. Accordingly, we sample $k$ points, $\mathbf{x}_1, \mathbf{x}_2, ..., \mathbf{x}_k$, on each camera ray and compute $\sigma$ at each of these locations. We then compute the probability of ray termination $\alpha_i$ between every pair of consecutive points ($\mathbf{x}_i$, $\mathbf{x}_{i+1}$) to determine the distance at which the ray is terminated, i.e., the depth $\hat{d}_\mathbf{u}$.
Mathematically, 
\begin{gather}
    \alpha_i = \text{exp}(1 - \sigma_{\mathbf{x}_i}\delta_i) \label{eqn:nerf-alpha}, \\
    \hat{d}_{\mathbf{u}} = \sum_{i=1}^{K} (\prod_{j=1}^{i-1}(1-\alpha_j))\alpha_i d_i, \label{eqn:newf-depth}
\end{gather}
where $d_i$ is the distance of $\mathbf{x}_i$ from the camera center, and $\delta_i = d_{i+1} - d_i$. A depth map output from $\mathcal{G}$ is illustrated in \figref{fig:fv-pred}. We use the computed depth map to supervise the geometric pathway $\mathcal{G}$ using the photometric loss between RGB images generated using inverse and forward warping. Inverse warping is described as:
\begin{equation}
        I'_{\text{tgt}, \text{inv}} (p_{\text{src}}) = I_{\text{tgt}}
        \langle K T_{\text{src} \rightarrow \text{tgt}} d(p_{\text{src}}) K^{-1} p_{\text{src}} \rangle, 
    \label{eqn:inverse-warping}
\end{equation}
where $K$ is the intrinsic camera matrix, $\langle\cdot\rangle$ denotes the bilinear sampling operator, and $p_{\text{src}}$ is a pixel coordinate in the source image. Similarly, forward warping is described as:
\begin{equation}
        I'_{\text{tgt}, \text{fwd}} (K T_{\text{src} \rightarrow \text{tgt}} d(p_\text{src})  K^{-1} p_{\text{src}}) = I_{\text{src}}(p_{\text{src}}), 
    \label{eqn:forward-warping}
\end{equation}
We reduce the impact of occlusions and disocclusions across timesteps from corrupting the overall photometric loss by only computing the pixelwise minimum for each of the forward and inverse photometric losses. The photometric loss is then computed as: 
\begin{equation}
    \mathcal{L}_\text{photom} = \lVert I'_{\text{tgt}, \text{fwd}} - I_{\text{tgt}} \rVert_1 + \lVert I'_{\text{tgt}, \text{inv}} - I_{\text{src}} \rVert_1
\end{equation}

\subsection{Semantic Pathway}
\label{subsec:scene-semantics}
The semantic pathway $\mathcal{S}$ aims to facilitate the learning of holistic feature representations for various scene elements in a label-free manner. This rich pretrained representation enables efficient adaptation to semantic classes during finetuning. To this end, we learn the representations of scene elements by masking out random patches in the input image and then forcing the network to generate pixel-wise predictions for each of the masked patches (\figref{fig:fv-pred}). Moreover, we also exploit the temporal consistency of static elements in the scene by reconstructing the RGB images at future timesteps $t_1, t_2, ..., t_n$ using the masked RGB input at timestep $t_0$. This novel formulation of temporal masked autoencoding (T-MAE) allows our network to learn spatially- and semantically consistent features which improve its occlusion reasoning ability and accordingly its performance on semantic BEV mapping. 

Our semantic pathway $\mathcal{S}$, shown in \figref{fig:overall-network}, masks the input image $\mathcal{I}_0$ using a binary mask $M_0$ with a masking ratio $m$, and generates the corresponding masked semantic 3D voxel grid $V^{\mathcal{S}}_0$. We then multiply $V^{\mathcal{S}}_0$ with the volumetric density $\sigma$ obtained from the geometric pathway $\mathcal{G}$ to generate the intermediate masked voxel grid $V_0$. During pretraining, we densify $V_0$ by filling the masked regions using a common mask token \texttt{[M]}, and generating pseudo voxel grids $V_{0\rightarrow i}$ by warping $V_0$ using the known camera poses between the current and the $i^{\text{th}}$ timesteps. Mathematically,
\begin{equation}
    V_{0 \rightarrow i} = T_{0 \rightarrow i} V_0,
\end{equation}
where $T_{0 \rightarrow i}$ is the transformation between camera poses at timesteps $t_0$ and $t_i$. We then independently use the voxel grids $V_0, V_{0 \rightarrow 1}, V_{0 \rightarrow 2}, ..., V_{0 \rightarrow i}$ as inputs to an RGB reconstruction head to reconstruct the RGB images $\hat{\mathcal{I}}_0, \hat{\mathcal{I}}_{0\rightarrow 1}, \hat{\mathcal{I}}_{0\rightarrow 2}, ..., \hat{\mathcal{I}}_{0\rightarrow i}$. We compute the L\textsubscript{2} loss on the normalized pixel values of every patch between $\mathcal{I}_k$ and $\hat{\mathcal{I}}_k$ to generate the supervision for the semantic pathway $\mathcal{S}$. We thus compute the reconstruction loss as:
\begin{equation}
    \mathcal{L}_\text{rgb} = \sum_{i=0}^{n}\lVert \mathcal{I}^{p}_i - \hat{\mathcal{I}}^{p}_{0 \rightarrow i} \rVert_{2},
\end{equation}
where $\mathcal{I}^{p}$ denotes the per-patch normalized image.

\subsection{BEV Finetuning}
\label{subsec:bev-finetuning}
We set up the network for finetuning by disabling image masking and discarding the RGB reconstruction head. We finetune the network on semantic BEV mapping by training the model on a fraction of BEV ground truth semantic labels using the cross entropy loss function. Mathematically,
\begin{equation}
    \mathcal{L}_\text{bev} = CE(B, \hat{B}),
\end{equation}
where $B$ and $\hat{B}$ are the semantic BEV ground truth and semantic BEV prediction masks, respectively.
\section{Experimental Results}
\label{sec:experiments}

In this section, we present quantitative and qualitative results of our novel unsupervised label-efficient semantic BEV mapping framework, \net, and provide extensive ablative experiments to demonstrate the benefit of our proposed contributions.

\subsection{Datasets}
We evaluate \net~on two large-scale autonomous driving datasets, i.e., KITTI-360~\cite{cit:dataset-kitti360} and nuScenes~\cite{cit:dataset-nuscenes}. Since neither dataset provides semantic BEV labels, we adopt the label generation pipeline outlined in PoBEV~\cite{cit:bev-seg-panopticbev} with minor modifications to discard the \textit{occlusion} mask to generate the semantic BEV ground truth labels. We sample one forward-facing perspective image from either fisheye camera for multi-camera supervision in KITTI-360 but use only a single camera when training on nuScenes due to the lack of sufficient field-of-view overlap between the spatial cameras. For the KITTI-360 dataset, we hold out sequence $10$ for validation and use the remaining $8$ sequences for training. For the nuScenes dataset, we follow the train-val split from~\cite{cit:bev-seg-pon} and obtain $702$ train and $142$ validation sequences.

\subsection{Training Protocol}
We train \net~on images of size $448 \times 1344$, and $448 \times 896$ for KITTI-360 and nuScenes, respectively. 
We select these image sizes to ensure compatibility with both the image encoder as well as the lower scales of the BiFPN adapter module since they are divisible by both $14$ and $32$.
The pretraining phase follows a label-free paradigm and trains the network using only spatio-temporal FV images with a window size of $4$, masking ratio of $0.75$, and masking patch size of $28$ for $20$ epochs with an initial learning rate (LR) of $0.005$ which is decayed by a factor of $0.5$ at epoch $15$ and $0.2$ at epoch $18$. We finetune the network on the task of semantic BEV mapping for $100$ epochs using only $1\%$ of BEV labels for the KITTI-360 dataset and one sample from every scene for the nuScenes dataset ($\approx\frac{1}{40}\%$). We use an LR of $0.005$ during finetuning and decay it by a factor of $0.5$ at epoch $75$ and $0.2$ at epoch $90$. We optimize \net~using the SGD optimizer with a batch size of $12$, momentum of $0.9$, and weight decay of $0.0001$.

\subsection{Quantitative Results}
We evaluate the performance of \net~on the KITTI-360 dataset by comparing it with the self-supervised approach SkyEye~\cite{cit:bev-seg-skyeye} as well as the fully-supervised baselines outlined in SkyEye. However, since SkyEye cannot be trained on nuScenes due to the lack of FV labels, we compare our approach with only the fully-supervised baselines on the nuScenes dataset. For all experiments, we use the code provided by the authors and ensure fair comparison by using the training protocols described in their original manuscripts. We use the standard mIoU metric for quantifying the performance~\cite{hurtado2022semantic}. \tabref{tab:quant-eval-overall-kitti} and \tabref{tab:quant-eval-overall-nuscenes} present the results of this evaluation for KITTI-360 and nuScenes respectively. For these experiments, we report metrics obtained when fully-supervised approaches are trained using $100\%$ of BEV labels, SkyEye is pretrained using $100\%$ of FV labels and finetuned on a tiny fraction of BEV labels, while \net~is trained on only a tiny fraction of BEV labels, i.e., $1\%$ on KITTI-360 and one sample per scene ($\approx\frac{1}{40}\%$) on nuScenes.

\begin{table}[t]
\caption{Evaluation of semantic BEV mapping on the KITTI-360 dataset. All metrics are reported in $[\%]$.}
\vspace{-0.3cm}
\scriptsize
\centering
\setlength\tabcolsep{1.5pt}
 \begin{tabular}{l|cc|cccccccc|c}
 \toprule
 \textbf{Method} & \textbf{FV} & \textbf{BEV} & \textbf{Road} & \textbf{Side.} & \textbf{Build.} & \textbf{Terrain} & \textbf{Person} & \textbf{2-Wh.} & \textbf{Car} & \textbf{Truck} & \textbf{mIoU}
 \\
 \midrule

IPM~\cite{cit:bev-seg-ipm-original} & 100\% & - & 53.03 & 24.90 & 15.19 & 32.31 & 0.20 & 0.36 & 11.59 & 1.90 & 17.44 \\
VED~\cite{cit:bev-seg-lu2019ved} & - & 100\% & 65.97 & \textbf{35.41} & \textbf{37.28} & 34.34 & 0.13 & 0.07 & 23.83 & 8.89 & 25.74 \\
VPN~\cite{cit:bev-seg-pan2020vpn} & - & 100\% & 69.90 & 34.31 & 33.65 & 40.17 & 0.56 & 2.26 & 27.76 & 6.10 & 26.84 \\
PON~\cite{cit:bev-seg-pon} & - & 100\% & 67.98 & 31.13 & 29.81 & 34.28 & 2.28 & 2.16 & 37.99 & 8.10 & 26.72 \\
PoBEV~\cite{cit:bev-seg-panopticbev} & - & 100\% & \textbf{70.14} & 35.23 & 34.68 & \textbf{40.72} & \textbf{2.85} & \textbf{5.63} & \textbf{39.77} & \textbf{14.38} & \textbf{30.42} \\
\cmidrule{1-12}
PoBEV~\cite{cit:bev-seg-panopticbev} & - & 1\% & 60.41 & 20.97 & 24.65 & 23.38 & 0.15 & 0.23 & 21.71 & 1.23 & 19.09 \\
SkyEye~\cite{cit:bev-seg-skyeye} & 100\% & 1\% & 69.26 & 33.48 & 32.79 & 39.46 & 0.00 & 0.34 & 32.36 & 7.93 & 26.94 \\
\net~(Ours) & 0\% & 1\% & 70.58 & 34.26 & 40.68 & 38.53 & 1.35 & 4.74 & 30.94 & 10.58 & 28.96 \\
\bottomrule
\end{tabular}
\label{tab:quant-eval-overall-kitti}
\vspace{-0.3cm}
\end{table}

\begin{table}[t]
\caption{Evaluation of semantic BEV mapping on the nuScenes dataset. All metrics are reported in $[\%]$.}
\vspace{-0.3cm}
\scriptsize
\centering
\setlength\tabcolsep{1.5pt}
 \begin{tabular}{l|cc|cccccccc|c}
 \toprule
 \textbf{Method} & \textbf{FV} & \textbf{BEV} & \textbf{Road} & \textbf{Side.} & \textbf{Manm.} & \textbf{Terrain} & \textbf{Person} & \textbf{2-Wh.} & \textbf{Car} & \textbf{Truck} & \textbf{mIoU}
 \\
 \midrule
IPM~\cite{cit:bev-seg-ipm-original} & 100\% & - & 43.51 & 9.05 & 26.21 & 16.60 & 0.14 & 0.72 & 4.65 & 3.67 & 13.07 \\
VED~\cite{cit:bev-seg-lu2019ved} & - & 100\% & 67.97 & 25.23 & 49.69 & 31.51 & 0.80 & 1.28 & 21.85 & 17.51 & 26.98  \\
VPN~\cite{cit:bev-seg-pan2020vpn} & - & 100\% & 66.47 & 23.94 & 47.65 & 33.19 & 2.02 & 4.13 & 22.66 & 18.33 & 27.30 \\
PON~\cite{cit:bev-seg-pon} & - & 100\% &  67.50 & 24.49 & 47.02 & 30.86 & 2.49 & 6.85 & 26.68 & 18.85 & 28.09 \\
PoBEV~\cite{cit:bev-seg-panopticbev} & - & 100\% & \textbf{70.15} & \textbf{27.87} & \textbf{50.04} & \textbf{35.32} & \textbf{3.89} & \textbf{7.06} & \textbf{31.60} & \textbf{21.27} & \textbf{30.90} \\
\cmidrule{1-12}
PoBEV~\cite{cit:bev-seg-panopticbev} & - & $\approx\frac{1}{40}\%$ & 64.55 & 19.85 & 45.21 & 28.45 & 1.20 & 1.06 & 20.45 & 11.48 & 24.03\\
\net~(Ours) & 0\% & $\approx\frac{1}{40}\%$ & \textbf{67.72} & \textbf{27.06} & \textbf{47.10} & \textbf{34.78} & \textbf{3.31} & \textbf{5.79} & \textbf{21.92} & \textbf{13.57} & \textbf{27.66} \\
\bottomrule
\end{tabular}
\label{tab:quant-eval-overall-nuscenes}
\vspace{-0.3cm}
\end{table}

We observe from \tabref{tab:quant-eval-overall-kitti} that our approach, \net, outperforms four of the five fully-supervised baselines by more than \SI{2}{pp} while using only $1\%$ of BEV labels. Notably, \net~also exceeds SkyEye by \SI{2.02}{pp} without using any additional labeled data. We also note that our approach significantly outperforms SkyEye on the static classes of \textit{road} and \textit{building}, as well as the dynamic classes of \textit{person}, \textit{2-wheeler}, and \textit{truck}. This improvement stems from explicit modeling of both scene geometry and representation which ensures well-constrained extents of dynamic objects as well as efficient mapping of scene elements to BEV classes using only $1\%$ of BEV labels. Although better than SkyEye, we observe that \net~underperforms PoBEV for most dynamic classes, reporting \SI{8.83}{pp} and \SI{3.80}{pp} lower on \textit{car} and \textit{truck} respectively. This is likely due to either insufficient views for training the implicit field or the presence of moving objects which results in its sub-optimal performance. Increasing the number of timesteps and sampling more perspective images from the fisheye cameras could address this limitation. 

On the nuScenes dataset, we note that \net~is comparable to most of the fully-supervised baselines but is consistently outperformed by the state-of-the-art approach PoBEV. nuScenes, being extremely dynamic and diverse, presents a significant challenge to our implicit field formulation which enforces a static scene constraint. This is especially evident in the \textit{car} and \textit{truck} classes which report \SI{9.68}{pp} and \SI{7.70}{pp} lower than PoBEV. Nonetheless, \net~is able to efficiently learn the scene representations of static classes, resulting in a comparable performance with all baselines while using only $1\%$ of annotated~data.

\subsection{Ablation Study}
\label{subsec:ablation-study}

\begin{table}[t]
 \caption{Ablation study on the impact of our unsupervised pretraining on the overall network performance. The column ``FV'' shows whether the models leverage FV pretraining, and the column ``PT'' denotes whether the models have been pretrained. All experiments are on the KITTI-360 dataset.}
 \vspace{-0.3cm}
\centering
\scriptsize
 \begin{tabular}{c|lcc|c|cccccccc|c}
 \toprule
 \textbf{BEV} & \textbf{Model} & \textbf{FV} & \textbf{PT} & \textbf{Epochs} & \textbf{Road} & \textbf{Side} & \textbf{Build} & \textbf{Terr.} & \textbf{Pers.} & \textbf{2-Wh.} & \textbf{Car} & \textbf{Truck} & \textbf{mIoU} \\
 \midrule
 \multirow{4}{*}{1\%} & PoBEV  & \xmark & - & \multirow{4}{*}{100} & 60.41 & 20.97 & 24.65 & 23.38 & 0.15 & 0.23 & 21.71 & 1.23 & 19.09 \\
 & SkyEye  & \cmark & \cmark & & 69.26 & 33.48 & 32.79 & \textbf{39.46} & 0.00 & 0.34 & \textbf{32.36} & 7.93 & 26.94 \\
 & \net & \xmark & \xmark & & 69.40 & 32.09 & 34.75 & 35.27 & 1.01 & 2.79 & 28.76 & 7.66 & 26.47 \\
 & \net & \xmark & \cmark & & \textbf{70.58} & \textbf{34.26} & \textbf{40.68} & 38.53 & \textbf{1.35} & \textbf{4.74} & 30.94 & \textbf{10.58} & \textbf{28.96} \\
 \midrule
 \multirow{4}{*}{5\%} & PoBEV  & \xmark & - & \multirow{4}{*}{80} & 64.45 & 27.36 & 30.15 & 31.66 & 0.69 & 0.98 & 29.75 & 6.06 & 23.89 \\
 & SkyEye  & \cmark & \cmark & & 72.16 & 37.20 & 34.89 & \textbf{42.97} & \textbf{4.77} & \textbf{9.16} & \textbf{40.74} & 9.88 & 31.47 \\
 & \net & \xmark & \xmark & & 72.80 & 37.89 & 38.59 & 40.06 & 2.34 & 5.62 & 34.86 & 16.26 & 31.05 \\
 & \net & \xmark & \cmark & & \textbf{73.74} & \textbf{39.56} & \textbf{42.07} & 41.49 & 2.46 & 6.32 & 34.68 & \textbf{14.88} & \textbf{31.90} \\
 \midrule
 \multirow{4}{*}{10\%} & PoBEV  & \xmark & - & \multirow{4}{*}{50} & 66.58 & 30.28 & 31.76 & 34.50 & 1.22 & 3.28 & 33.43 & 7.56 & 26.08  \\
 & SkyEye  & \cmark & \cmark & & 73.36 & 38.30 & 37.54 & \textbf{44.62} & \textbf{4.80} & \textbf{9.67} & \textbf{42.84} & 10.06 & 32.65 \\
 & \net & \xmark & \xmark & & 74.31 & 38.45 & 40.04 & 41.26 & 3.19 & 6.02 & 35.56 & 16.53 & 31.92 \\
 & \net & \xmark & \cmark & & \textbf{74.74} & \textbf{39.40} & \textbf{43.63} & 43.33 & 2.91 & 6.95 & 37.62 & \textbf{18.09} & \textbf{33.33} \\
 \midrule
 \multirow{4}{*}{50\%} & PoBEV  & \xmark & - & \multirow{4}{*}{30} & 69.88 & 33.81 & 33.40 & 40.48 & 2.47 & 4.63 & 38.81 & 9.84 & 29.16 \\
 & SkyEye  & \cmark & \cmark & & 73.10 & \textbf{39.23} & 38.08 & \textbf{45.72} & \textbf{4.05} & \textbf{10.44} & \textbf{44.72} & 12.10 & \textbf{33.43} \\
 & \net & \xmark & \xmark & & 73.89 & 38.42 & 42.25 & 41.46 & 2.26 & 6.26 & 37.20 & 15.08 & 32.10 \\
 & \net & \xmark & \cmark & & \textbf{74.29} & 38.48 & \textbf{43.87} & 42.77 & 2.80 & 5.22 & 37.68 & \textbf{15.20} & 32.54 \\
\midrule
 \multirow{4}{*}{100\%} & PoBEV & \xmark & - & \multirow{4}{*}{20} & 70.14 & 35.23 & 34.68 & 40.72 & 2.85 & 5.63 & 39.77 & 14.38 & 30.42 \\
 & SkyEye & \cmark & \cmark & & 73.57 & \textbf{39.45} & 38.74 & \textbf{46.06} & \textbf{3.95} & \textbf{9.66} & \textbf{45.21} & 10.92 & \textbf{33.44} \\
 & \net & \xmark & \xmark & & 74.22 & 39.39 & \textbf{42.86} & 42.96 & 2.55 & 6.66 & 35.68 & \textbf{17.11} & 32.68 \\
 & \net & \xmark & \cmark & & \textbf{74.81} & 38.59 & 42.58 & 43.67 & 3.52 & 6.21 & 38.47 & 15.24 & 32.88 \\
 \bottomrule
 \end{tabular}
\label{tab:ablation-percentages}
\vspace{-0.3cm}
\end{table}

In this section, we investigate the influence of various components of our approach by performing an ablation study on the KITTI-360 dataset. Specifically, we evaluate the impact of model pretraining when presented with varying amounts of labeled BEV data, the benefit of each of our neural pathways, and the effect of varying masking ratios on the overall performance of the network.

{\parskip=5pt
\noindent\textbf{Impact of Model Pretraining}: In this section, we study the impact of model pretraining by finetuning our model \textit{with} and \textit{without} pretraining with varying percentages of labeled BEV data. Accordingly, we establish five percentage splits of BEV labels, i.e., $1\%$, $5\%$, $10\%$, $50\%$, and $100\%$, and sample three random sets for each percentage split. We train each percentage split three times, once using each random set, and report the mean value to mitigate the risk of random chance affecting the final results. Moreover, we also train the best two baselines, i.e., PoBEV and SkyEye, across all percentage splits as a reference for evaluating our approach. \tabref{tab:ablation-percentages} presents the results of this ablation study.}

We observe that our model trained using our unsupervised pretraining strategy, \net, consistently outperforms our model without pretraining across all percentage splits. The most substantial improvements of \SI{2.49}{pp} and \SI{1.41}{pp} occur when finetuning with only $1\%$ and $10\%$ of BEV labels, respectively. We also note that \net~outperforms PoBEV by \SI{9.87}{pp} and SkyEye by \SI{2.02}{pp} when using only $1\%$ of BEV labels. At extremely low percentage splits, PoBEV does not encounter enough BEV labels to learn the mapping from FV to BEV, while the FV semantic-based pretraining of SkyEye does not impart sufficient geometric modeling and representation learning ability to the network. The notable improvement over SkyEye is primarily attributed to the superior segmentation performance on static classes such as \textit{road} and \textit{building} as well as non-moving dynamic objects such as trucks and buses. This improvement directly stems from the use of implicit neural fields to model scene geometry which helps the network to effectively reason about static elements in the scene. Moreover, we highlight that \net~finetuned using only $5\%$ of BEV labels already outperforms the state-of-the-art fully supervised approach PoBEV trained using $100\%$ of BEV labels; thus underscoring the impact of model pretraining in reducing the dependence on large quantities of labeled data. We also note that SkyEye consistently outperforms our approach across four of the five percentage splits on \textit{person}, \textit{two-wheeler}, and \textit{car}. We believe that the superior performance of SkyEye stems from the presence of $100\%$ FV labels which provide unparalleled semantic knowledge during the pretraining phase. Nevertheless, our approach still yields competitive results without using any additional labeled data, thus highlighting the impact of our unsupervised pretraining mechanism.

\begin{table}[t]
 \caption{Ablation study to investigate the efficacy of various network components. All experiments are on the KITTI-360 dataset using 1\% of BEV labels.}
 \vspace{-0.3cm}
\centering
\scriptsize
 \begin{tabular}{c|cc|cccccccc|c}
 \toprule
 \textbf{Model} & \textbf{Geometric} & \textbf{Semantic} & \textbf{Road} & \textbf{Side.} & \textbf{Build.} & \textbf{Terr.} & \textbf{Pers.} & \textbf{2-Wh.} & \textbf{Car} & \textbf{Truck} & \textbf{mIoU} \\
 \midrule
  L1 & \xmark & \xmark & 69.40 & 32.09 & 34.75 & 35.27 & 1.01 & 2.79 & 28.76 & 7.66 & 26.47 \\
  L2 & \cmark & \xmark & 70.85 & \textbf{34.34} & 38.12 & 35.03 & 0.93 & 4.06 & 29.79 & 8.84 & 27.75 \\ 
  L3 & \cmark & \cmark & 70.58 & 34.26 & \textbf{40.68} & \textbf{38.53} & \textbf{1.35} & \textbf{4.74} & \textbf{30.94} & \textbf{10.58} & \textbf{28.96}  \\
 \bottomrule
 \end{tabular}
\label{tab:ablation-network-components}
 \vspace{-0.3cm}
\end{table}

{\parskip=5pt
\noindent\textbf{Influence of Network Components}: In this section, we quantify the impact of the geometric and semantic pathways on the overall performance of the network by incrementally incorporating each component into the pretraining step and finetuning the resultant model on $1\%$ of BEV labels. \tabref{tab:ablation-network-components} presents the results of this ablation study. 
The first row, comprising model L1, illustrates a network without our unsupervised pretraining and serves as a baseline to assess the improvement brought about by the other components. Model L2 incorporates the geometric pathway into the pretraining which results in an improvement of \SI{1.28}{pp} over model L1. The inclusion of geometric pathway during pretraining allows the implicit field to learn the scene geometry and reason about occlusions which helps improve the IoU metric on most of the classes by nearly \SI{1}{pp}. 
Upon the incorporation of our novel temporal MAE strategy via the semantic pathway in model L3, we observe a significant \SI{1.21}{pp} improvement over model L2. By learning to reconstruct the missing information in the masked patches over multiple timesteps, the network learns spatially- and temporally consistent representations of various scene components which allows it to easily map the learned representation to the semantic BEV task using only $1\%$ of BEV labels. 
}

{\parskip=5pt
\noindent\textbf{Impact of Mask Ratios}: In this experiment, we evaluate the impact of different masking ratios on the overall performance of the model and present the results in \tabref{tab:ablation-maskratios}. We observe that a masking ratio of $75\%$ is ideal for our novel temporal masked autoencoding mechanism. Lower masking ratios do not present a sufficiently challenging pretraining task and thus result in only marginal improvements over model L2 in \tabref{tab:ablation-network-components}, while higher masking ratios mask out a significant portion of vital information resulting in worse performance as compared to a model with no masked autoencoding.
}

\begin{table}[t]
\caption{Ablation study on the impact of masking ratio. All experiments are on the KITTI-360 dataset using 1\% of BEV labels.}
\vspace{-0.3cm}
\centering
\scriptsize
 \begin{tabular}{c|ccccc}
 \toprule
 \textbf{Masking Ratio} & \textbf{0\%} & \textbf{25\%} & \textbf{50\%} & \textbf{75\%} & \textbf{90\%} \\
 \midrule
 \textbf{mIoU} & 27.75 & 27.87 & 28.22 & \textbf{28.96} & 27.31 \\
 \bottomrule
 \end{tabular}
\label{tab:ablation-maskratios}
\vspace{-0.3cm}
\end{table}

\begin{figure}[!t]
\centering
\footnotesize
\setlength{\tabcolsep}{0.05cm}
{
\renewcommand{\arraystretch}{0.15}
\newcolumntype{M}[1]{>{\centering\arraybackslash}m{#1}}
\begin{tabular}{cM{4.5cm}M{2.25cm}M{2.25cm}M{2.25cm}}
& \scriptsize{Input FV Image} & \scriptsize{SkyEye/PoBEV} & \scriptsize{\net~(Ours)} & \scriptsize{Improvement/Error} \\ 
\\
\rotatebox[origin=c]{90}{\scriptsize{(a) KITTI-360}} & {\includegraphics[width=\linewidth, height=0.455\linewidth, frame]{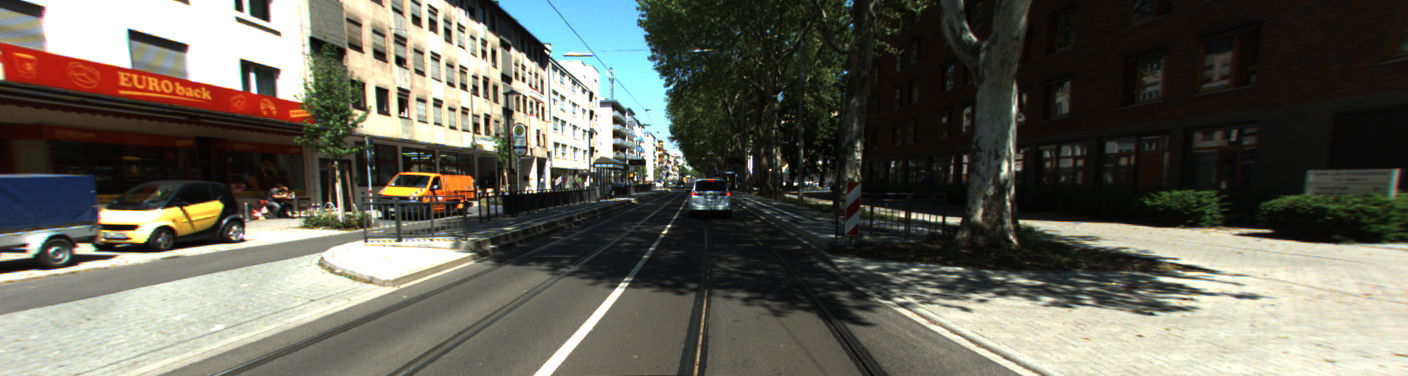}} & {\includegraphics[width=\linewidth, frame]{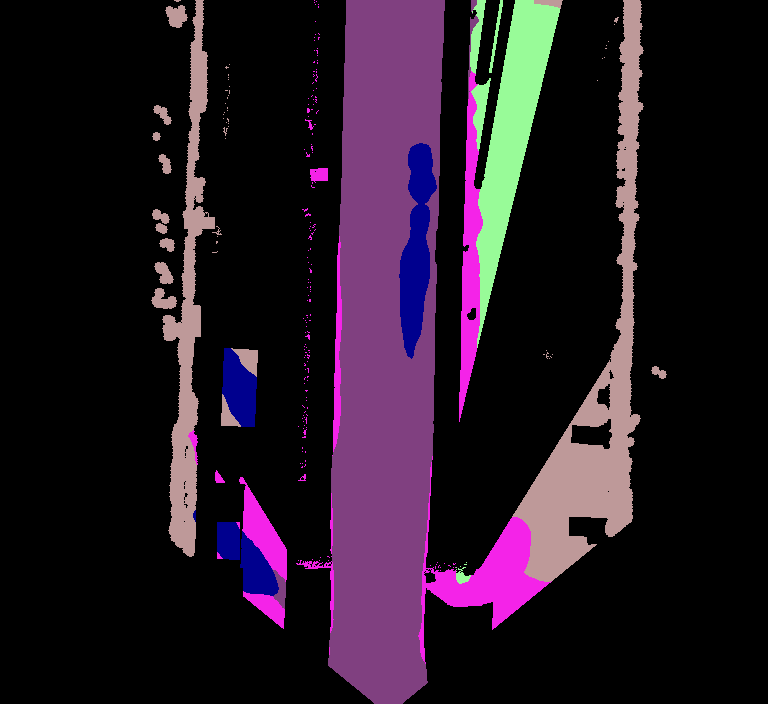}} & {\includegraphics[width=\linewidth, frame]{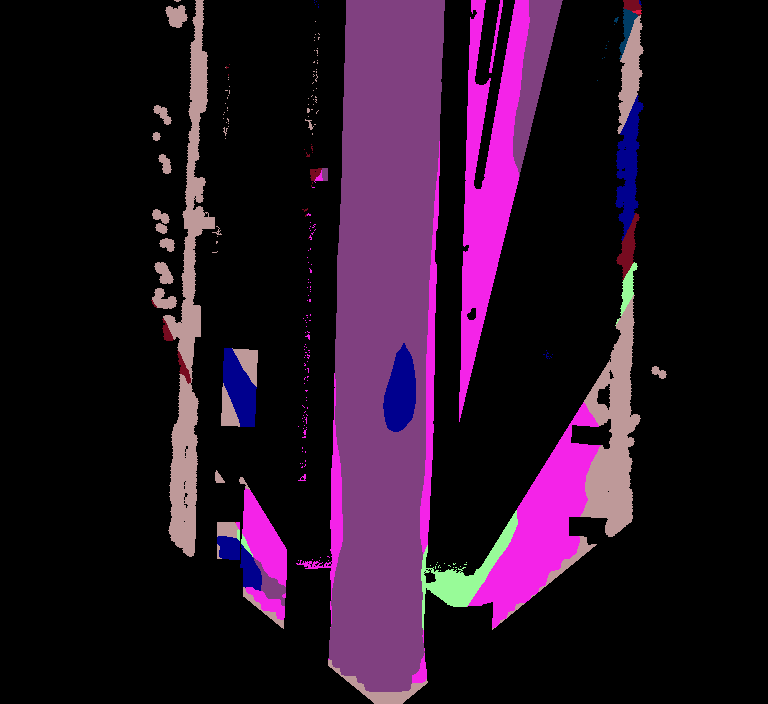}} & {\includegraphics[width=\linewidth, frame]{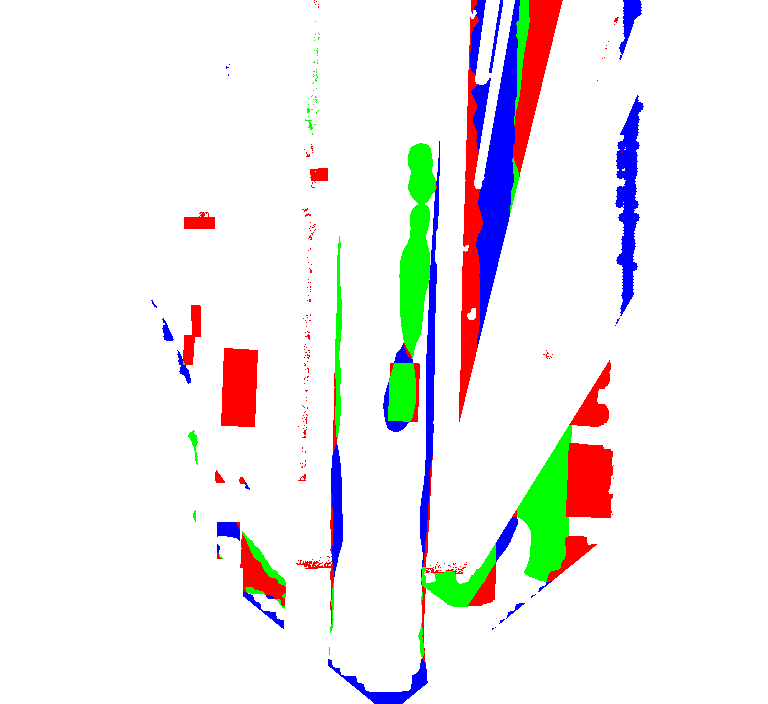}} \\
\\
\rotatebox[origin=c]{90}{\scriptsize{(b) KITTI-360}} & \includegraphics[width=\linewidth, height=0.455\linewidth, frame]{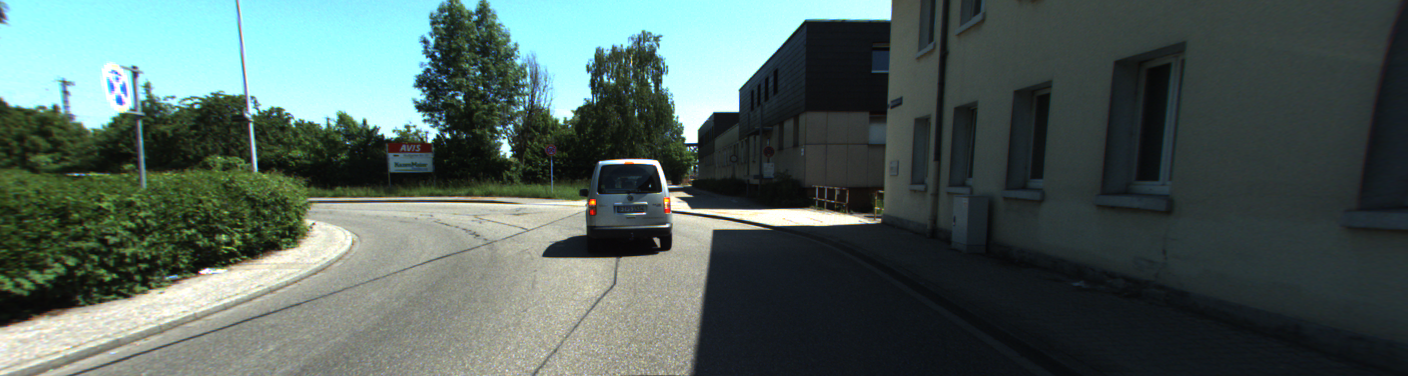} & \includegraphics[width=\linewidth, frame]{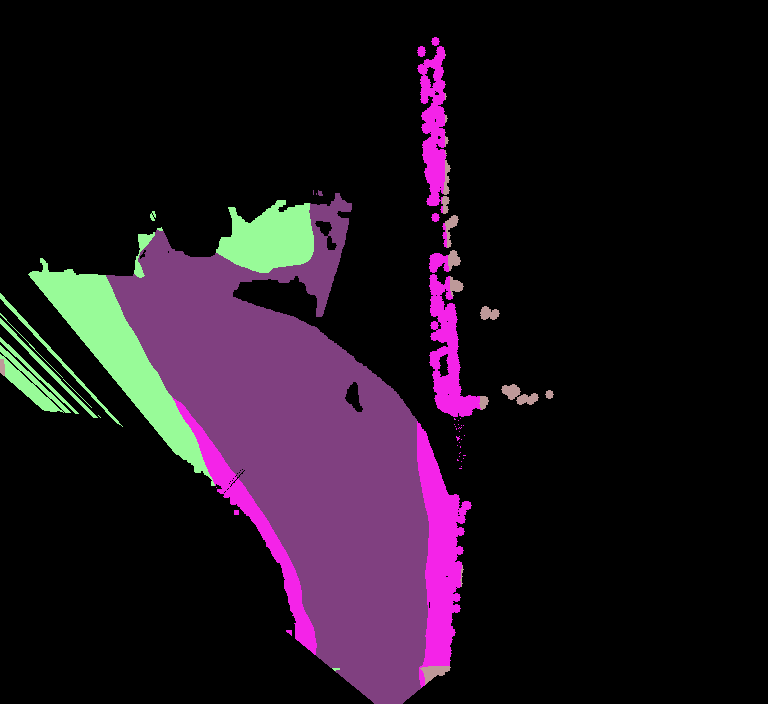} & \includegraphics[width=\linewidth, frame]{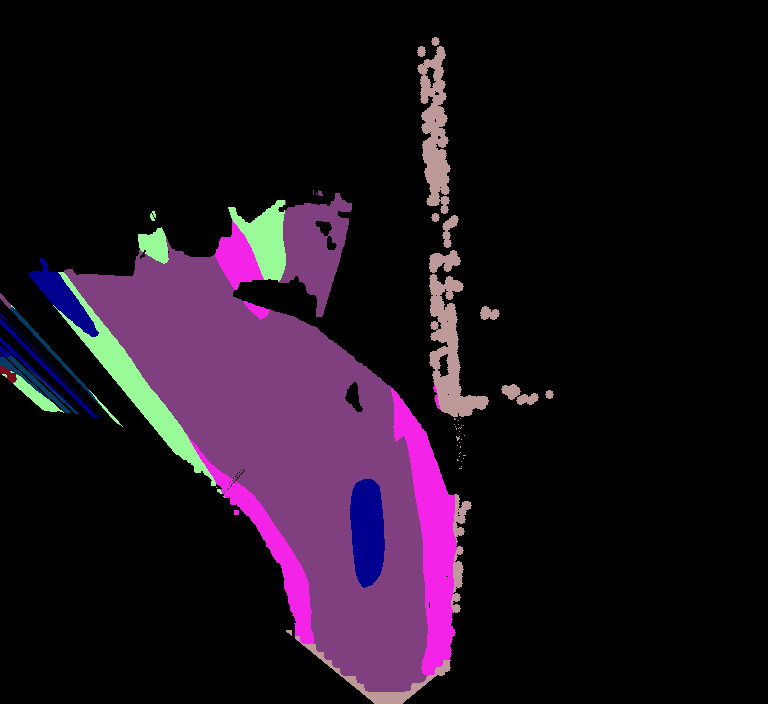} & \includegraphics[width=\linewidth, frame]{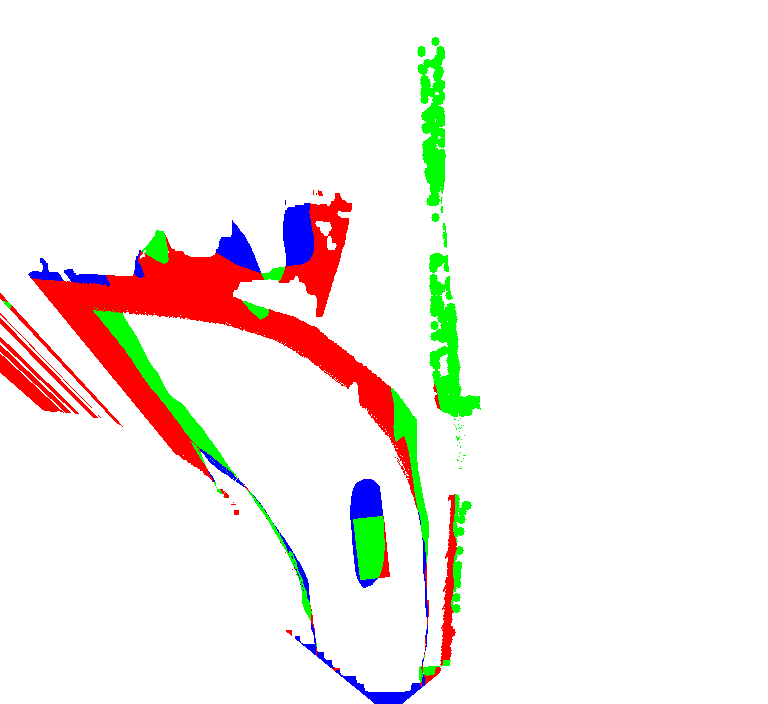} \\
\\
\rotatebox[origin=c]{90}{\scriptsize{(c) KITTI-360}} & \includegraphics[width=\linewidth, height=0.455\linewidth, frame]{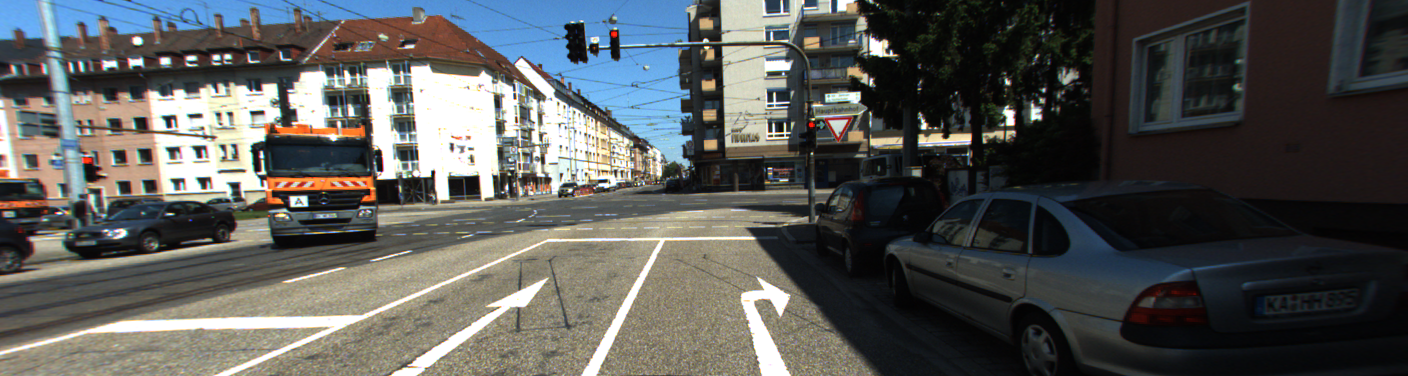} & \includegraphics[width=\linewidth, frame]{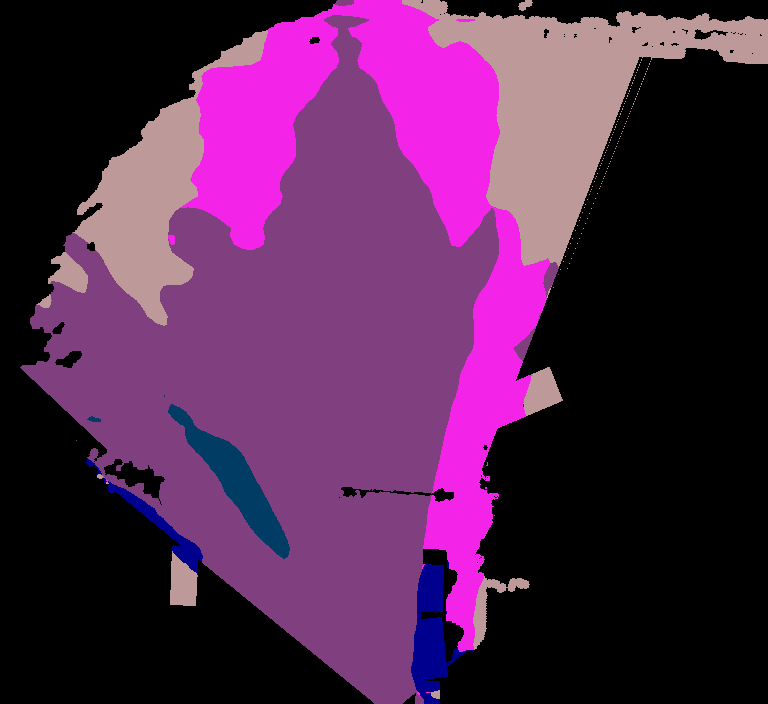} & \includegraphics[width=\linewidth, frame]{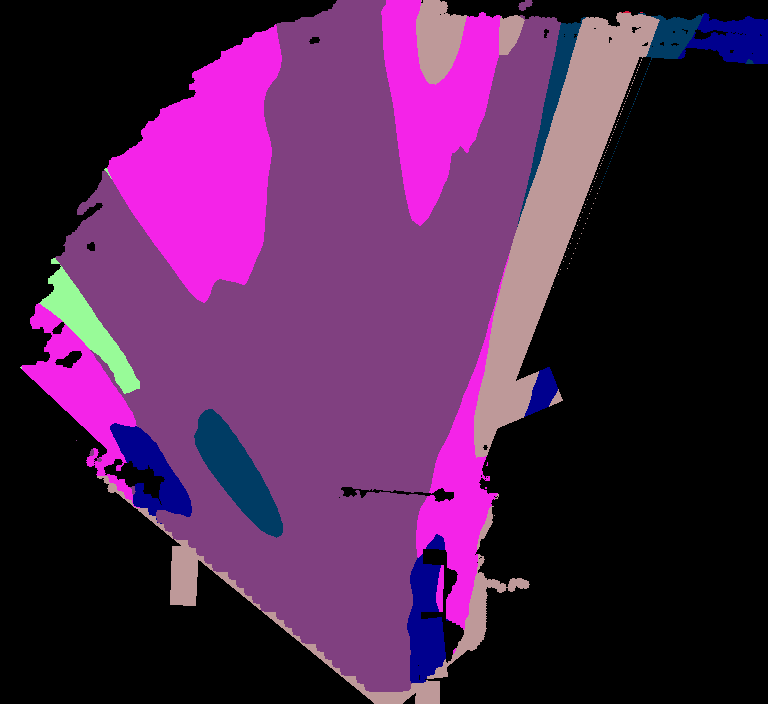} & \includegraphics[width=\linewidth, frame]{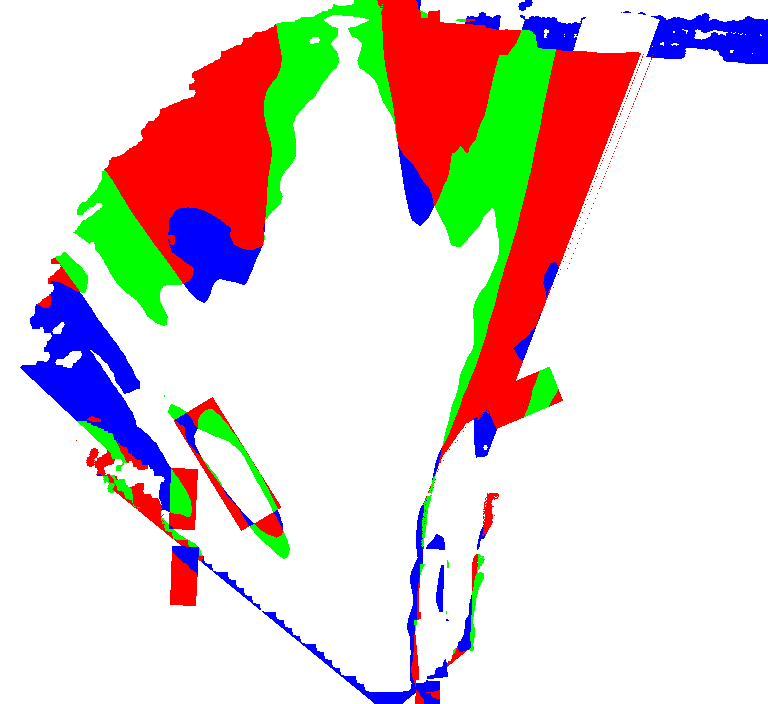} \\
\\
\rotatebox[origin=c]{90}{\scriptsize{(d) nuScenes}} & \includegraphics[width=\linewidth, height=0.43\linewidth, frame]{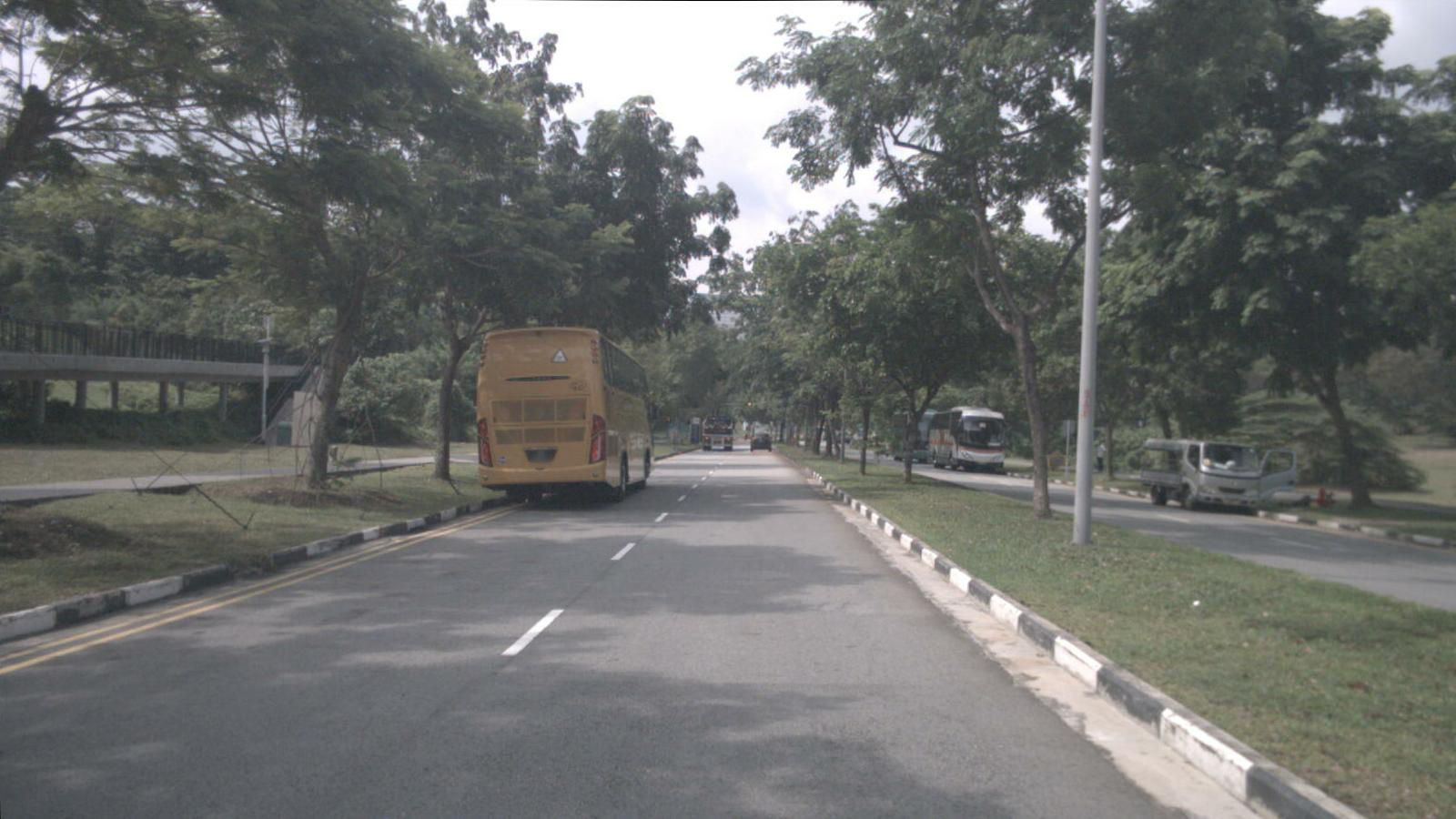} & \includegraphics[width=\linewidth, frame]{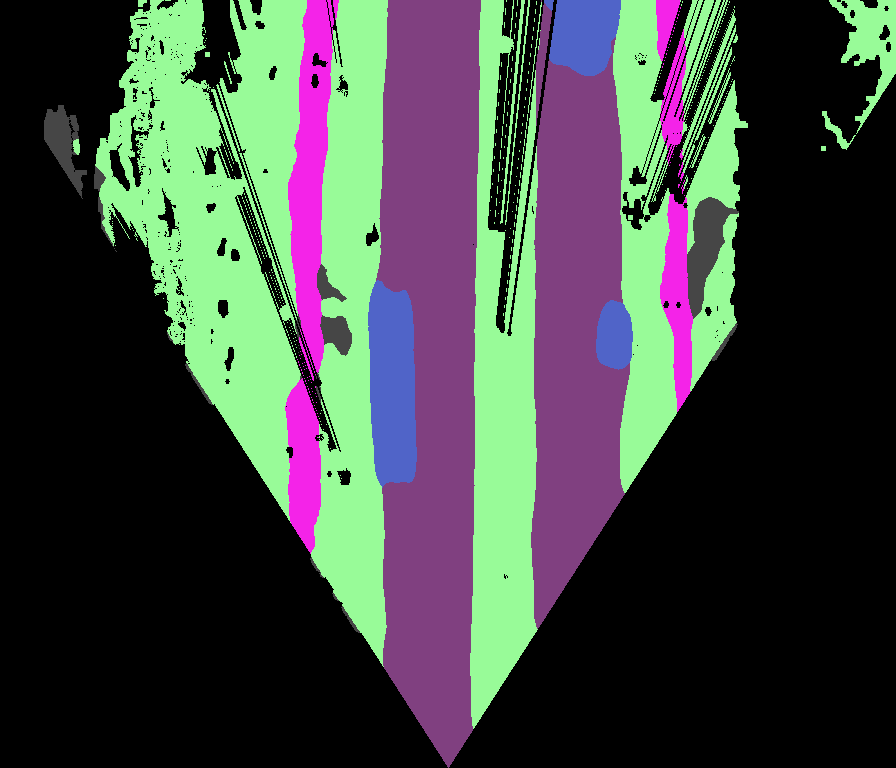} & \includegraphics[width=\linewidth, frame]{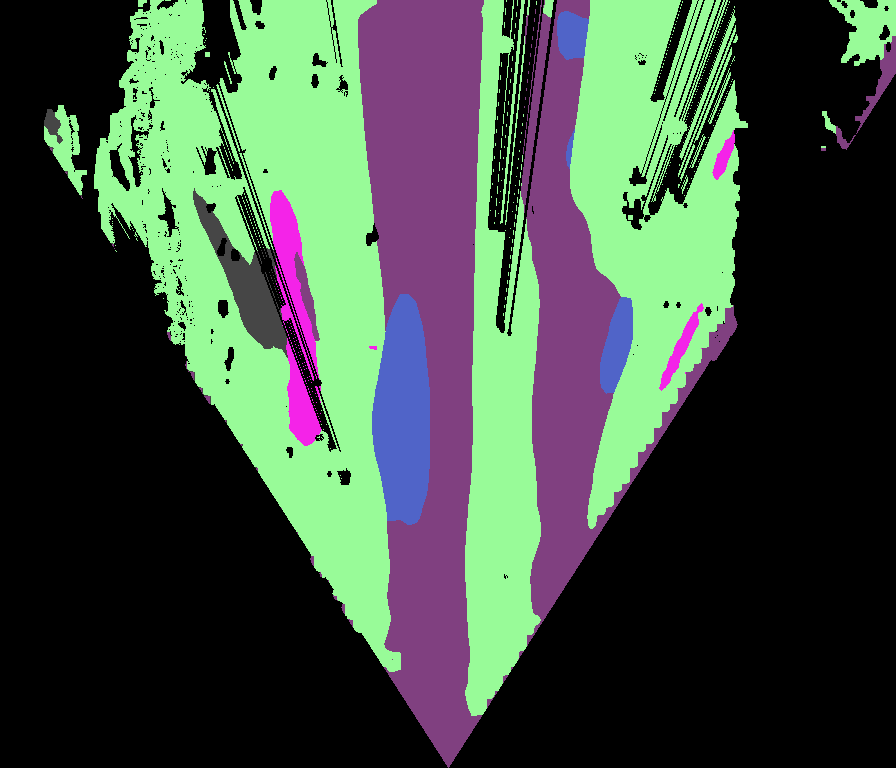} & \includegraphics[width=\linewidth, frame]{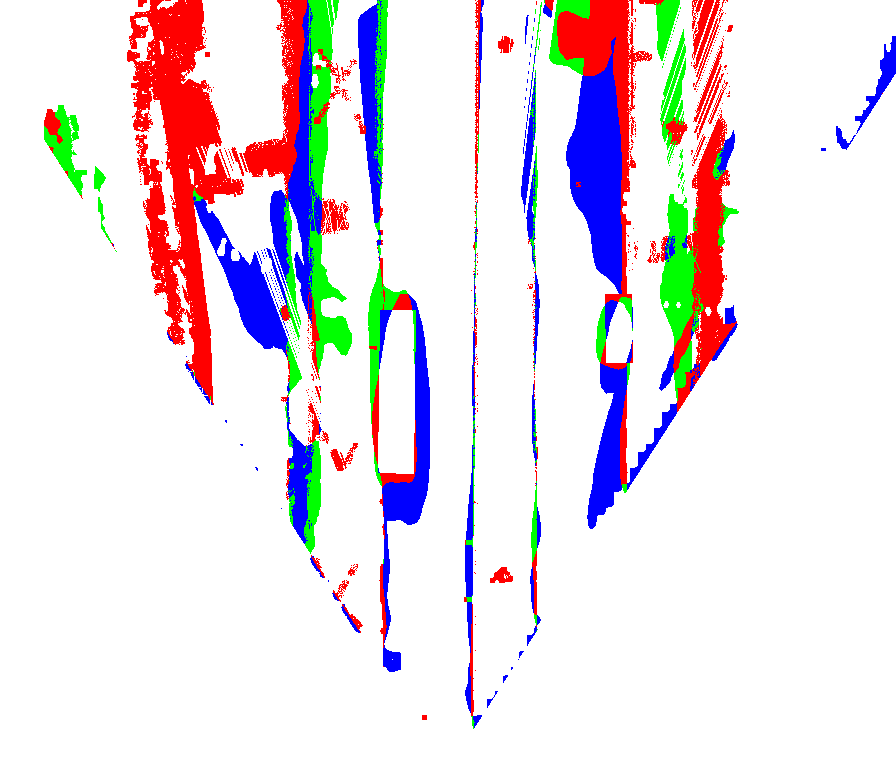} \\
\\
\rotatebox[origin=c]{90}{\scriptsize{(e) nuScenes}} & \includegraphics[width=\linewidth, height=0.43\linewidth, frame]{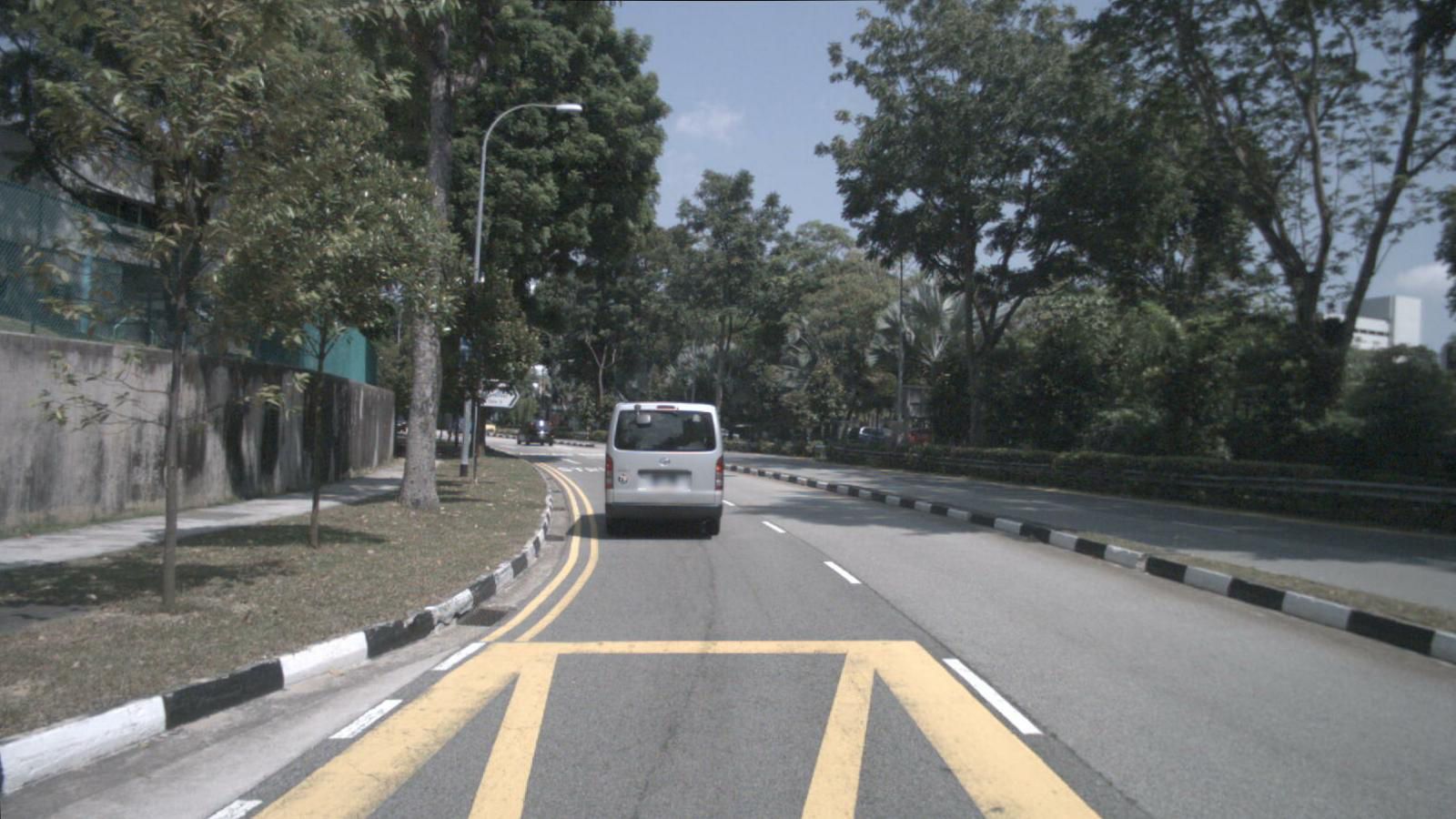} & \includegraphics[width=\linewidth, frame]{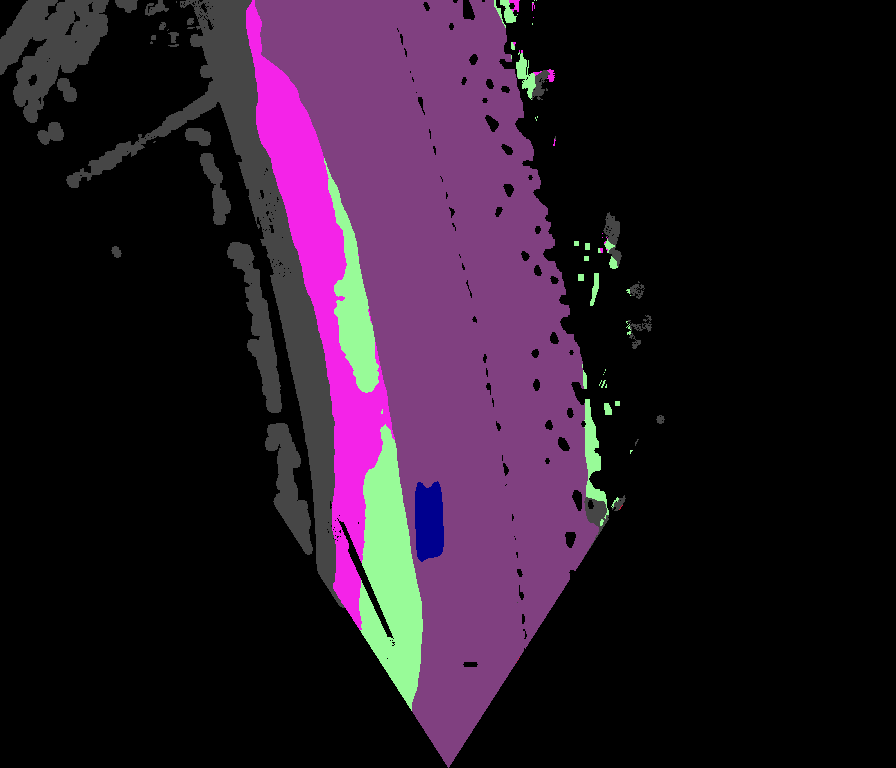} & \includegraphics[width=\linewidth, frame]{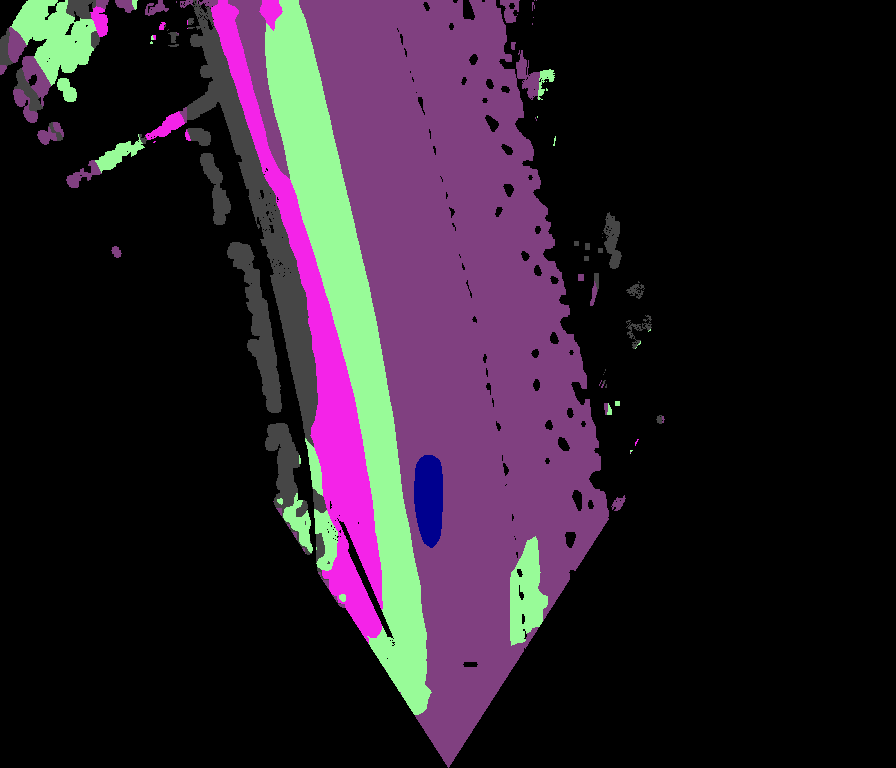} & \includegraphics[width=\linewidth, frame]{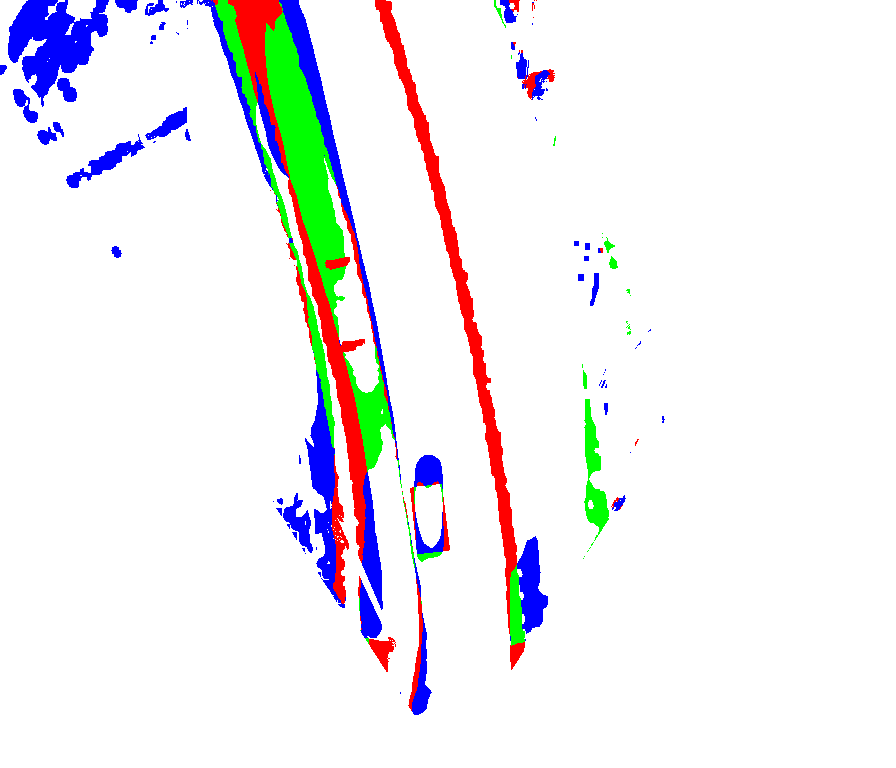} \\
\\
\rotatebox[origin=c]{90}{\scriptsize{(f) nuScenes}} & \includegraphics[width=\linewidth, height=0.43\linewidth, frame]{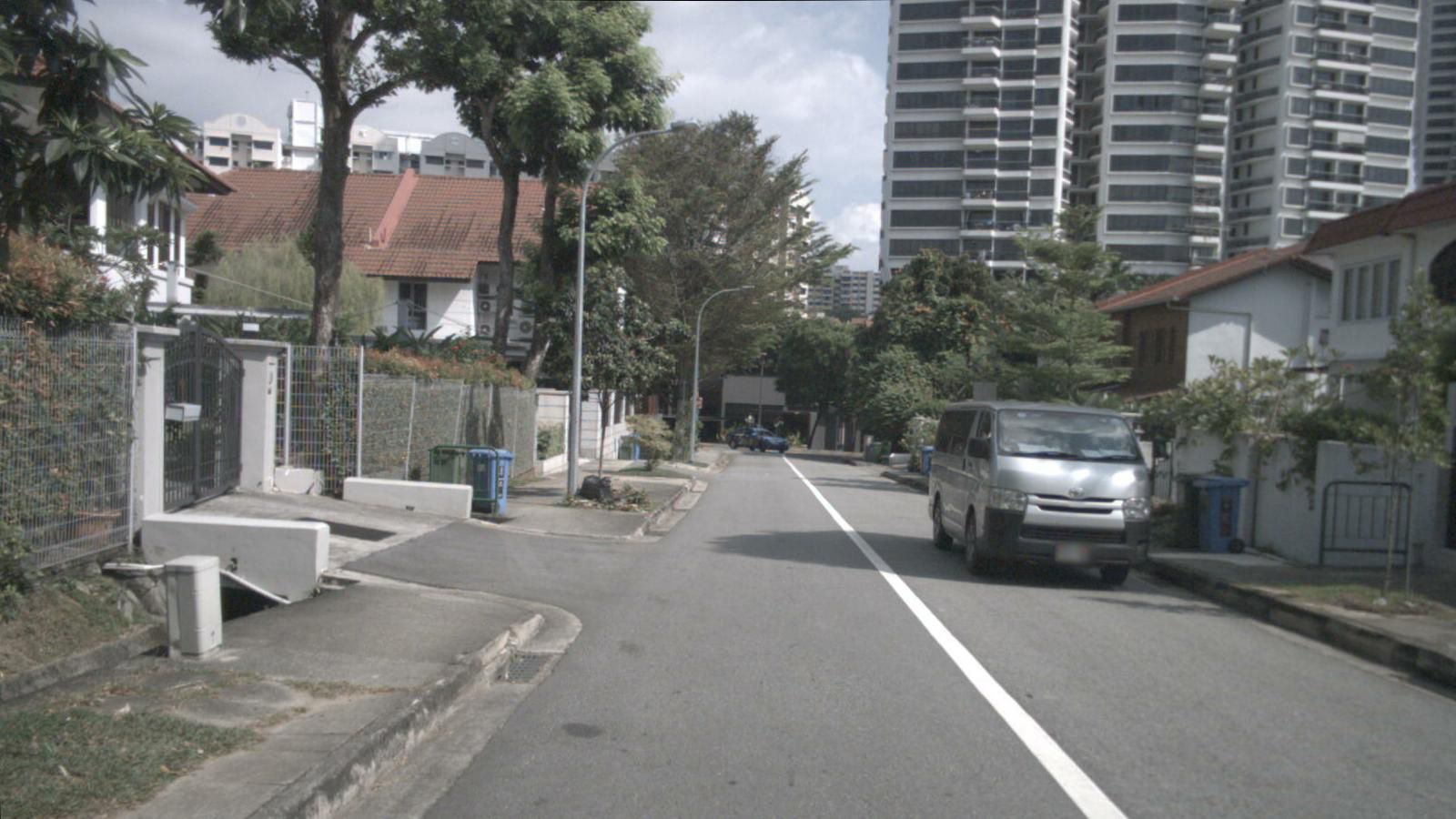} & \includegraphics[width=\linewidth, frame]{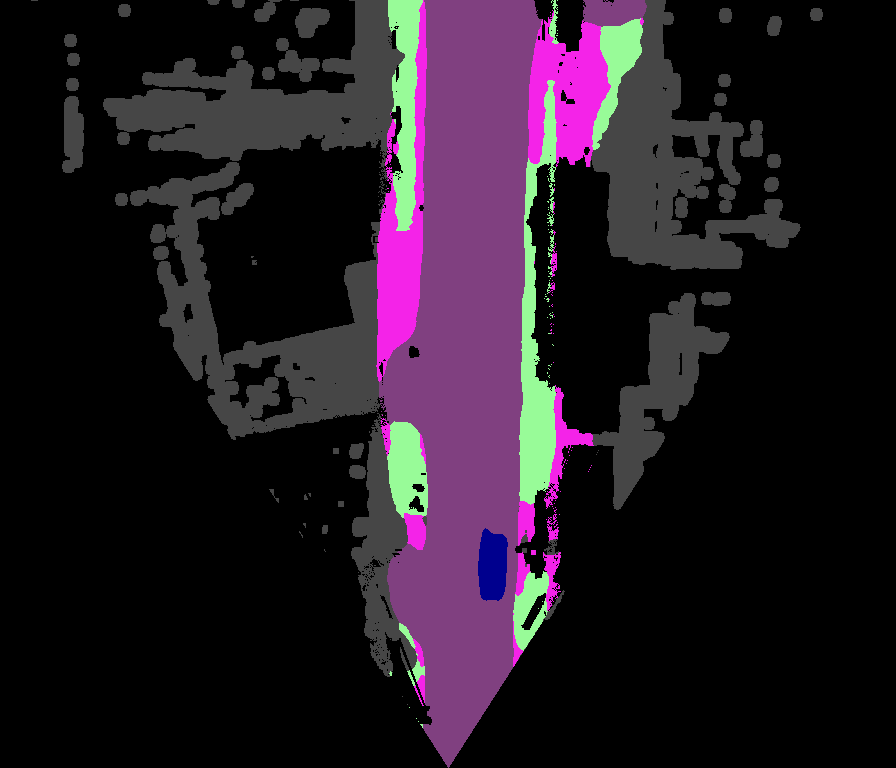} & \includegraphics[width=\linewidth, frame]{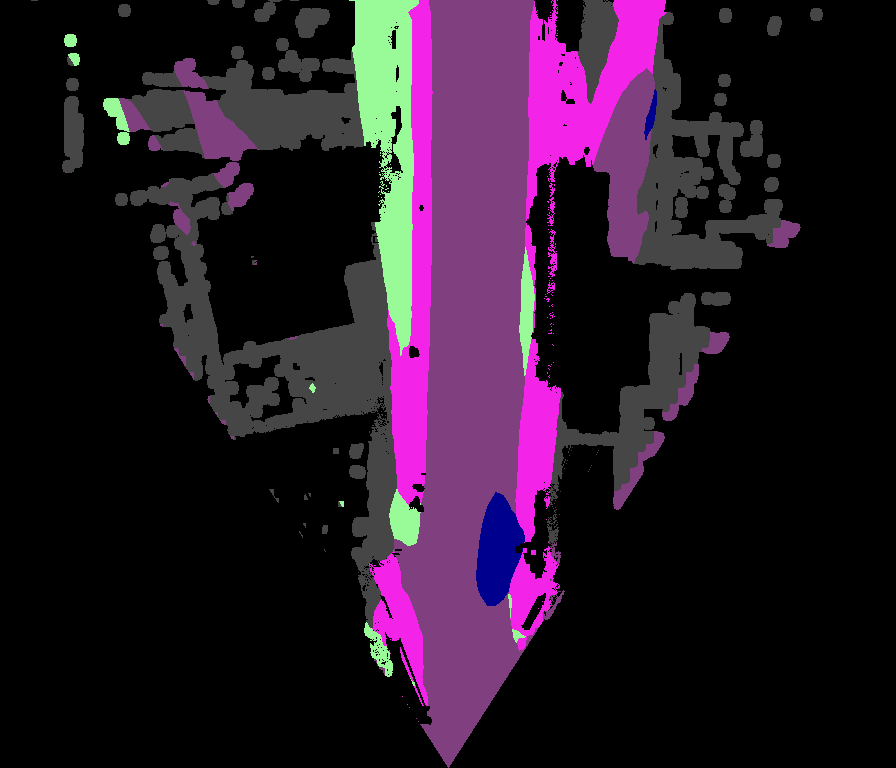} & \includegraphics[width=\linewidth, frame]{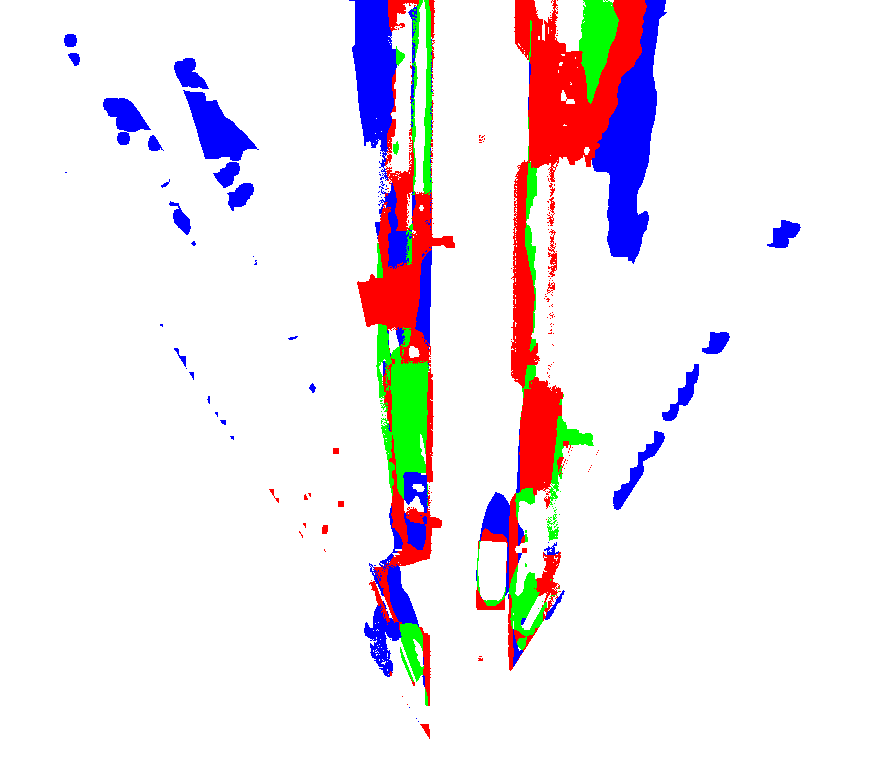} \\
\end{tabular}
}
\caption{Qualitative results of our unsupervised learning framework \net~in comparison with SkyEye~\cite{cit:bev-seg-skyeye} on the KITTI-360 dataset, and PoBEV~\cite{cit:bev-seg-panopticbev} on the nuScenes dataset. We also display an Improvement/Error map which highlights pixels where \net~outperforms the baseline in green, where baseline outperforms \net~in blue, and where both models misclassify in red.}
\label{fig:qual-analysis}
\vspace{-0.5cm}
\end{figure}

\subsection{Qualitative Results}

We qualitatively evaluate the performance of \net~in \figref{fig:qual-analysis}~by comparing it with SkyEye~\cite{cit:bev-seg-skyeye} on the KITTI-360 dataset, and PoBEV~\cite{cit:bev-seg-panopticbev} on the nuScenes dataset.
We observe from \figref{fig:qual-analysis}(a) that both SkyEye and \net~are able to predict static classes such as road and sidewalk to a high degree of accuracy, but SkyEye fails to properly localize the car and significantly stretches it along the depth dimension. Our approach, on the other hand, can both properly localize the car in the BEV map as well as predict its extent. 
In \figref{fig:qual-analysis}(b) we observe that SkyEye fails to detect the car in the scene, while our approach not only detects the object in the scene but also accurately estimates its extent. Further, we observe in \figref{fig:qual-analysis}(c) that our approach is able to better predict the extent of the truck as well as predict the location of the \textit{road} class in far-off regions. 
We hypothesize that our approach efficiently leverages the rich geometric and semantic knowledge learned by the disjoint neural pathways to effectively transfer the knowledge from FV to BEV even when using only $1\%$ of BEV labels.
We observe from \figref{fig:qual-analysis}(d-f) that our model accurately estimates the static elements of the scene which is evident from the error/improvement map in the last column. Although trained on only $1\%$ of BEV labels as compared to $100\%$ for PoBEV, our approach manages to precisely capture the locations of \textit{car} and \textit{truck} instances in the BEV map. 
\figref{fig:qual-analysis}(d) also highlights one of the limitations of our approach wherein dynamic objects such as cars, trucks, and pedestrians are often radially stretched. This limitation is primarily caused by the lack of sufficient camera views to learn the entire 3D representation of the dynamic objects which could be addressed by exploiting the cross-view correlation and spatio-temporal consistency of surround view cameras. Interestingly, \figref{fig:qual-analysis}(e) reveals that \net~is also able to leverage representative scene priors learned during the pretraining step to infer knowledge about occluded regions. In this prediction, \net~is able to predict that the \textit{terrain} class extends further into the occluded region, unlike PoBEV which incorrectly predicts this region as \textit{sidewalk}; thus highlighting the benefit of our unsupervised pretraining protocol. 

\subsection{Discussion of Limitations}
\net~suffers from three main limitations, all of which pertain to modeling scene geometry using neural implicit fields. Firstly, existing implicit field formulations enforce a strong static scene assumption which is often violated in real-world autonomous driving environments. An explicit dynamic object handling module as discussed in~\cite{ingale2021real} could be used to address this limitation in such environments. Secondly, implicit fields rely on a large and diverse set of camera views of a given object to learn its optimal scene geometry. However, this is infeasible in autonomous driving which results in the generation of sub-optimal volumetric grids in their current form. Lastly, our formulation of implicit fields is supervised using the photometric loss between temporal multi-camera images. However, the photometric loss is often sensitive to varying lighting conditions, occlusions, and disocclusions, as well as object motion - all of which are exacerbated when the ego motion between two frames is large.
This problem can typically be addressed by adding a stereo camera to capture slightly offset images and provide a reliable frame for loss computation.\looseness=-1

\section{Conclusion}
\label{sec:conclusion}

In this paper, we present the first unsupervised representation learning approach, \net, for predicting semantic BEV maps from monocular FV images using a label-efficient learning paradigm. Our approach leverages the spatio-temporal consistency and rich scene semantics offered by FV image sequences to independently learn the sub-tasks of BEV mapping, i.e., scene geometry estimation and scene representation learning, in an unsupervised pretraining step. It finetunes the resultant model on the BEV segmentation task using only a small fraction of labels in BEV. Using extensive evaluations on the KITTI-360 and nuScenes datasets, we demonstrate that \net~performs on par with the existing fully-supervised and self-supervised approaches while using only 1\% of BEV labels and without relying on any additional source of labeled supervision.

{
    \bibliographystyle{splncs04}
    \bibliography{references}
}

\clearpage
\setcounter{page}{1}

\begin{center}
\Large
\textbf{\thetitle}\\
\vspace{0.5em}Supplementary Material \\
\vspace{1.0em}
\end{center}

\setcounter{section}{0}
\setcounter{equation}{0}
\setcounter{figure}{0}
\setcounter{table}{0}
\makeatletter

\renewcommand{\thesection}{S.\arabic{section}}
\renewcommand{\thesubsection}{S.\arabic{section}.\arabic{subsection}}
\renewcommand{\thetable}{S.\arabic{table}}
\renewcommand{\thefigure}{S.\arabic{figure}}

\noindent In this supplementary material, we present additional experimental results to analyze the performance of \net, our unsupervised representation learning framework for semantic BEV mapping. To this end, 
we present further ablation experiments in \secref{sec:supp-ablation-experiments} and additional qualitative results in \secref{sec:supp-qualitative-results}.

\section{Additional Ablative Experiments}
\label{sec:supp-ablation-experiments}
In this section, we present additional ablative experiments to further study the impact of various parameters on the overall performance of the model. Specifically, we study the influence of (1) different DINOv2 variants (\secref{subsec:supp-ablation-dino-variants}) 
and (2) masking patch size in our novel T-MAE module (\secref{subsec:supp-ablation-patchsize}) on the overall performance of the model. Further, we also present the results obtained when the native backbones of the best baselines are replaced with our backbone (\secref{subsec:supp-baseline-dinov2}) and when the BEV percentage split defined in SkyEye~\cite{cit:bev-seg-skyeye} is used for model finetuning (\secref{subsec:supp-finetuning-skyeye-split}).

\subsection{DINOv2 Backbone Variants}
\label{subsec:supp-ablation-dino-variants}

In this section, we study the influence of different variants of the DINOv2 backbone on the overall performance of the model. Specifically, we first pretrain the model using four variants, namely, \textit{vit-b}, \textit{vit-s}, \textit{vit-l}, and \textit{vit-g}, and finetune each of them using $1\%$ of semantic BEV labels. 
\tabref{tab:ablation-dino-backbones} presents the results of this ablation study. We observe that \textit{vit-s} yields the lowest performance among all variants, achieving \SI{3.41}{pp} lower than \textit{vit-b}. Being the smallest of all variants, \textit{vit-s} does not generate features that are as representative as its larger counterparts, thus resulting in its reduced overall performance.

{\parskip=3pt
\noindent The three larger variants, i.e., \textit{vit-b}, \textit{vit-l}, and \textit{vit-g}, however, yield very similar mIoU scores with the difference between the highest and lowest performance being only \SI{0.80}{pp}. In other words, the performance of semantic BEV mapping saturates after \textit{vit-b} and does not improve upon using a larger backbone. We infer that this behavior can be attributed to one of the following reasons: (1) larger DINOv2 backbones provide better features for FV tasks, but these features do not easily transfer to the task of BEV mapping, or (2) $1\%$ of BEV labels are insufficient to leverage the full potential of larger backbones. Given the similar performance of \textit{vit-b} as compared to \textit{vit-l} and \textit{vit-g} while being more efficient in terms of number of parameters, we use the \textit{vit-b} variant of the DINOv2 backbone in this work. 
}

\begin{table}[t]
\caption{Ablation study on the impact of different DINOv2 backbones on the overall model performance. All models in this experiment are finetuned using only $1\%$ of BEV labels. All metrics are reported in $[\%]$ on the KITTI-360 dataset.}
\vspace{-0.3cm}
\scriptsize
\centering
 \begin{tabular}{c|cccc}
 \toprule
 \textbf{Backbone} & \textbf{vit-s} & \textbf{vit-b} & \textbf{vit-l} & \textbf{vit-g} \\
 \midrule
 \textbf{mIoU} & 25.55 & \textbf{28.96} & 28.40 & 28.16 \\
\bottomrule
\end{tabular}
\vspace{-0.3cm}
\label{tab:ablation-dino-backbones}
\end{table}

\begin{table}[t]
\caption{Ablation study on the impact of masking patch size in T-MAE on the overall performance of the model. All models are finetuned using only $1\%$ of labels in BEV. All metrics are reported in $[\%]$ on the KITTI-360 dataset.}
\vspace{-0.3cm}
\scriptsize
\centering
 \begin{tabular}{c|cccccccc|c}
 \toprule
 \textbf{Patch Size} & \textbf{Road} & \textbf{Side.} & \textbf{Build.} & \textbf{Terr.} & \textbf{Per.} & \textbf{2-Wh.} & \textbf{Car} & \textbf{Truck} & \textbf{mIoU}
 \\
 \midrule
 14 & \textbf{71.45} & 33.41 & 36.89 & 37.48 & 0.75 & 3.69 & 30.05 & 9.23 & 27.87 \\
 28 & 70.58 & \textbf{34.26} & \textbf{40.68} & \textbf{38.53} & 1.35 & \textbf{4.74} & \textbf{30.94} & \textbf{10.58} & \textbf{28.96} \\
 56 & 70.02 & 34.10 & 38.87 & 37.88 & \textbf{1.37} & 4.71 & 30.91 & 9.66 & 28.44 \\
\bottomrule
\end{tabular}
\vspace{-0.3cm}
\label{tab:ablation-mask-patch}
\end{table}

\subsection{Masking Patch Size}
\label{subsec:supp-ablation-patchsize}
In this section, we analyze the influence of masking patch sizes used for masking the input image in our novel temporal MAE (T-MAE) module on the overall performance of the model. To this end, we first pretrain the model using masking patches of size $14$, $28$, and $56$, and then finetune the resultant model on $1\%$ of BEV labels. \tabref{tab:ablation-mask-patch} presents the results of this ablation study.

{\parskip=3pt
\noindent We observe that a masking patch size of $28$ gives the highest mIoU score across all the evaluated patch sizes. A smaller patch size does not mask out enough of an object and consequently does not present a challenging reconstruction task during the unsupervised pretraining phase. In contrast, a larger patch size masks out significant distinguishing regions in the image which hinders the representation learning ability of the network during the pretraining phase. The effect of patch sizes is noticeable across all classes while being significant for dynamic objects which experience a substantial reduction in the IoU scores when too little of the object is masked out. Given these observations, we use a patch size of $28$ in our \net~framework. 
}

\subsection{Impact of DINOv2 on Baseline Approaches}
\label{subsec:supp-baseline-dinov2}
In this section, we analyze the impact of the DINOv2 backbone on the overall performance of the baseline models. Specifically, we replace the native backbones of the two best baselines, PanopticBEV~\cite{cit:bev-seg-panopticbev} and SkyEye~\cite{cit:bev-seg-skyeye}, with a pretrained DINOv2 backbone as used in our model. We follow the setting defined in \secref{subsec:ablation-study} of the main paper and report the results when finetuning with varying percentages of BEV labels in \tabref{tab:supp-baseline-dinov2}. We observe that PoBEV reports slightly better performance across all percentage splits when using the DINOv2 backbone with the highest improvement of \SI{1.41}{pp} observed when using $100\%$ of BEV labels. In contrast, we observe that the performance of SkyEye deteriorates when the native encoder is replaced with the DINOv2 backbone. At lower percentage splits of $1\%$, $5\%$, and $10\%$, the BEV segmentation performance drops by \SI{2.85}{pp}, \SI{3.76}{pp}, and \SI{2.62}{pp} which indicates that the SkyEye framework is unable to adapt the DINOv2 features to this task. The performance drop is observed across all classes and is especially large for \textit{car} and \textit{two-wheeler} which we believe is a consequence of not having an explicit scene geometry estimation module to estimate the extent of objects in the scene. We infer that the native backbone of SkyEye absorbs a significant chunk of scene geometry, but when replaced with a frozen backbone as in our model, SkyEye fails to learn sufficient geometric information. We thus conclude that using our backbone in the baseline approaches results in only a slight improvement in PoBEV and deteriorates the BEV segmentation performance in SkyEye.

\begin{table}[th]
\caption{Performance of baseline approaches when using the DINOv2 backbone as used in \net. All experiments are on the KITTI-360 dataset. }
\vspace{-0.3cm}
\centering
\scriptsize
 \begin{tabular}{c|lcc|c|cccccccc|c}
 \toprule
 \textbf{BEV} & \textbf{Model} & \textbf{FV} & \textbf{PT} & \textbf{Backbone} & \textbf{Road} & \textbf{Side.} & \textbf{Build.} & \textbf{Terr.} & \textbf{Pers.} & \textbf{2-Wh.} & \textbf{Car} & \textbf{Truck} & \textbf{mIoU} \\
 \midrule
 \multirow{5}{*}{1\%} & PoBEV  & \xmark & - & \multirow{2}{*}{Native} & 60.41 & 20.97 & 24.65 & 23.38 & 0.15 & 0.23 & 21.71 & 1.23 & 19.09 \\
 & SkyEye  & \cmark & \cmark & & 69.26 & 33.48 & 32.79 & \textbf{39.46} & 0.00 & 0.34 & \textbf{32.36} & 7.93 & 26.94 \\
 \cmidrule{2-14}
 & PoBEV & \xmark & - & \multirow{3}{*}{DINOv2} & 62.36 & 21.02 & 27.18 & 24.22 & 0.04 & 0.12 & 17.50 & 0.95 & 19.17 \\
 & SkyEye & \cmark & \cmark & & 65.13 & 29.56 & 29.02 & 34.22 & 0.78 & 2.87 & 26.04 & 5.12 & 24.09 \\
 & \net & \xmark &\cmark & & \textbf{70.58} & \textbf{34.26} & \textbf{40.68} & 38.53 & \textbf{1.35} & \textbf{4.74} & 30.94 & \textbf{10.58} & \textbf{28.96} \\
\midrule
\multirow{5}{*}{5\%} & PoBEV  & \xmark & - & \multirow{2}{*}{Native} & 64.45 & 27.36 & 30.15 & 31.66 & 0.69 & 0.98 & 29.75 & 6.06 & 23.89 \\
 & SkyEye  & \cmark & \cmark & & 72.16 & 37.20 & 34.89 & \textbf{42.97} & \textbf{4.77} & \textbf{9.16} & \textbf{40.74} & 9.88 & 31.47 \\
 \cmidrule{2-14}
 & PoBEV  & \xmark & - & \multirow{3}{*}{DINOv2} & 67.61 & 30.73 & 30.97 & 32.80 & 0.42 & 0.47 & 25.48 & 5.58 & 24.26 \\
 & SkyEye  & \cmark & \cmark & & 69.84 & 34.19 & 32.80 & 37.13 & 2.54 & 4.74 & 32.49 & 7.93 & 27.71 \\
 & \net & \xmark & \cmark & & \textbf{73.74} & \textbf{39.56} & \textbf{42.07} & 41.49 & 2.46 & 6.32 & 34.68 & \textbf{14.88} & \textbf{31.90} \\
 \midrule
 \multirow{5}{*}{10\%} & PoBEV  & \xmark & - & \multirow{2}{*}{Native} & 66.58 & 30.28 & 31.76 & 34.50 & 1.22 & 3.28 & 33.43 & 7.56 & 26.08  \\
 & SkyEye  & \cmark & \cmark & & 73.36 & 38.30 & 37.54 & \textbf{44.62} & \textbf{4.80} & \textbf{9.67} & \textbf{42.84} & 10.06 & 32.65 \\
 \cmidrule{2-14}
 & PoBEV & \xmark & - & \multirow{3}{*}{DINOv2} & 68.99 & 33.17 & 35.81 & 34.15 & 0.70 & 1.58 & 29.74 & 10.06 & 26.77 \\
 & SkyEye & \cmark & \cmark & & 72.19 & 36.18 & 35.26 & 39.84 & 3.78 & 5.61 & 36.95 & 10.44 & 30.03 \\
 & \net & \xmark & \cmark & & \textbf{74.74} & \textbf{39.40} & \textbf{43.63} & 43.33 & 2.91 & 6.95 & 37.62 & \textbf{18.09} & \textbf{33.33} \\
 \midrule
 \multirow{5}{*}{50\%} & PoBEV  & \xmark & - & \multirow{2}{*}{Native} & 69.88 & 33.81 & 33.40 & 40.48 & 2.47 & 4.63 & 38.81 & 9.84 & 29.16 \\
 & SkyEye  & \cmark & \cmark & & 73.10 & \textbf{39.23} & 38.08 & \textbf{45.72} & \textbf{4.05} & \textbf{10.44} & \textbf{44.72} & 12.10 & \textbf{33.43} \\
 \cmidrule{2-14}
 & PoBEV & \xmark & - & \multirow{3}{*}{DINOv2} & 73.04 & 37.38 & 37.86 & 41.31 & 1.82 & 3.83 & 37.13 & 14.85 & 30.90 \\
 & SkyEye  & \cmark & \cmark & & 73.66 & 38.85 & 41.49 & 41.73 & 2.90 & 6.99 & 38.43 & 12.42 & 32.06  \\
 & \net & \xmark & \cmark & & \textbf{74.29} & 38.48 & \textbf{43.87} & 42.77 & 2.80 & 5.22 & 37.68 & \textbf{15.20} & 32.54 \\
\midrule
 \multirow{5}{*}{100\%} & PoBEV & \xmark & - & \multirow{2}{*}{Native} & 70.14 & 35.23 & 34.68 & 40.72 & 2.85 & 5.63 & 39.77 & 14.38 & 30.42 \\
 & SkyEye & \cmark & \cmark & & 73.57 & \textbf{39.45} & 38.74 & \textbf{46.06} & \textbf{3.95} & \textbf{9.66} & \textbf{45.21} & 10.92 & \textbf{33.44} \\
 \cmidrule{2-14}
 & PoBEV & \xmark & - & \multirow{3}{*}{DINOv2} & 73.29 & 37.81 & 40.23 & 42.11 & 1.78 & 3.32 & 38.66 & \textbf{17.42} & 31.83 \\
 & SkyEye & \cmark & \cmark & & 73.51 & 39.13 & 40.04 & 42.08 & 3.17 & 5.90 & 39.29 & 12.72 & 31.98 \\
 & \net & \xmark & \cmark & & \textbf{74.81} & 38.59 & \textbf{42.58} & 43.67 & 3.52 & 6.21 & 38.47 & 15.24 & 32.88 \\
 \bottomrule
 \end{tabular}
\label{tab:supp-baseline-dinov2}
\vspace{-0.3cm}
\end{table}

\subsection{BEV Finetuning using SkyEye Split}
\label{subsec:supp-finetuning-skyeye-split}
In this section, we report the results obtained upon finetuning both the baselines as well as our model with the single random set generated for each BEV percentage split as defined in SkyEye~\cite{cit:bev-seg-skyeye}. Please note that all networks are finetuned using the corresponding percent of BEV ground truth labels. \tabref{tab:supp-percentages-skyeye-split} presents the results of this study. We observe that our pretraining strategy significantly improves the performance of our model across all three percentage splits with the largest improvement of \SI{1.98}{pp} observed when using $1\%$ of BEV labels. We also note that \net~outperforms SkyEye by \SI{0.99}{pp} and \SI{1.88}{pp} when using $1\%$ and $10\%$ which highlights the impact of our approach in low label regimes. Thus, in line with \secref{subsec:ablation-study} and \tabref{tab:ablation-percentages}, we conclude that our novel pretraining strategy positively influences network performance on the BEV segmentation task and results in competitive segmentation performance even in extremely low label regimes. 

\begin{table}[t]
\caption{Ablation study on the impact of our unsupervised pretraining on the overall network performance using the finetuning split defined in SkyEye. All experiments are on the KITTI-360 dataset. }
\vspace{-0.3cm}
\centering
\scriptsize
 \begin{tabular}{c|lcc|c|cccccccc|c}
 \toprule
 \textbf{BEV} & \textbf{Model} & \textbf{FV} & \textbf{PT} & \textbf{Epochs} & \textbf{Road} & \textbf{Side.} & \textbf{Build.} & \textbf{Terr.} & \textbf{Pers.} & \textbf{2-Wh.} & \textbf{Car} & \textbf{Truck} & \textbf{mIoU} \\
 \midrule
 \multirow{4}{*}{1\%} & PoBEV  & \xmark & - & \multirow{4}{*}{100} & 61.70 & 17.10 & 27.81 & 26.72 & 0.07 & 0.36 & 21.51 & 0.84 & 19.51 \\
 & SkyEye  & \cmark & \cmark & & 72.56 & 34.33 & 36.70 & \textbf{41.66} & 0.00 & 0.16 & \textbf{33.85} & 10.29 & 28.71 \\
 & \net & \xmark & \xmark & & 70.89 & 33.88 & 37.71 & 37.41 & 0.80 & \textbf{2.87} & 31.59 & 6.59 & 27.72 \\
 & \net & \xmark & \cmark & & \textbf{72.94} & \textbf{37.79} & \textbf{43.70} & 38.29 & \textbf{0.87} & 2.57 & 30.62 & \textbf{10.86} & \textbf{29.70} \\
 \midrule
 \multirow{4}{*}{10\%} & PoBEV  & \xmark & - & \multirow{4}{*}{50} & 70.00 & 32.75 & 38.07 & 34.43 & 0.80 & 3.33 & 34.46 & 9.25 & 27.89  \\
 & SkyEye  & \cmark & \cmark & & 76.07 & 40.30 & 40.30 & \textbf{45.33} & 3.75 & 8.15 & \textbf{42.64} & 10.73 & 33.41  \\
 & \net & \xmark & \xmark & & \textbf{76.69} & 40.41 & 42.55 & 42.17 & 1.33 & 6.57 & 40.46 & 18.06 & 33.53 \\
 & \net & \xmark & \cmark & & 74.47 & \textbf{41.16} & \textbf{46.31} & 43.31 & \textbf{5.48} & \textbf{8.80} & 41.55 & \textbf{21.24} & \textbf{35.29} \\
 \midrule
 \multirow{4}{*}{50\%} & PoBEV  & \xmark & - & \multirow{4}{*}{30} & 72.09 & 35.64 & 36.64 & 42.41 & 1.61 & 3.92 & 41.41 & 9.77 & 30.44 \\
 & SkyEye  & \cmark & \cmark & & 76.43 & 39.89 & 45.22 & \textbf{46.64} & \textbf{5.10} & 7.93 & \textbf{42.43} & 12.30 & 34.49 \\
 & \net & \xmark & \xmark & & 75.46 & 39.45 & 42.71 & 39.69 & 3.85 & 5.70 & 41.88 & 17.82 & 33.32 \\
 & \net & \xmark & \cmark & & \textbf{76.54} & \textbf{42.65} & \textbf{49.23} & 41.47 & 3.36 & \textbf{8.61} & 38.76 & \textbf{19.42} & \textbf{35.01} \\
 \bottomrule
 \end{tabular}
\label{tab:supp-percentages-skyeye-split}
\vspace{-0.3cm}
\end{table}

\section{Additional Qualitative Results}
\label{sec:supp-qualitative-results}

\begin{figure}
\centering
\footnotesize
\setlength{\tabcolsep}{0.05cm}
{
\renewcommand{\arraystretch}{0.15}
\newcolumntype{M}[1]{>{\centering\arraybackslash}m{#1}}
\begin{tabular}{cM{3.1cm}M{1.55cm}M{1.55cm}M{1.55cm}M{1.55cm}M{1.55cm}}
& Input FV Image & $1\%$ & $5\%$ & $10\%$ & $50\%$ & $100\%$ \\
\\
(a) & {\includegraphics[width=\linewidth, height=0.455\linewidth, frame]{images/qual_supp/kitti360/rgb_kitti-437.png}} & {\includegraphics[width=\linewidth, frame]{images/qual_supp/kitti360/letsmap_1_kitti-437.png}} & {\includegraphics[width=\linewidth, frame]{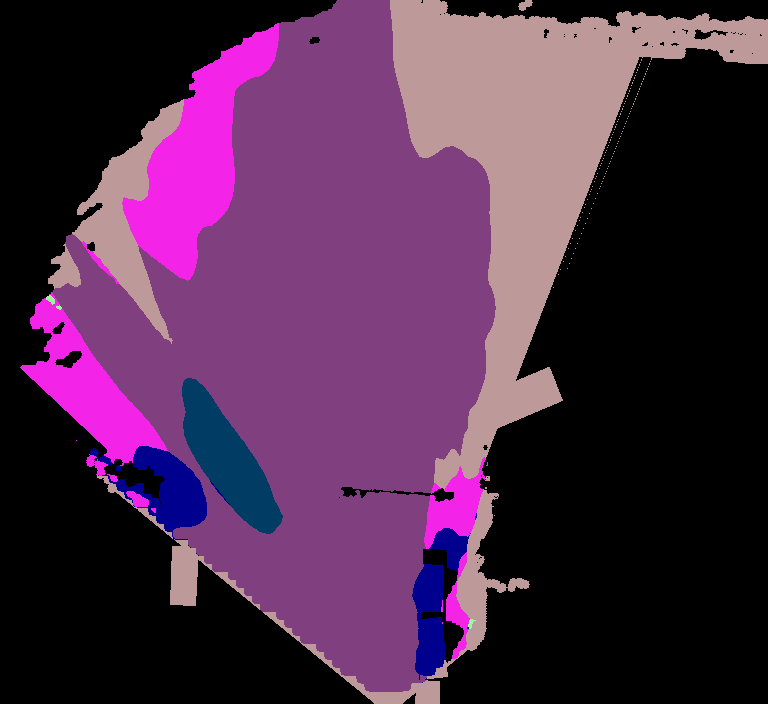}} & {\includegraphics[width=\linewidth, frame]{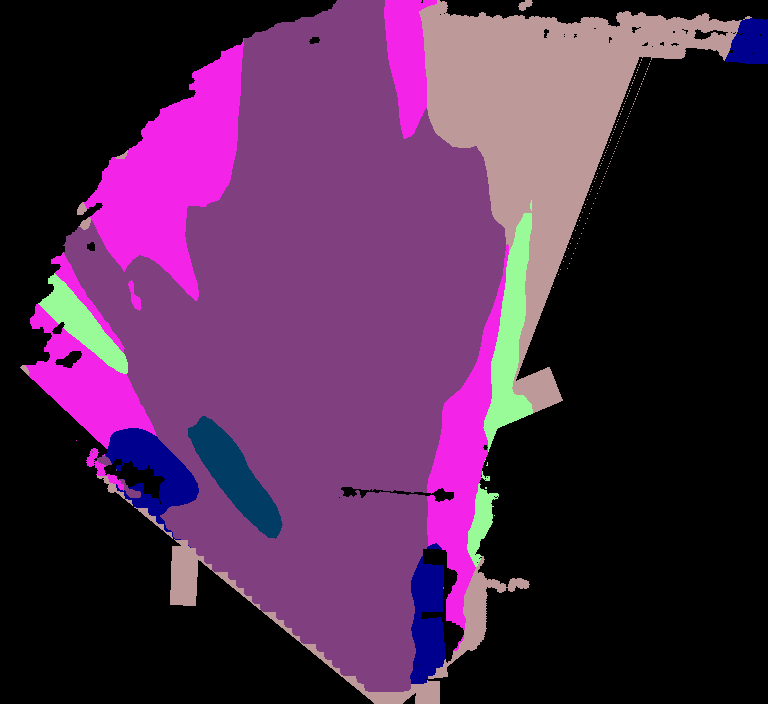}} & {\includegraphics[width=\linewidth, frame]{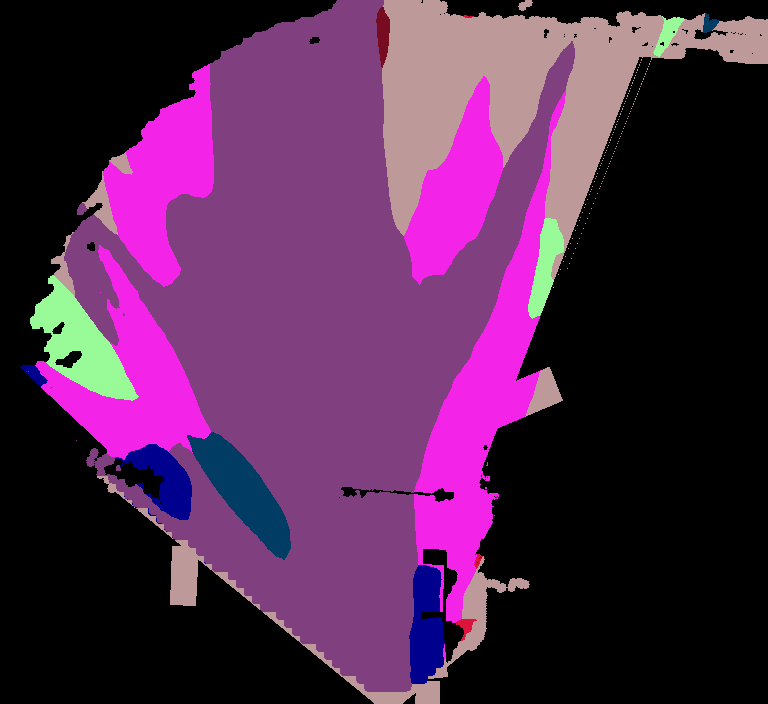}} & {\includegraphics[width=\linewidth, frame]{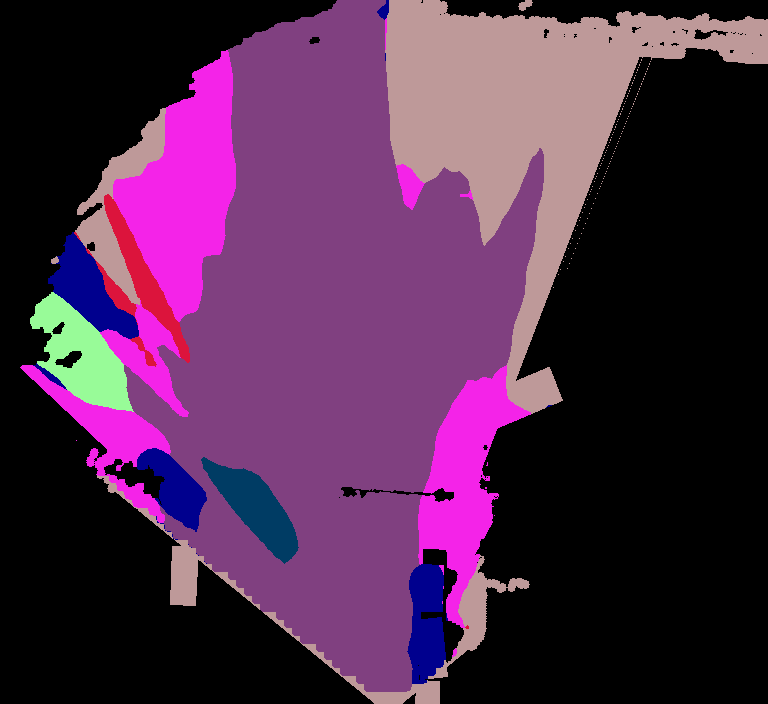}}  \\
\\
(b) & {\includegraphics[width=\linewidth, height=0.455\linewidth, frame]{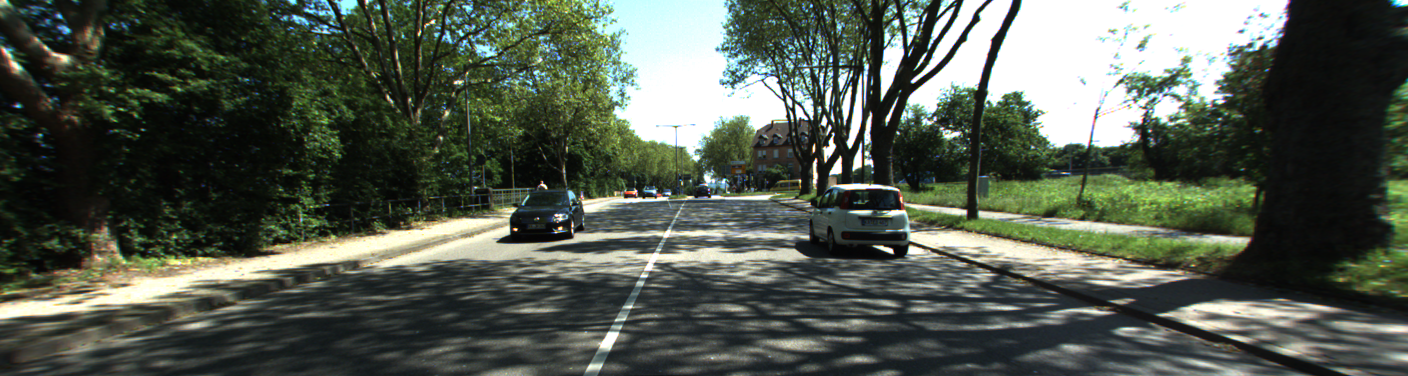}} & {\includegraphics[width=\linewidth, frame]{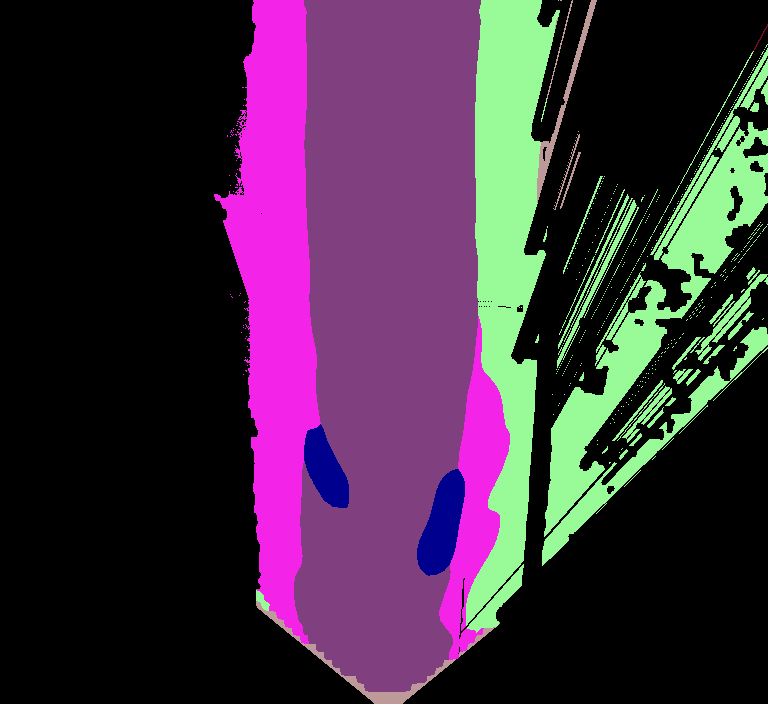}} & {\includegraphics[width=\linewidth, frame]{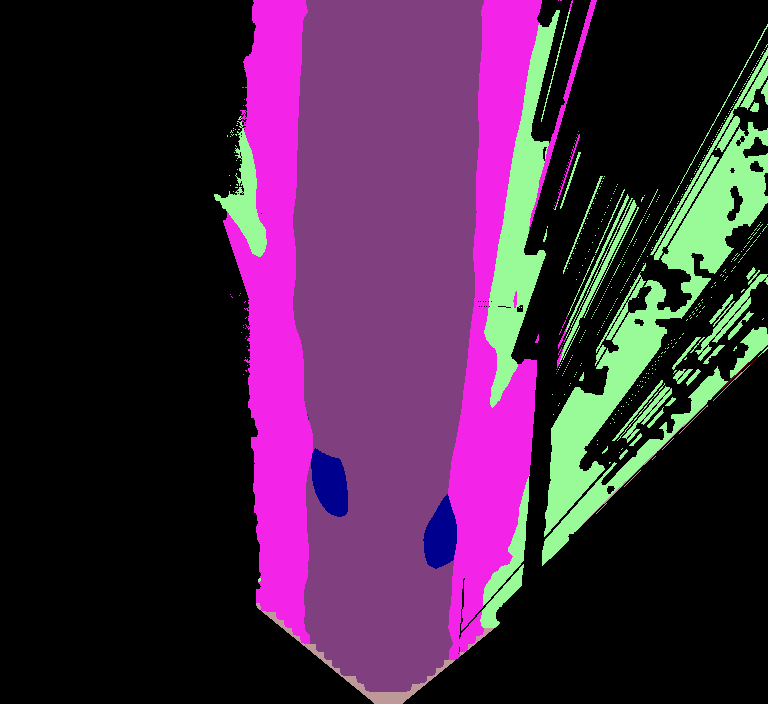}} & {\includegraphics[width=\linewidth, frame]{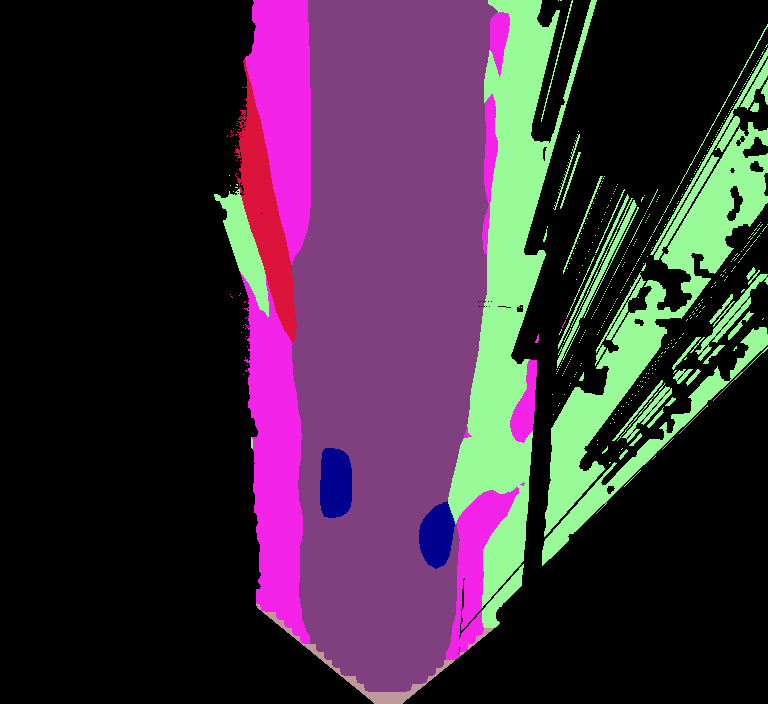}} & {\includegraphics[width=\linewidth, frame]{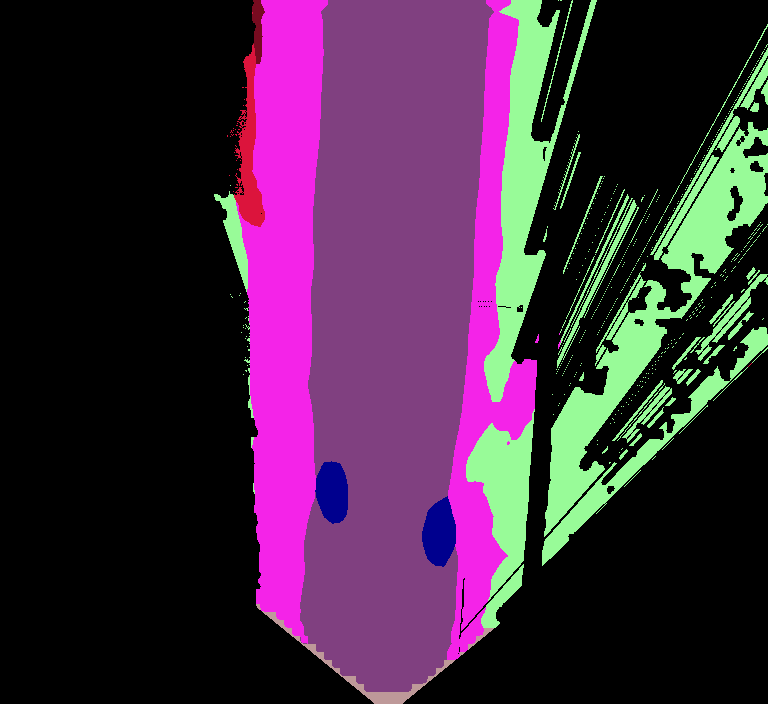}} & {\includegraphics[width=\linewidth, frame]{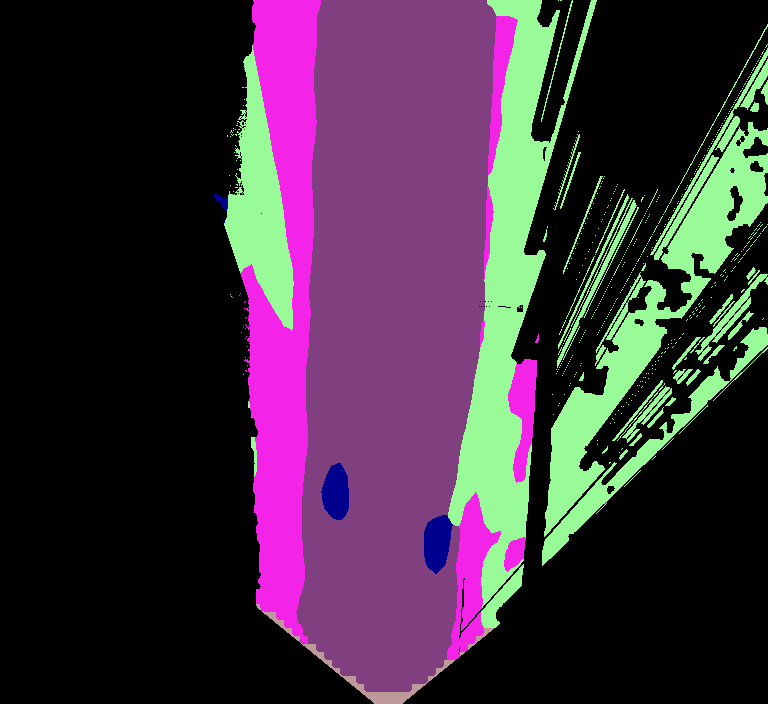}}  \\
\\
(c) & {\includegraphics[width=\linewidth, height=0.455\linewidth, frame]{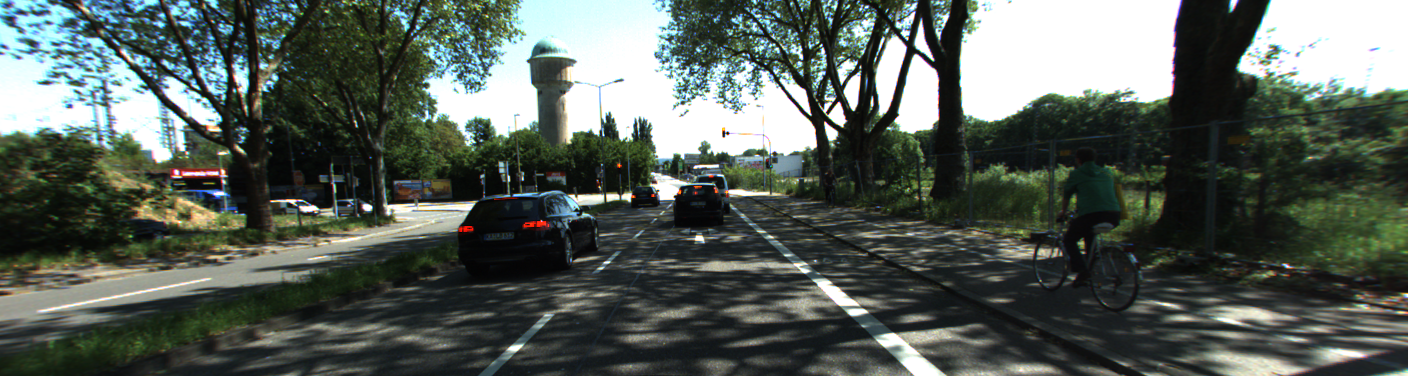}} & {\includegraphics[width=\linewidth, frame]{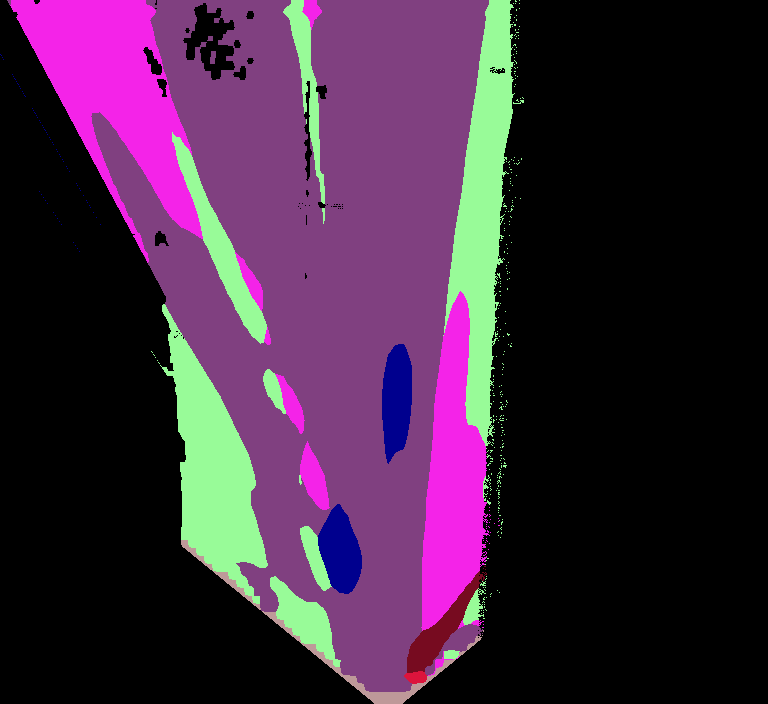}} & {\includegraphics[width=\linewidth, frame]{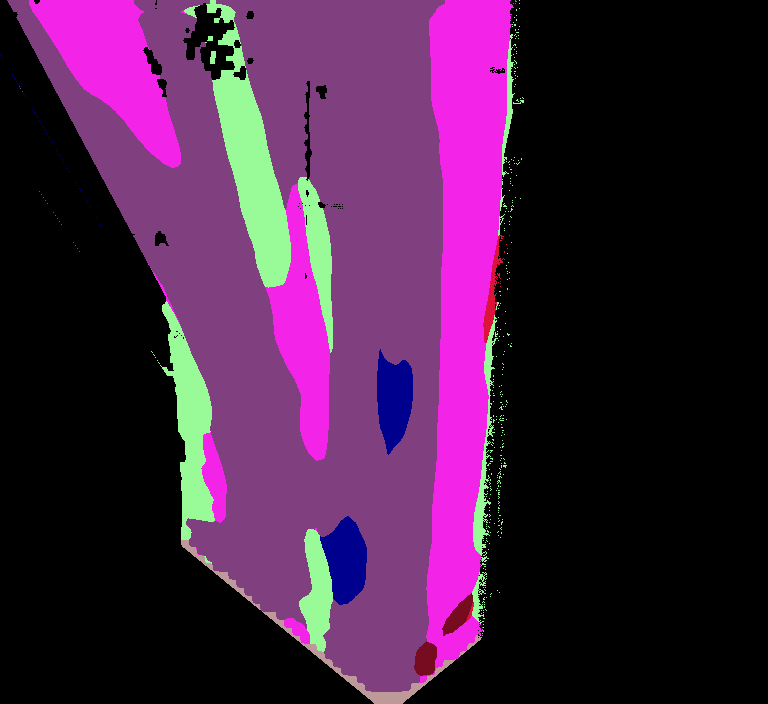}} & {\includegraphics[width=\linewidth, frame]{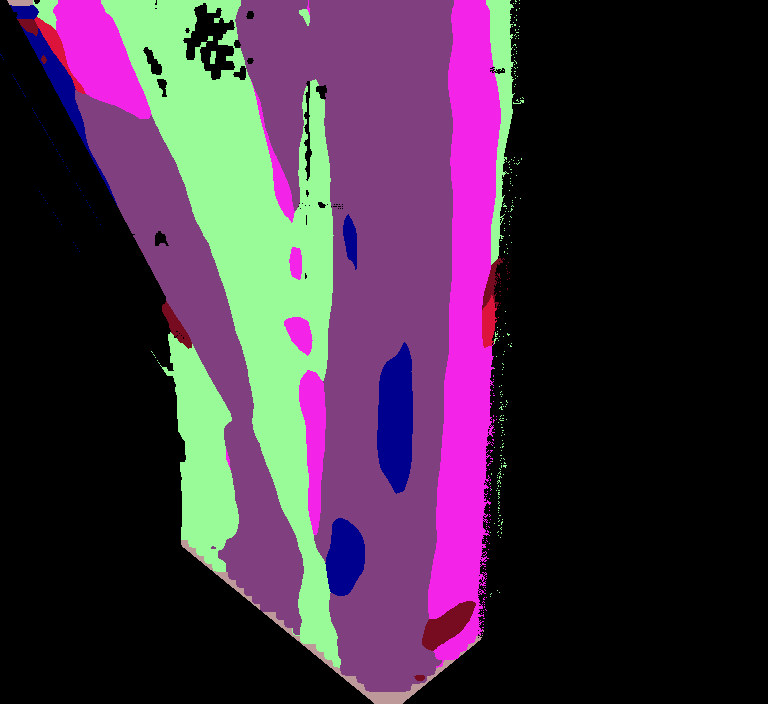}} & {\includegraphics[width=\linewidth, frame]{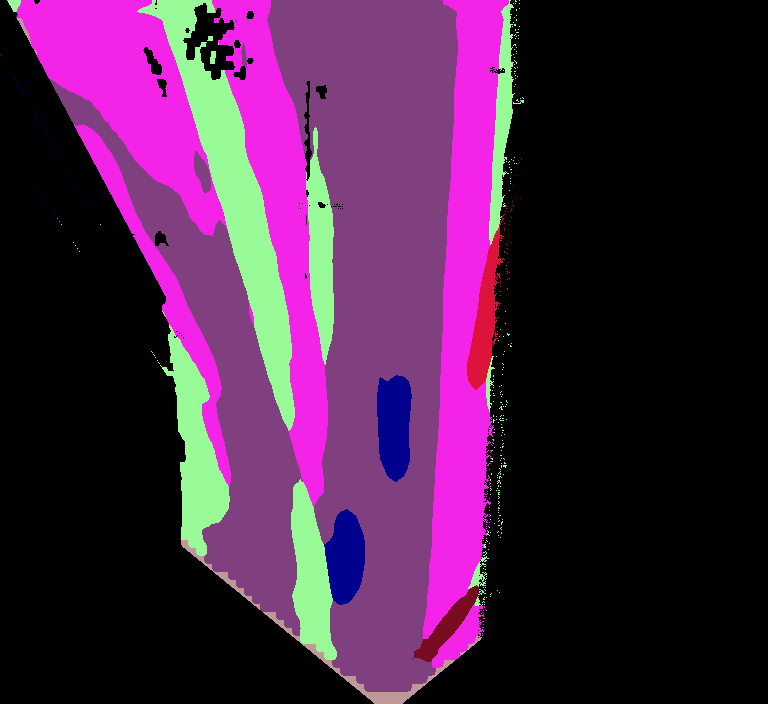}} & {\includegraphics[width=\linewidth, frame]{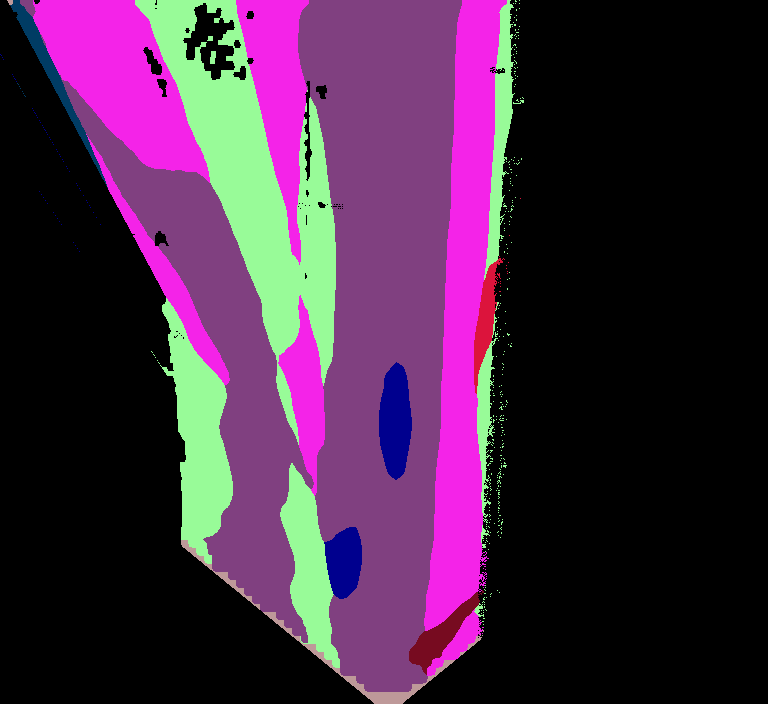}}  \\
\\
(d) & {\includegraphics[width=\linewidth, height=0.455\linewidth, frame]{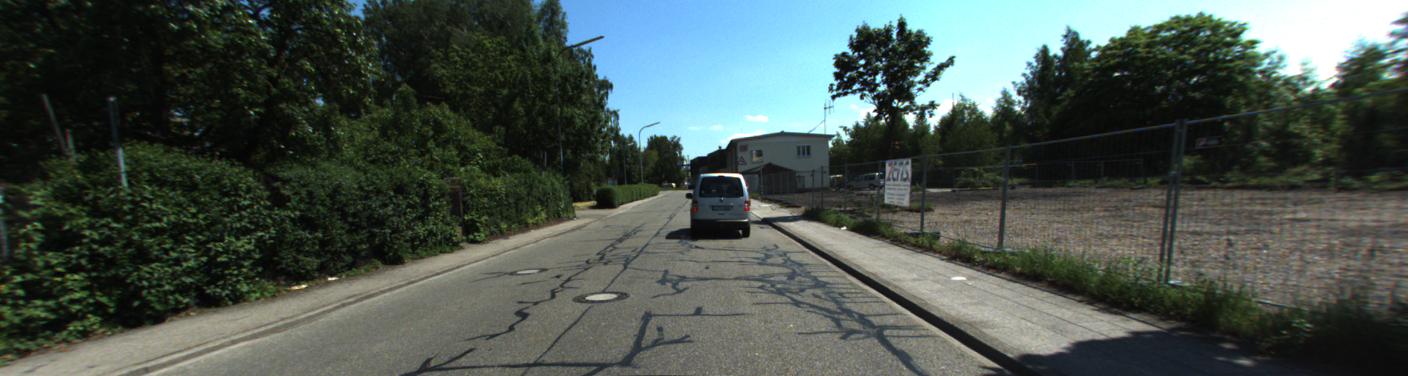}} & {\includegraphics[width=\linewidth, frame]{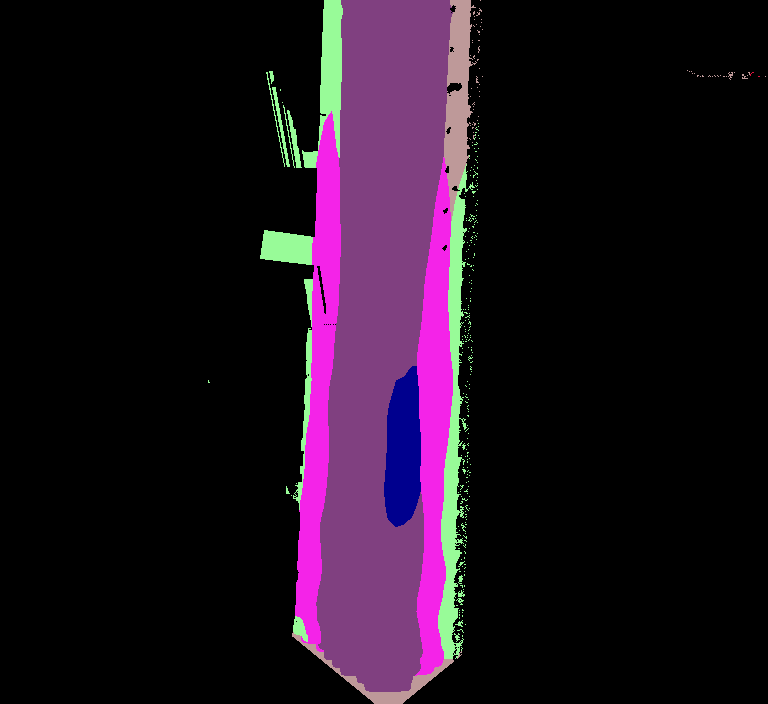}} & {\includegraphics[width=\linewidth, frame]{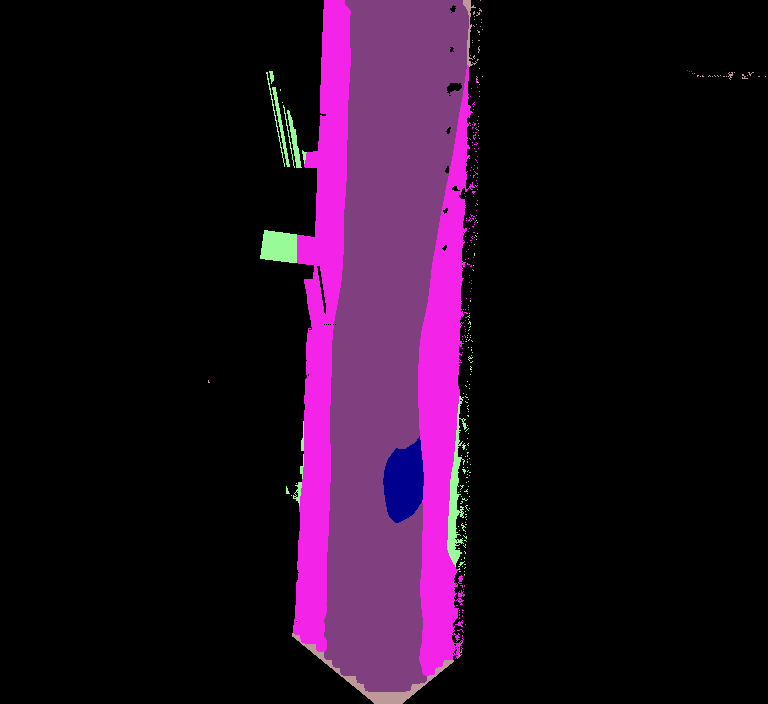}} & {\includegraphics[width=\linewidth, frame]{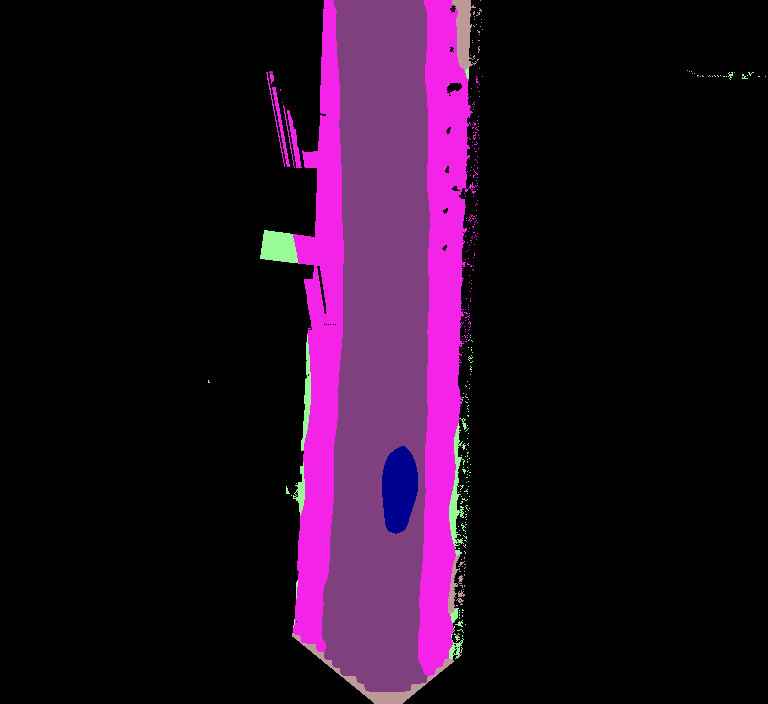}} & {\includegraphics[width=\linewidth, frame]{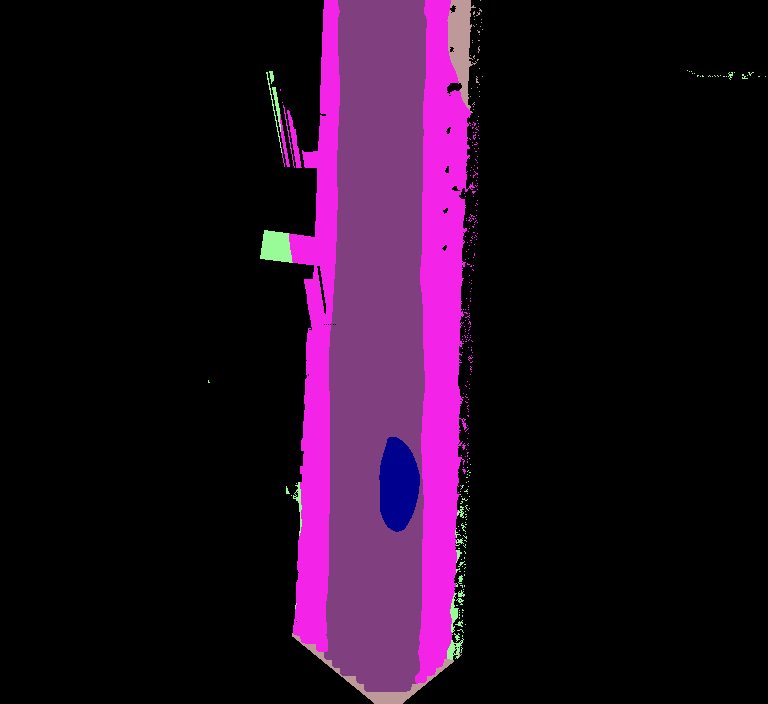}} & {\includegraphics[width=\linewidth, frame]{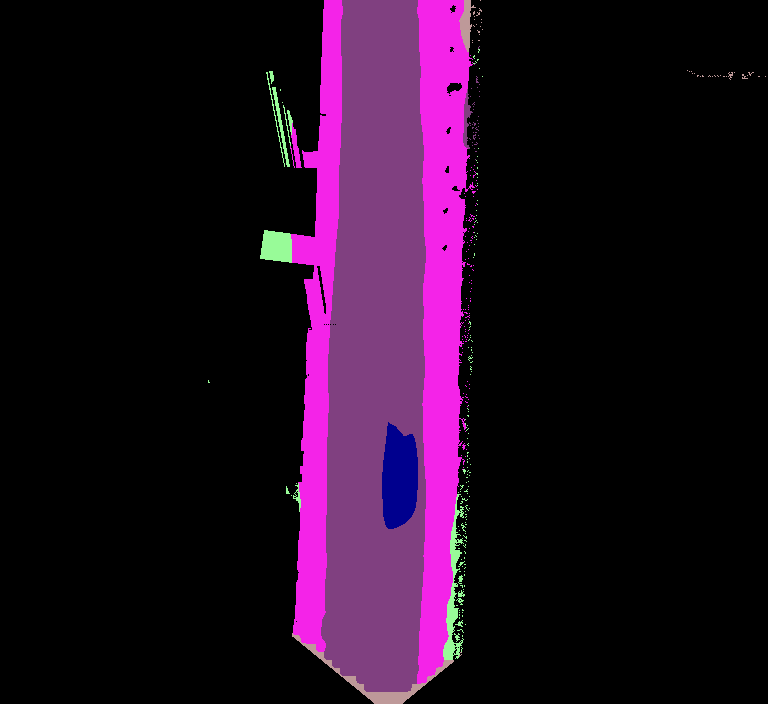}}  \\
\\
(e) & {\includegraphics[width=\linewidth, height=0.43\linewidth, frame]{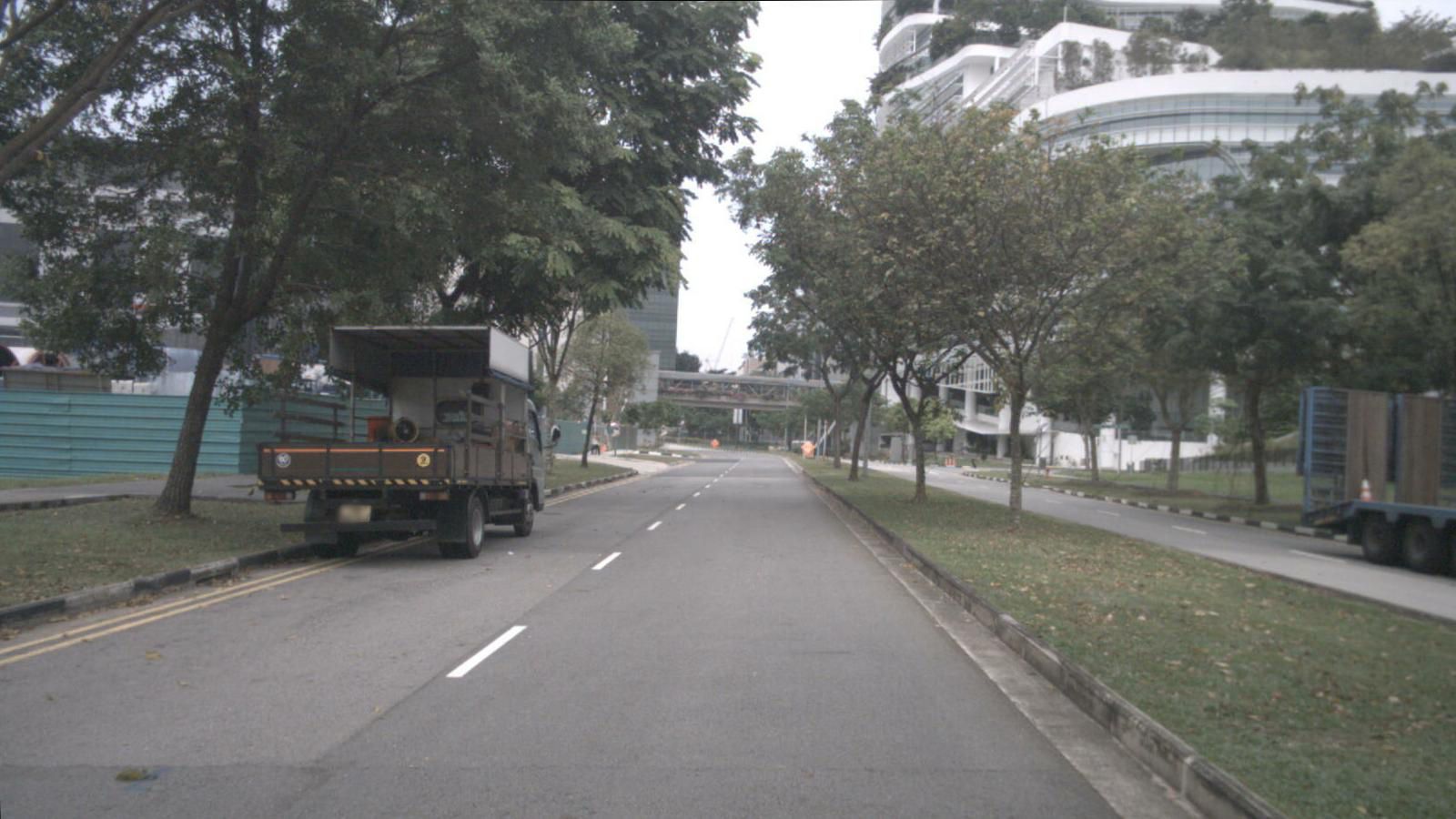}} & 
{\includegraphics[width=\linewidth, frame]{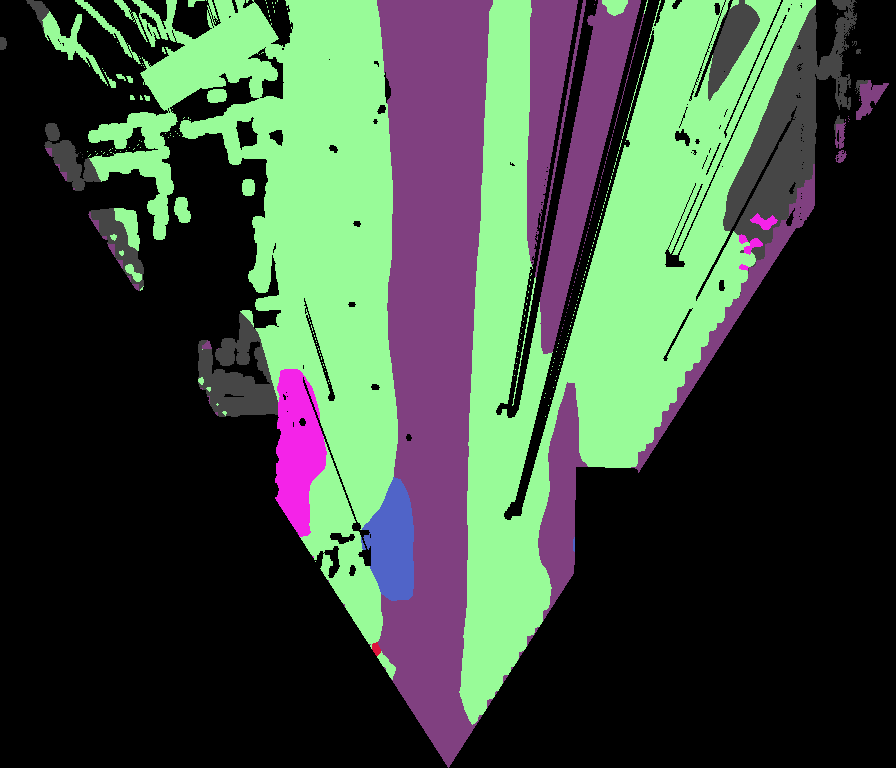}} & {\includegraphics[width=\linewidth, frame]{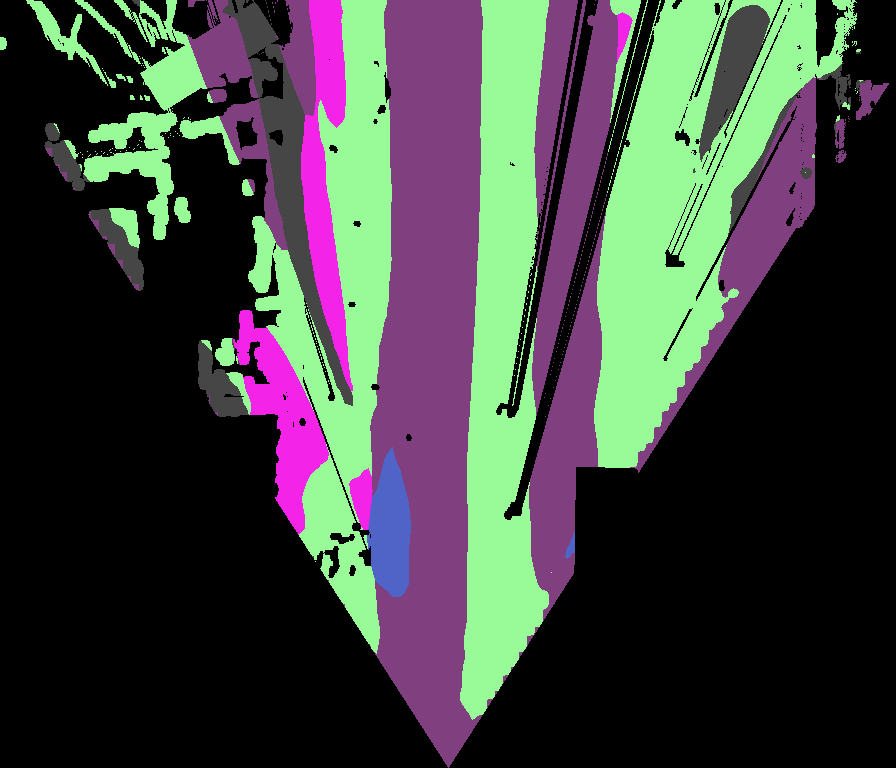}} & 
{\includegraphics[width=\linewidth, frame]{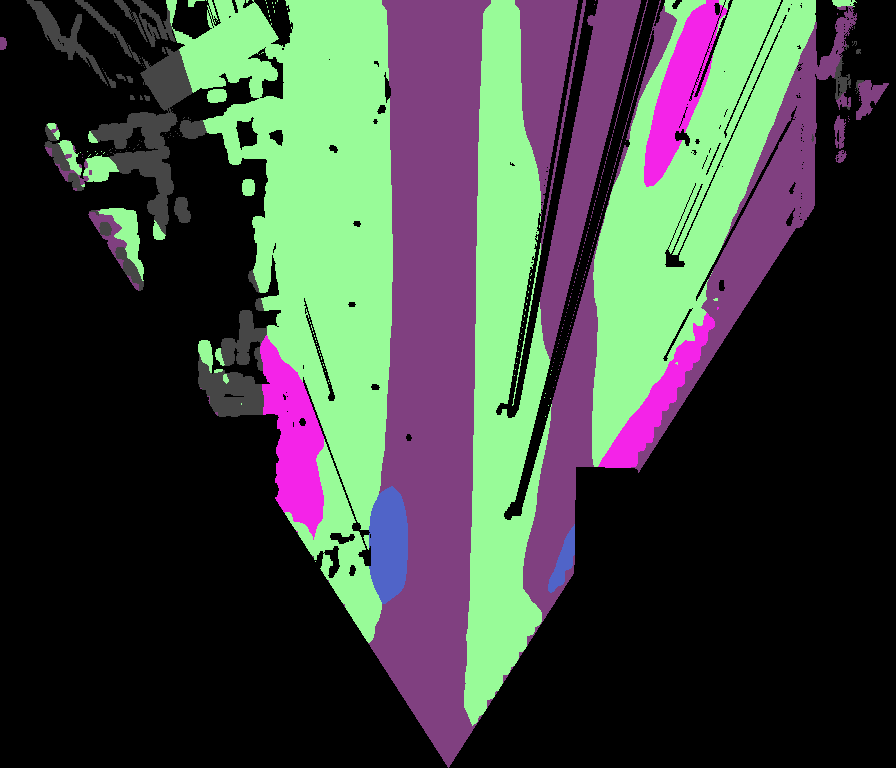}} & {\includegraphics[width=\linewidth, frame]{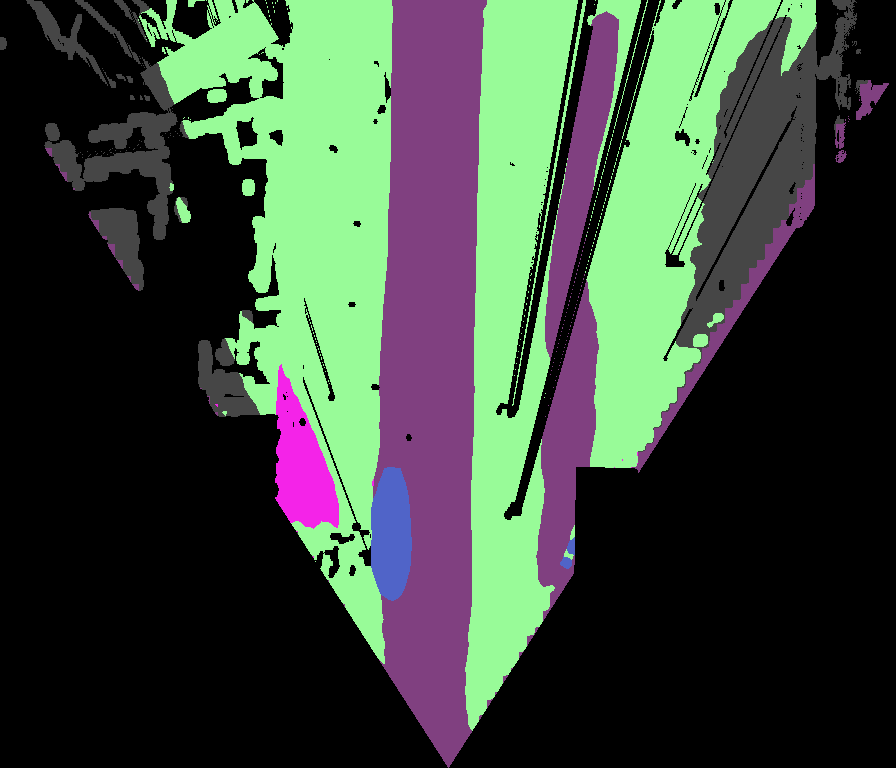}} & {\includegraphics[width=\linewidth, frame]{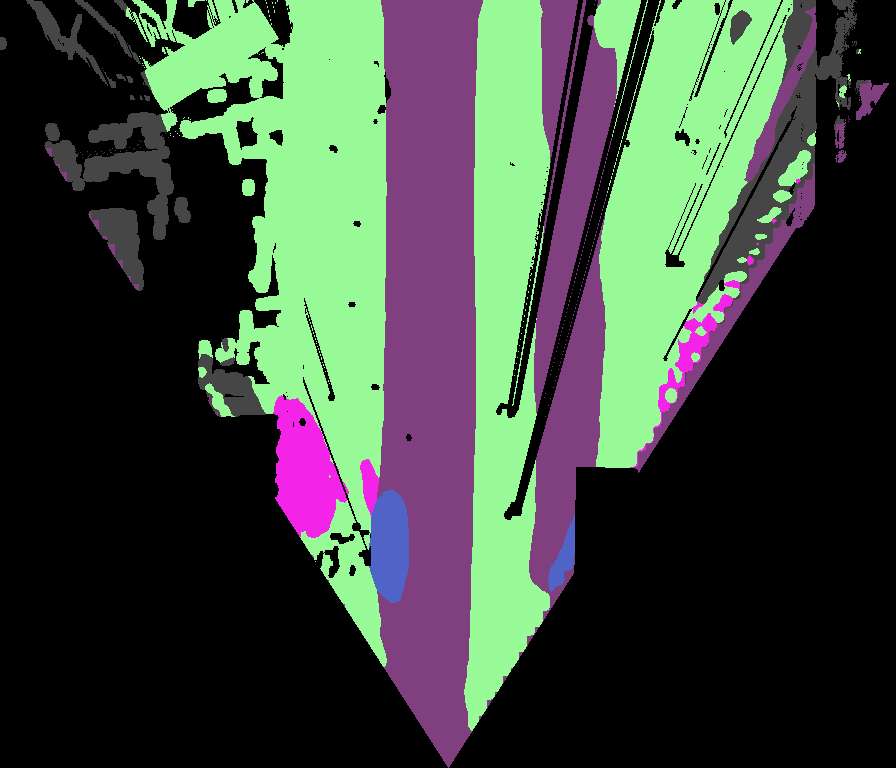}}  \\
\\
(f) & {\includegraphics[width=\linewidth, height=0.43\linewidth, frame]{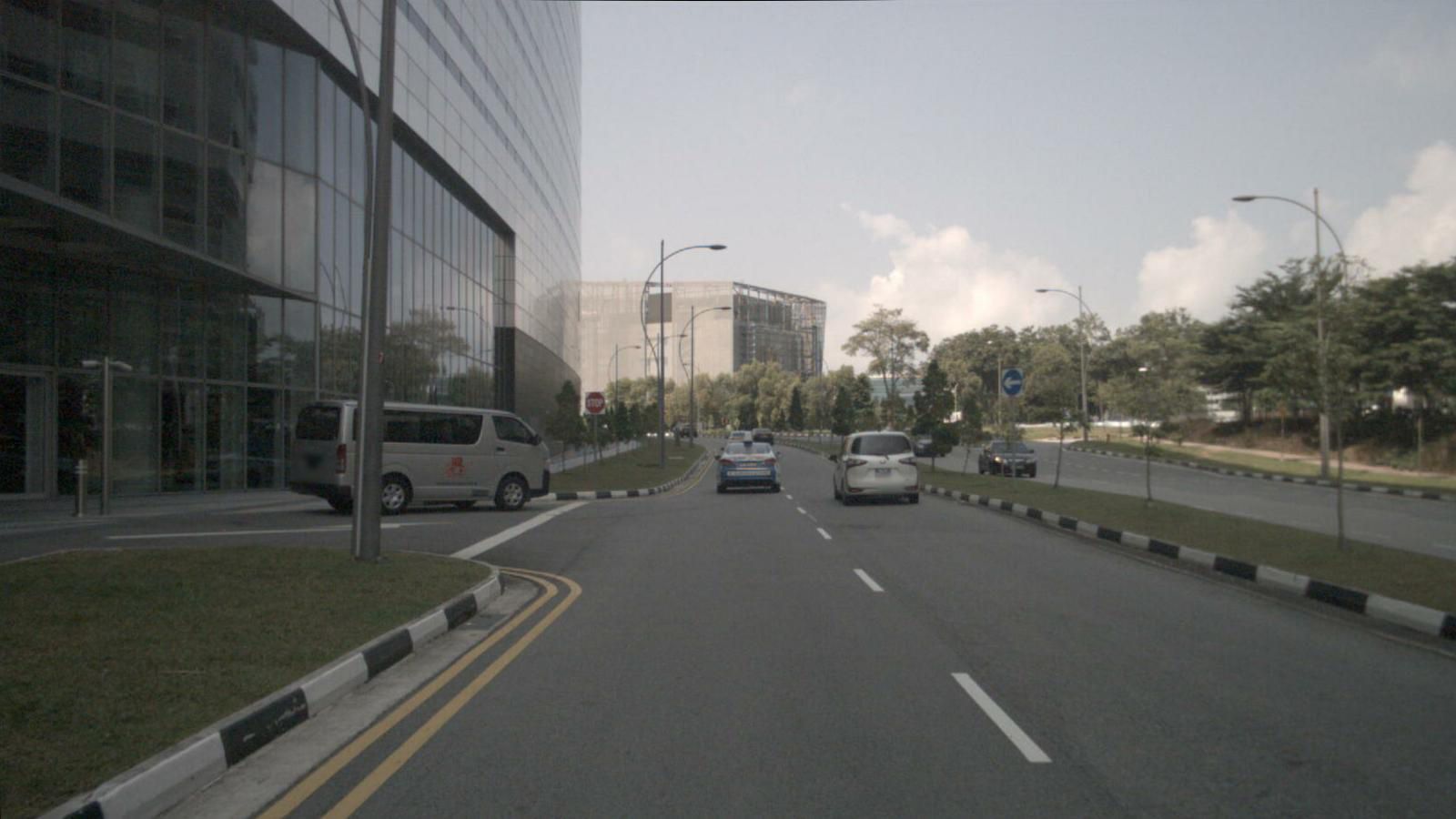}} & 
{\includegraphics[width=\linewidth, frame]{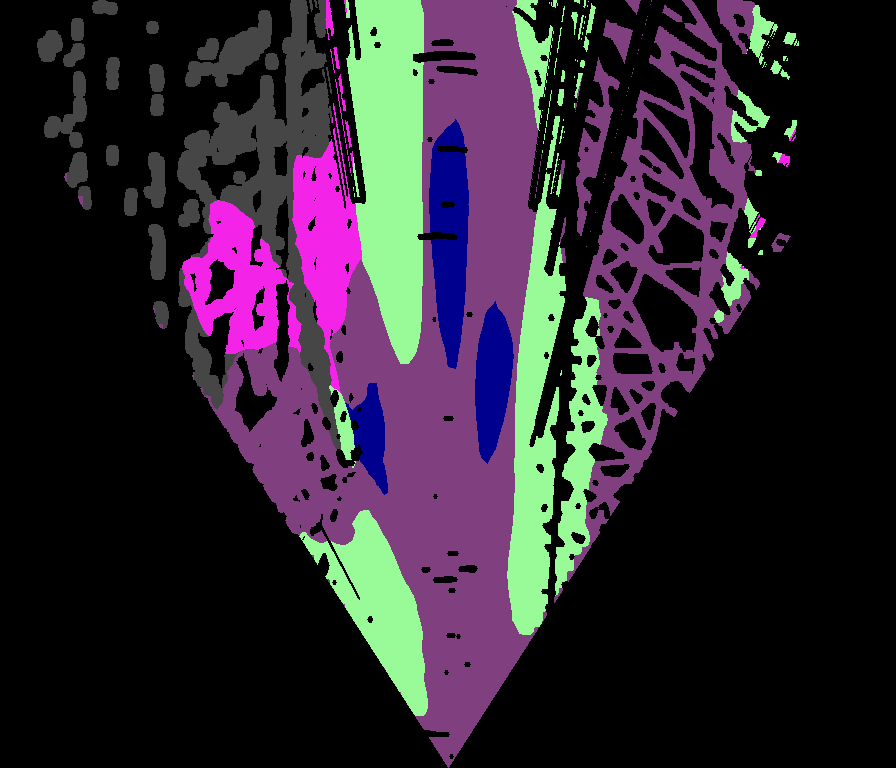}} & {\includegraphics[width=\linewidth, frame]{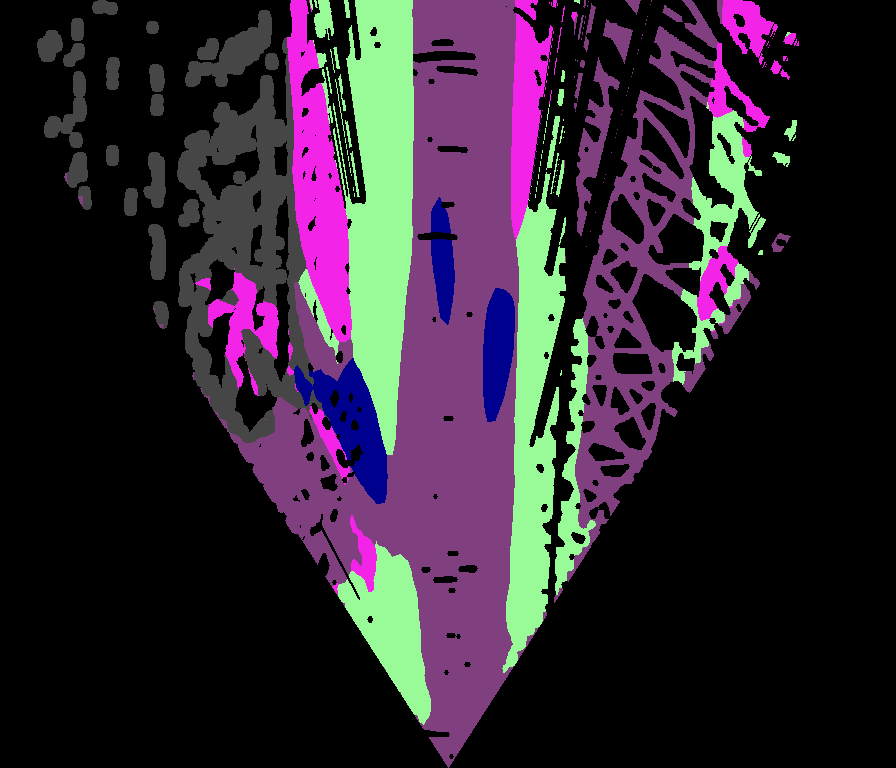}} & 
{\includegraphics[width=\linewidth, frame]{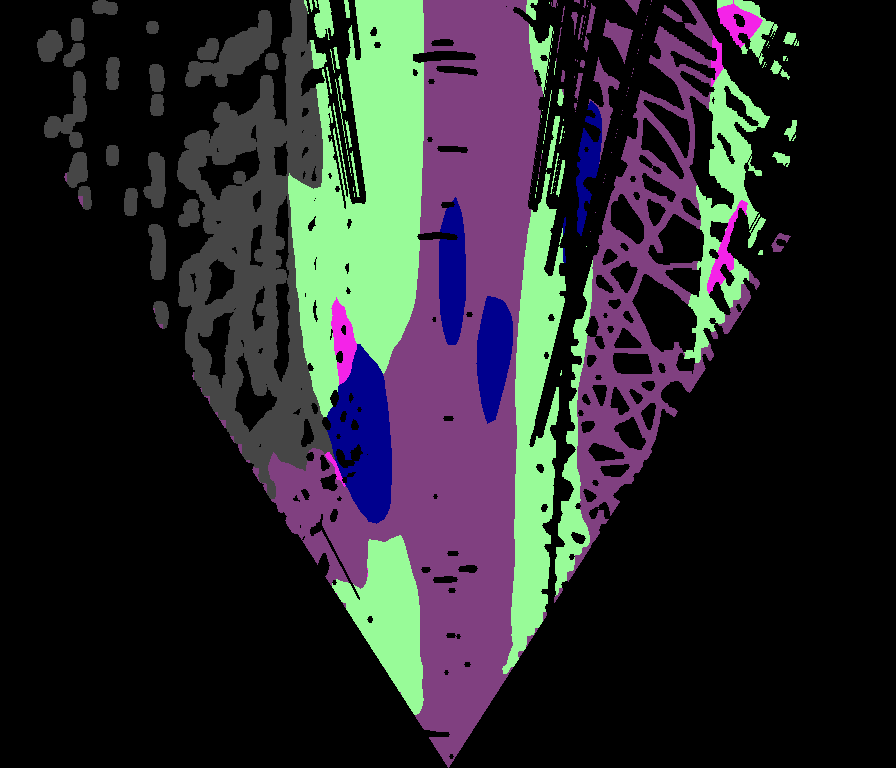}} & {\includegraphics[width=\linewidth, frame]{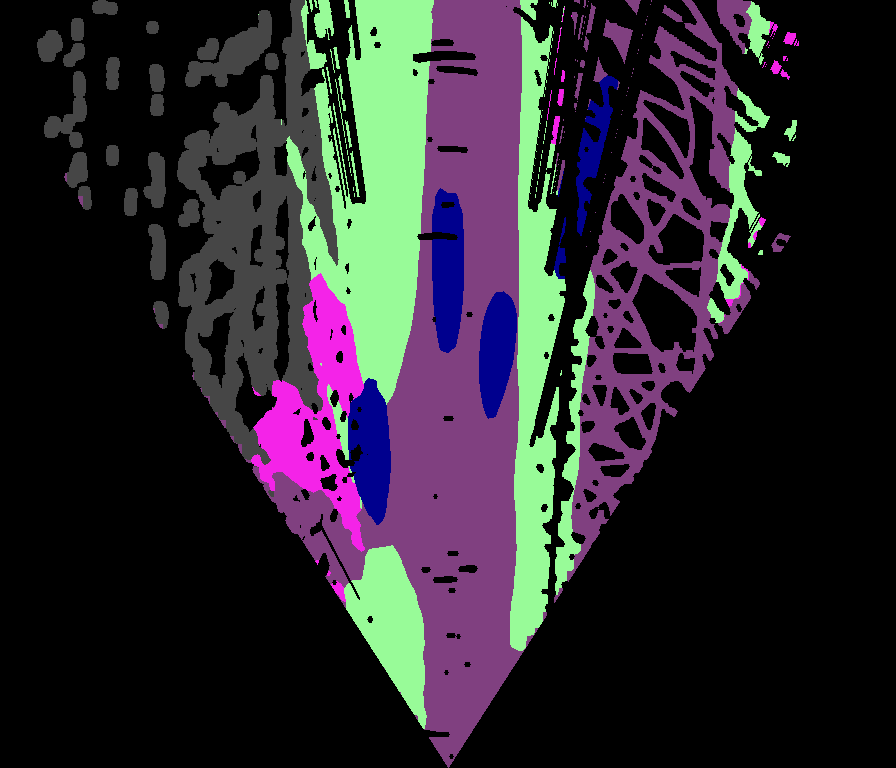}} & {\includegraphics[width=\linewidth, frame]{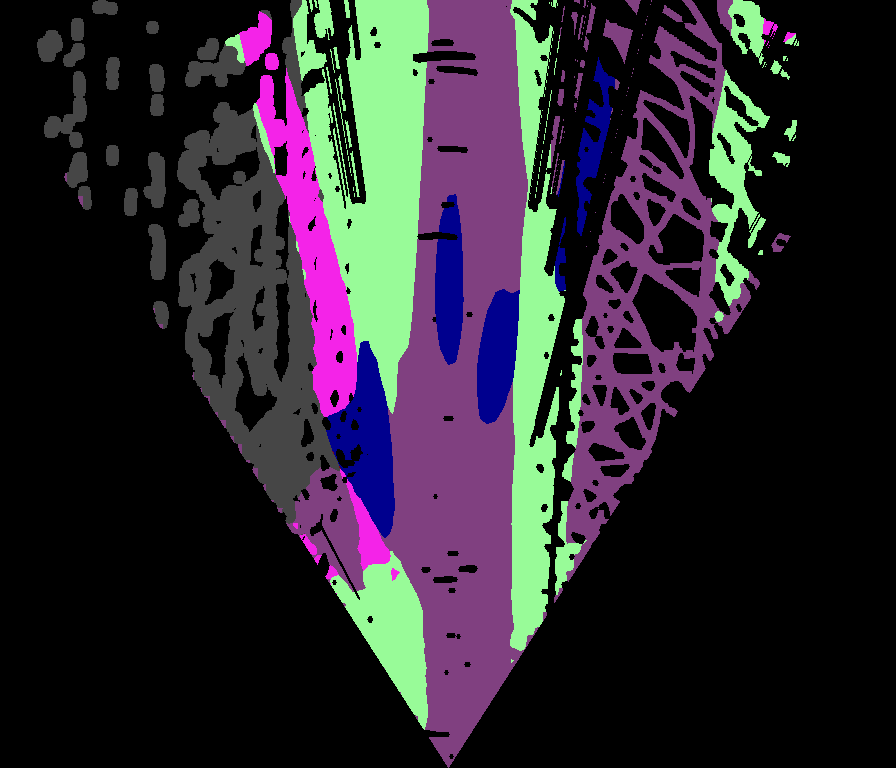}}  \\
\\
(g) & {\includegraphics[width=\linewidth, height=0.43\linewidth, frame]{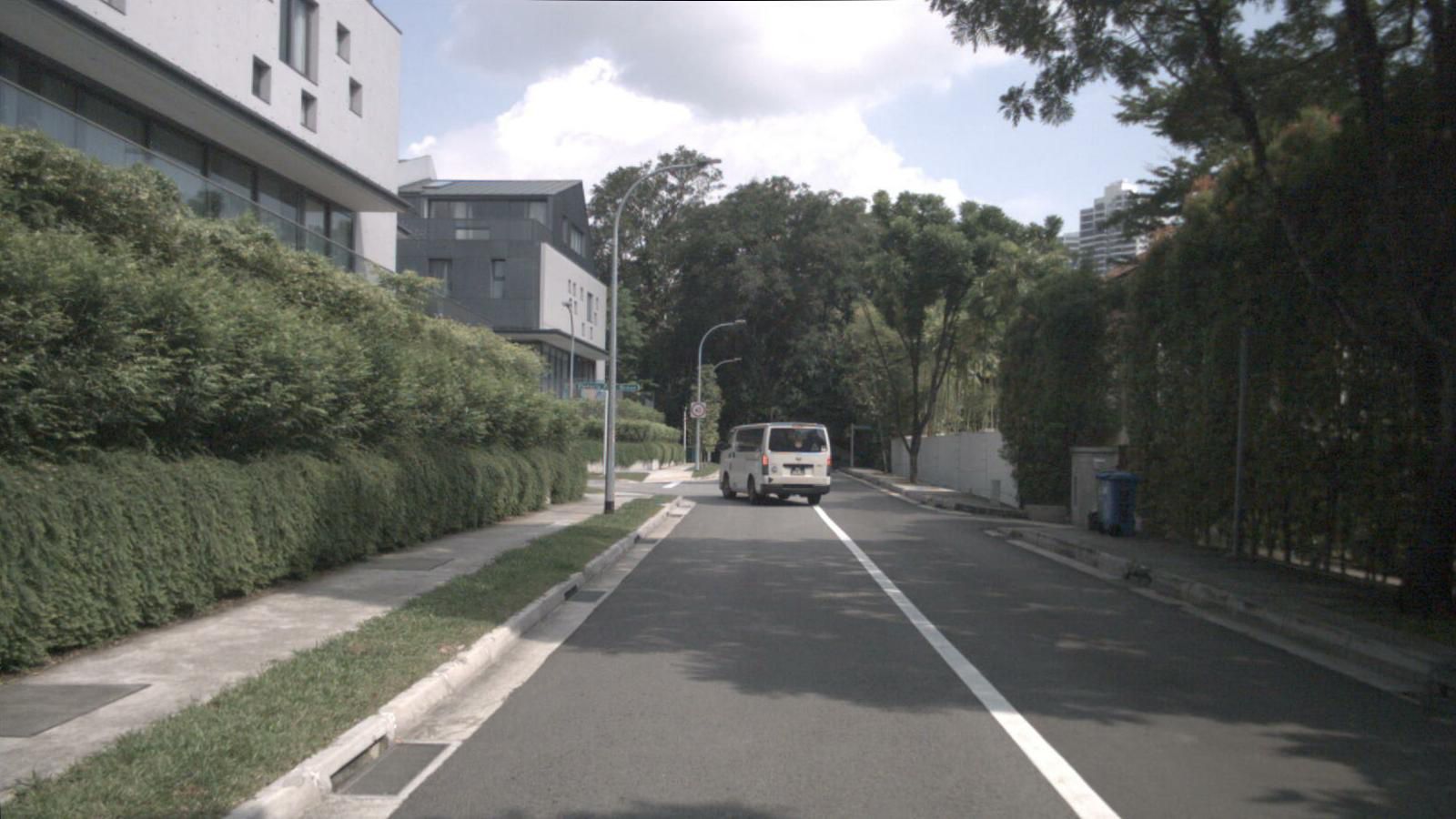}} & 
{\includegraphics[width=\linewidth, frame]{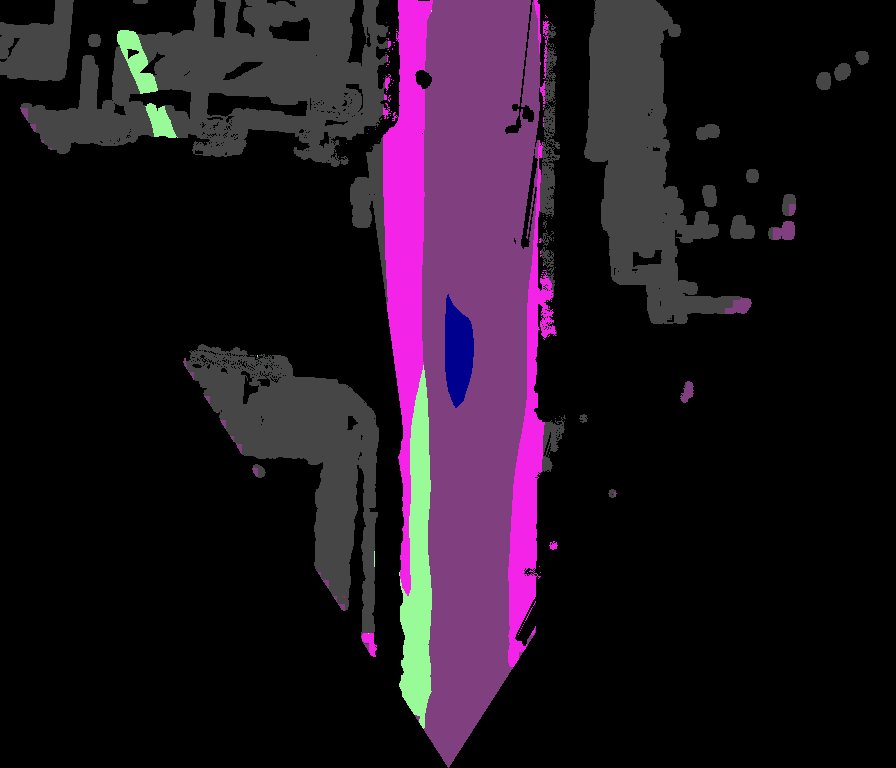}} & {\includegraphics[width=\linewidth, frame]{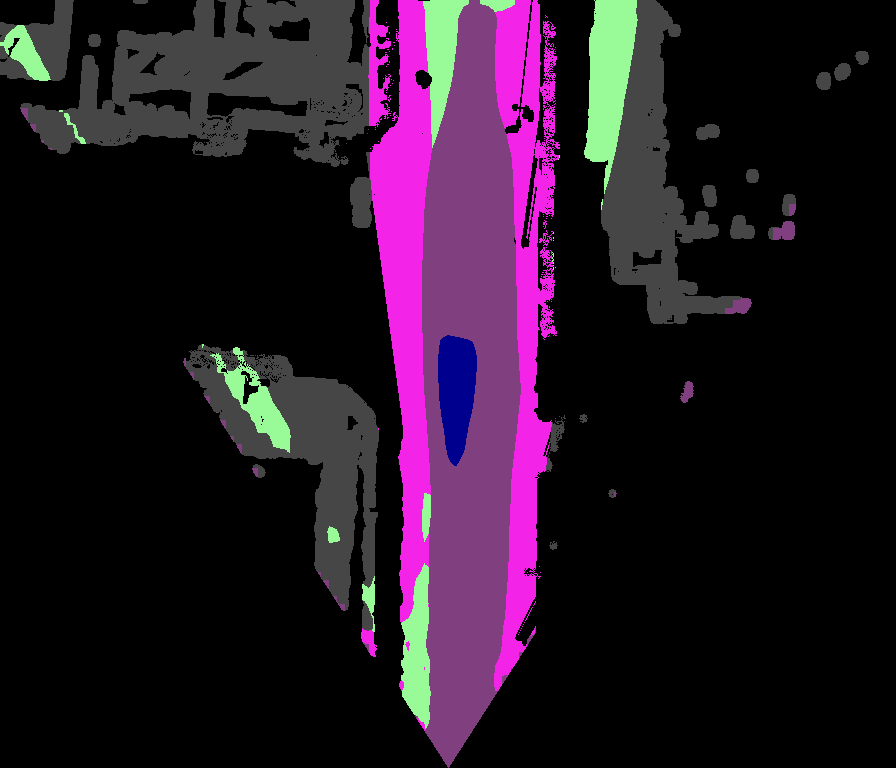}} & 
{\includegraphics[width=\linewidth, frame]{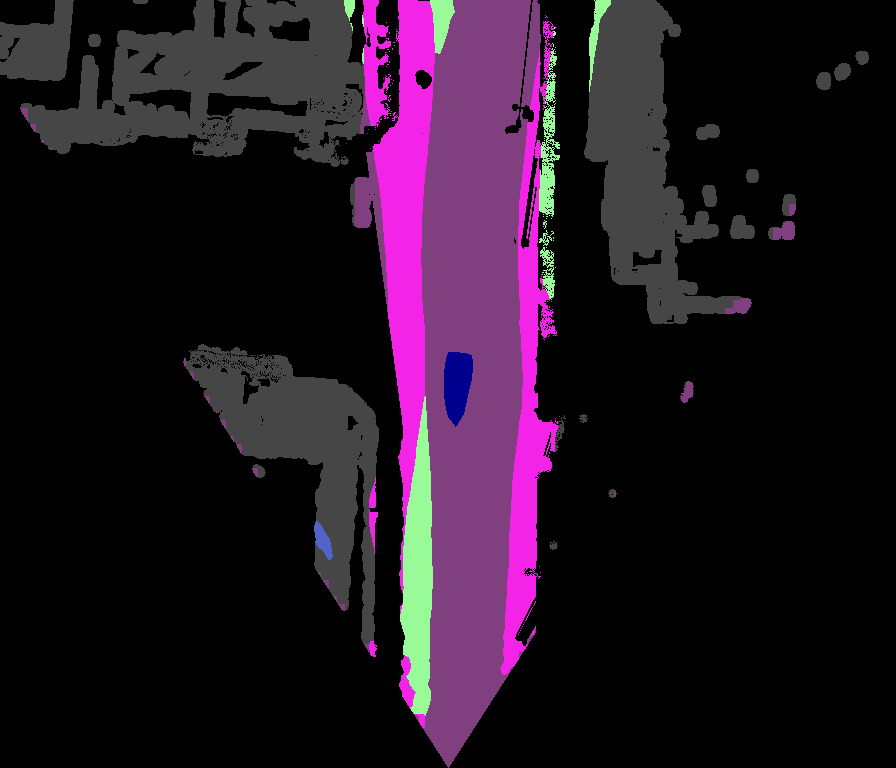}} & {\includegraphics[width=\linewidth, frame]{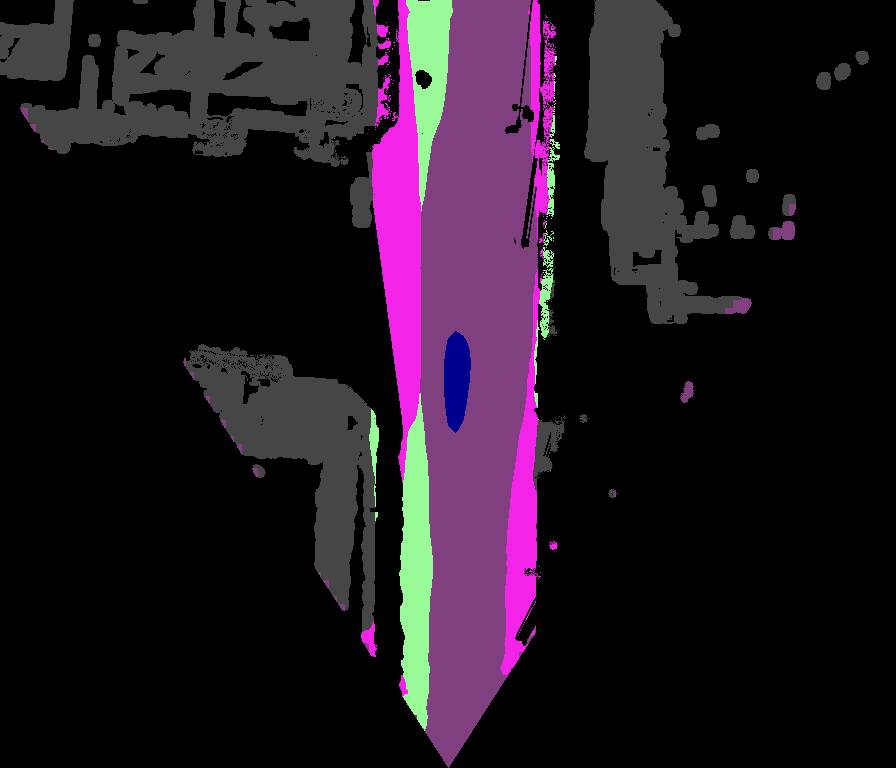}} & {\includegraphics[width=\linewidth, frame]{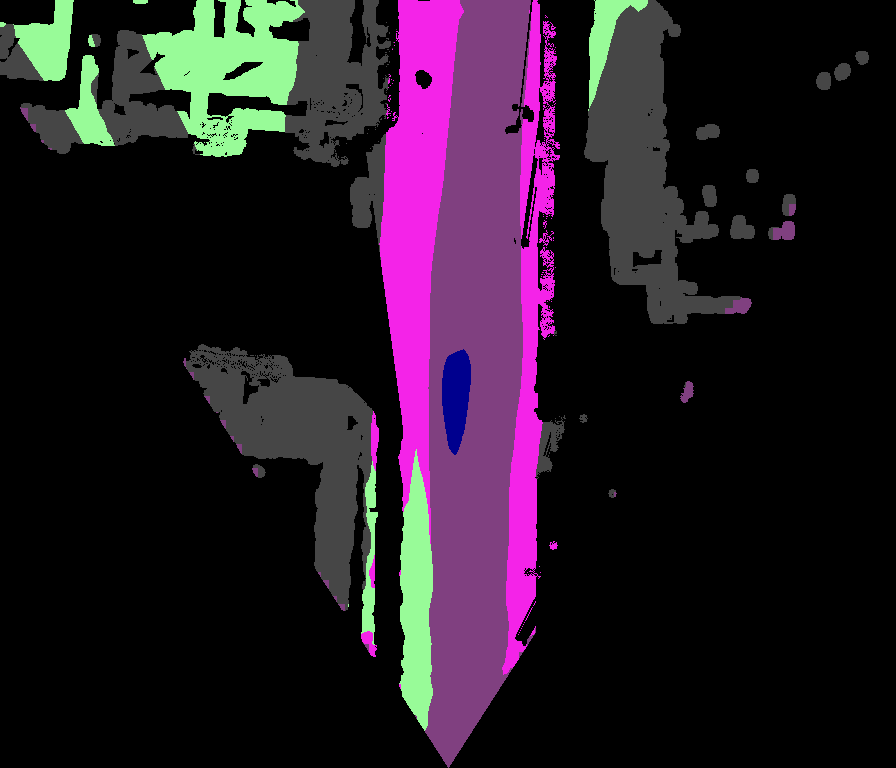}}  \\
\\
(h) & {\includegraphics[width=\linewidth, height=0.43\linewidth, frame]{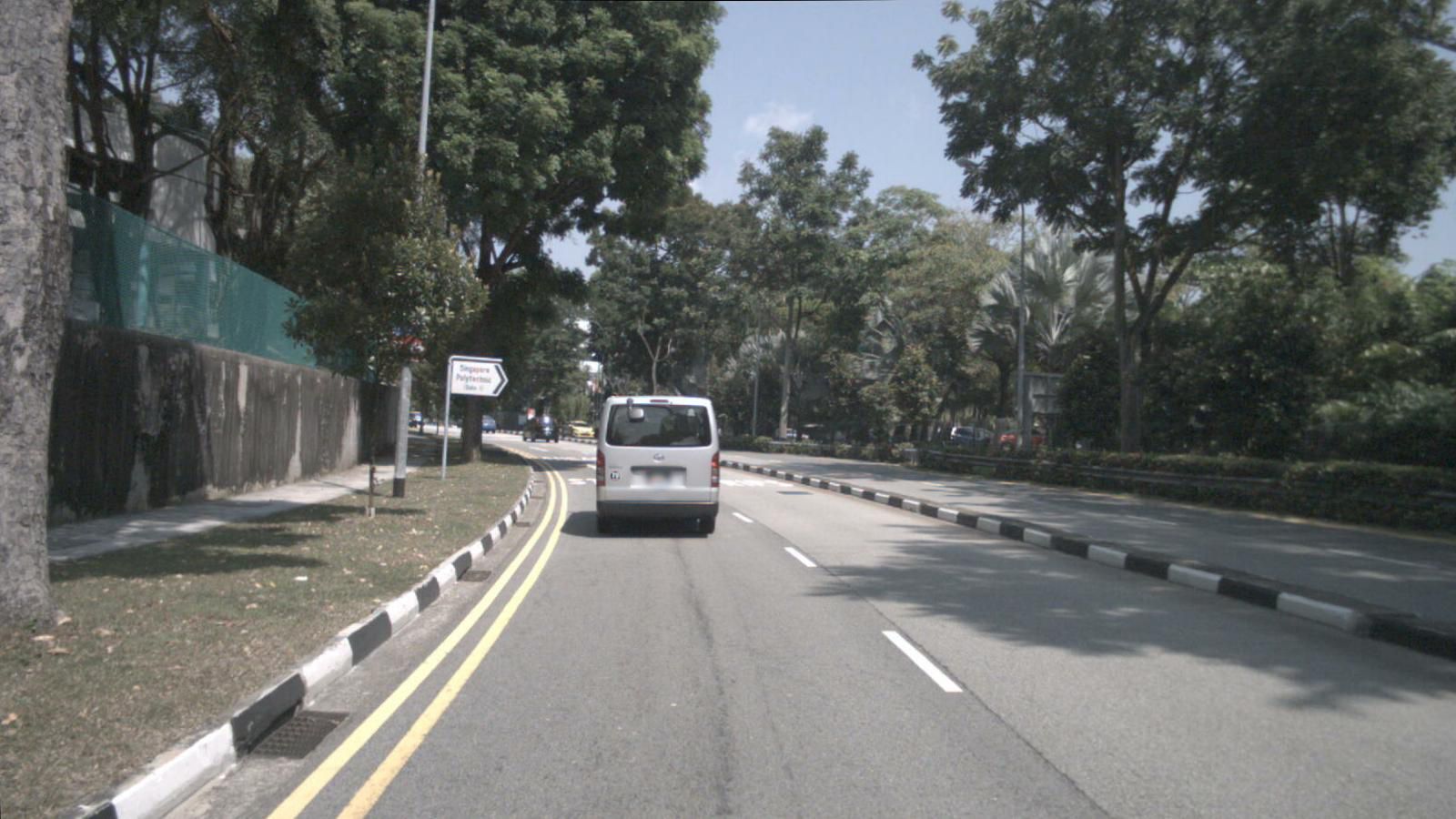}} & 
{\includegraphics[width=\linewidth, frame]{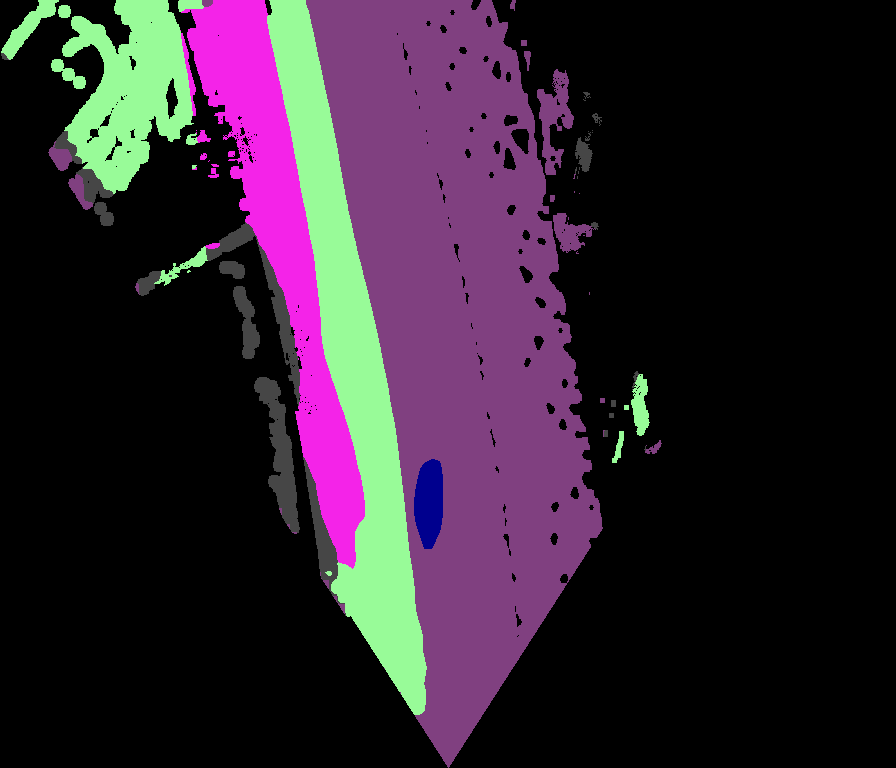}} & {\includegraphics[width=\linewidth, frame]{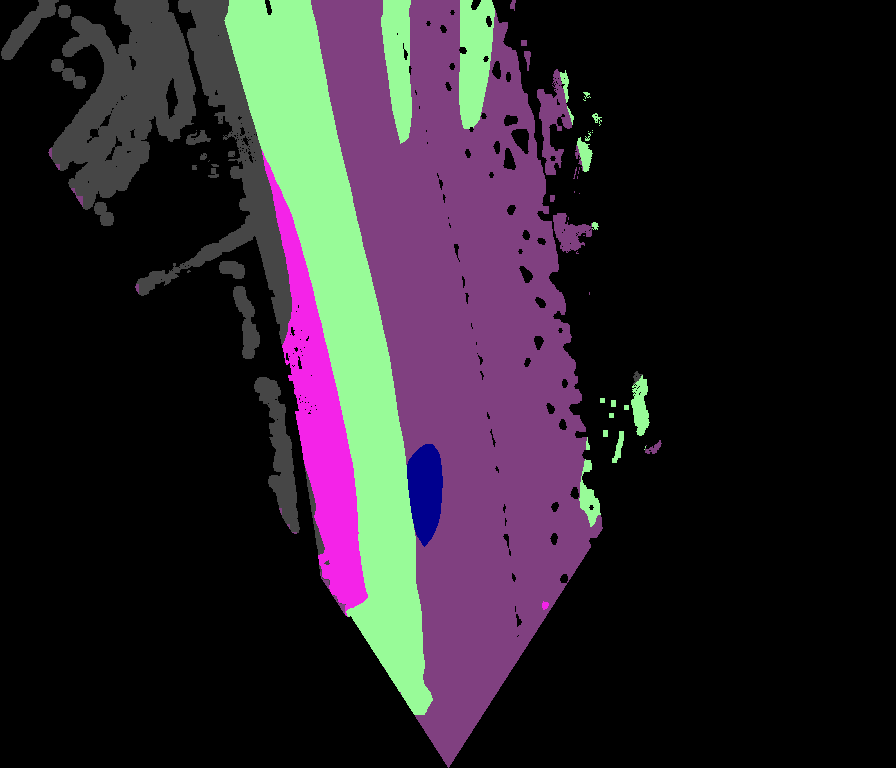}} & 
{\includegraphics[width=\linewidth, frame]{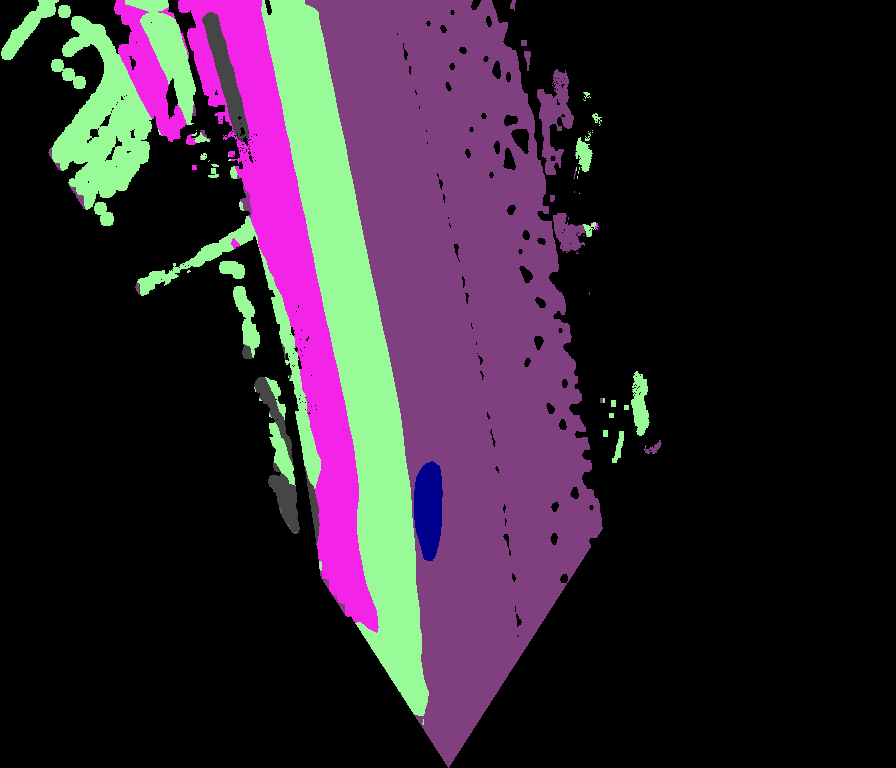}} & {\includegraphics[width=\linewidth, frame]{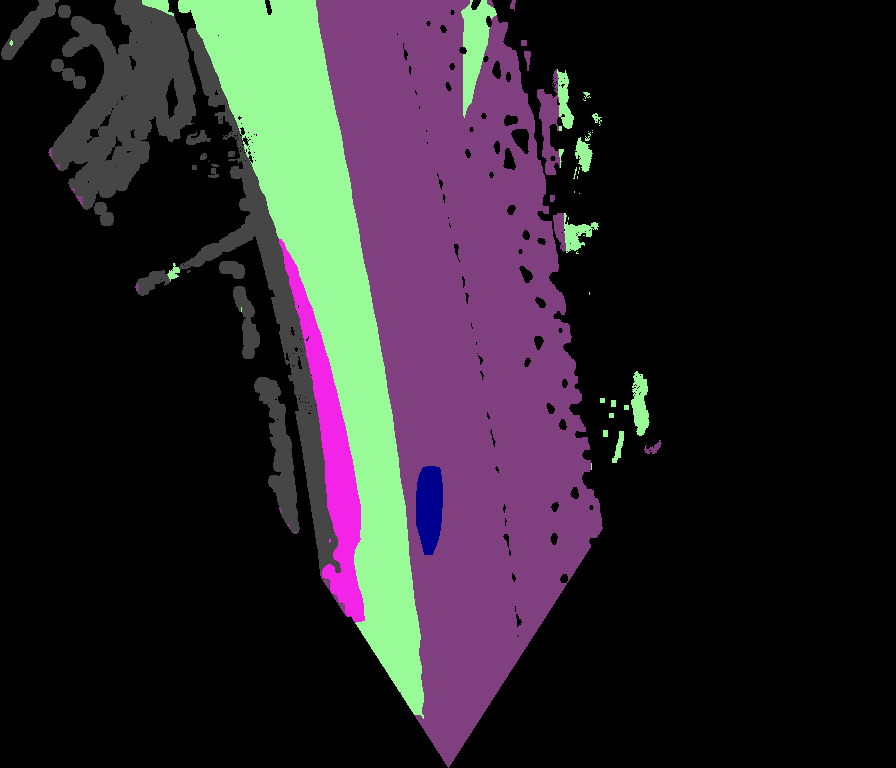}} & {\includegraphics[width=\linewidth, frame]{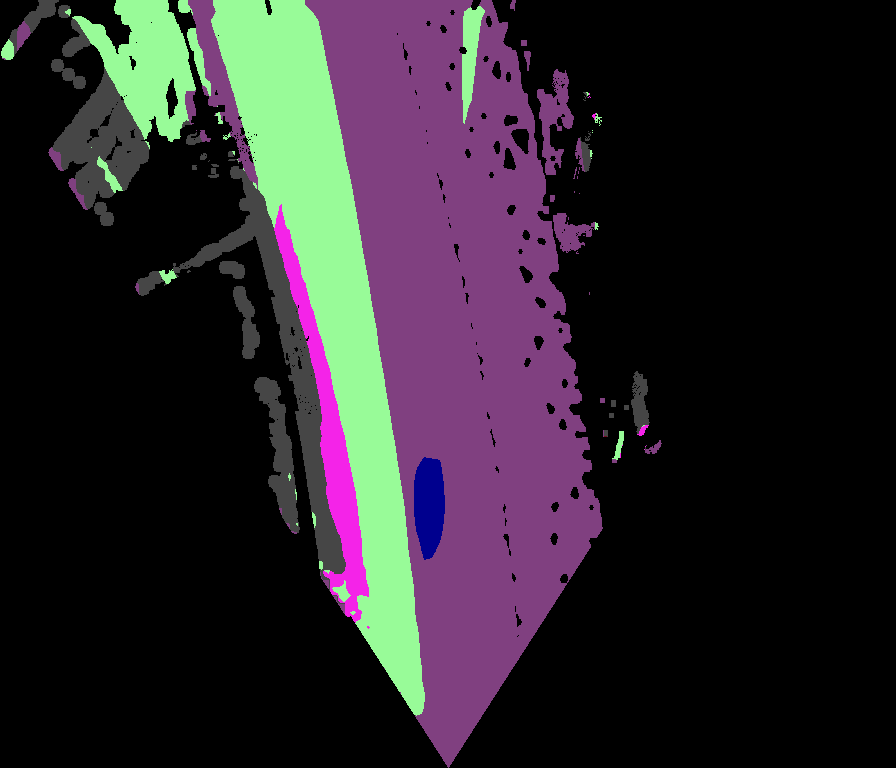}}   \\
\end{tabular}
}
\caption{Qualitative results obtained when \net~is finetuned using $1\%$, $5\%$, $10\%$, $50\%$ and $100\%$ of labels in BEV. Figures (a-d) depict predictions on the KITTI-360 dataset, while figures (e-h) show the predictions on the nuScenes dataset.}
\label{fig:supp-percentage-results}
\vspace{-0.3cm}
\end{figure}

In this section, we qualitatively evaluate the performance of our model by comparing the semantic BEV maps obtained when the amount of BEV supervision is gradually increased from $1\%$ to $100\%$. \figref{fig:supp-percentage-results} presents the results of this evaluation. \figref{fig:supp-percentage-results}(a, b, c, d) present the results on the KITTI-360 dataset and \figref{fig:supp-percentage-results}(e, f, g, h) present the results on the nuScenes dataset.

{\parskip=3pt
\noindent We observe that the semantic BEV map predictions are largely consistent across all the percentage splits of the two datasets with only minor differences pertaining to the predicted object extents. This behavior is evident in \figref{fig:supp-percentage-results}(d,~f) where the model finetuned with $1\%$ of BEV data tends to stretch objects along the radial direction, while models finetuned with higher percentage splits are not significantly affected by this factor. Moreover, we note that the $1\%$ model is able to both detect and localize all the objects in the BEV map to a high degree of accuracy, with only minor errors in the heading of the detected objects (\figref{fig:supp-percentage-results}(c)). Further, we observe in \figref{fig:supp-percentage-results}(a, f, h) that the model finetuned with $1\%$ labels is able to accurately reason about occlusions in the scene, such as the road behind the truck in \figref{fig:supp-percentage-results}(a) and the regions beyond the curve in the road in \figref{fig:supp-percentage-results}(h). This occlusion handling ability stems from the use of an independent implicit field-based geometry pathway to reason about the scene geometry in the unsupervised pretraining step. In some cases, however, the scene priors learned during the pretraining step do not generalize well to a given image input. For example, we observe in \figref{fig:supp-percentage-results}(c) that the grass patch next to the vehicle in the adjacent lane is erroneously predicted as a road for the $1\%$ model, while the models finetuned with more than $10\%$ BEV data accurately capture this characteristic.
Nonetheless, these observations reinforce the fact that our unsupervised pretraining step encourages the network to learn rich geometric and semantic representations of the scene which allows models finetuned with extremely small BEV percentage splits to generate accurate BEV maps. 
}


\end{document}